\documentclass[preprint,11pt]{article}

\usepackage{fullpage}
\usepackage[authoryear]{natbib}
\ifdefined\usebigfont

\usepackage{boondox-cal}
\usepackage{times}
\usepackage[italicdiff]{physics}
\usepackage[fontsize=13pt]{scrextend}
\usepackage[left=1.56in,right=1.56in,top=1.74in,bottom=1.74in]{geometry}
\usepackage{bigints}
\usepackage{csquotes}
\else
\fi

\usepackage{amssymb,amsmath,amsthm,amscd,dsfont,mathrsfs,bbm}
\usepackage{graphicx,float,psfrag,epsfig,amssymb}
\usepackage[usenames,dvipsnames,svgnames,table]{xcolor}
\definecolor{darkgreen}{rgb}{0.0,0,0.9}
\usepackage[pagebackref,letterpaper=true,colorlinks=true,pdfpagemode=none,citecolor=OliveGreen,linkcolor=BrickRed,urlcolor=BrickRed,pdfstartview=FitH]{hyperref}
\usepackage{wrapfig}
\usepackage{relsize}
\usepackage{color}
\usepackage{pict2e}
\usepackage{subcaption}
\usepackage{nameref}
\usepackage{makecell}
\usepackage[font={small}]{caption} 
\usepackage{mathtools}
\usepackage{enumitem}

\let\chapter\section
\usepackage[ruled,vlined]{algorithm2e}
\usepackage[noend]{algorithmic}
\usepackage{diagbox}
\usepackage{booktabs,tabularx}

\DeclareMathAlphabet{\mathpzc}{OT1}{pzc}{m}{it}

\newtheorem{propo}{Proposition}[section]
\newtheorem{lemma}[propo]{Lemma}

\newtheorem{assumption}{Assumption}
\newtheorem{proposition}[propo]{Proposition}
\newtheorem{coro}[propo]{Corollary}
\newtheorem{thm}[propo]{Theorem}

\newtheorem{defi}{Definition}

\theoremstyle{definition}

\def\cF{{\mathcal F}}
\def\cA{{\mathcal A}}
\def\cR{{\mathcal R}}

\def\cS{{\mathcal S}}

\def\reals{{\mathbb R}}

\def\eps{{\varepsilon}}
\def\prob{{\mathbb P}}
\def\E{{\mathbb E}}

\def\var{{\rm Var}}

\def\L0{{L_i}}

\def\de{{\rm d}}
\def\<{\langle}
\def\>{\rangle}

\def\hth{\widehat{\theta}}

\def\F{{\sf F}}
\def\ind{{\mathbb I}}

\def\F{{\sf F}}
\def\normal{{\sf N}}

\def\sT{{\sf T}}

\def\cov{{\rm Cov}}

\def\v*{v_i}
\def\T*{T_i}

\def\u*{u_i}
\def\F*{F_i}

\def\htheta{\widehat{\theta}}
\def\corr{{\rm corr}}

\def\th{\theta}
\def\hth{{\widehat{\theta}}}

\def \hq{\widehat{q}}

\def\l1u{W}



\DeclareMathAlphabet{\mathpzc}{OT1}{pzc}{m}{it}

\def\hw{\widehat{w}}

\def\normal{\mathsf{N}}

\def\tv{\mathsf{TV}}
\newcommand{\pinf}[1]{p^{\gamma}_{e}(#1;p_0)}

\def \hc{\widehat{c}}

\def\hb{\widehat{b}}

\def\hV{\widehat{V}}

\def \hQ{\widehat{Q}}
\def \cS{\mathcal{S}}
\def \cA{\mathcal{A}}
\def\hv{\hat{v}^{\gamma}}
\def\hq{\hat{q}^{\gamma}}
\def\homg{{\hat{\omega}}}
\def\hpsi{\widehat{\psi}}
\def\rhot{ {\hat{\th}}^{\mathsf{TDR}}}
\def\tdr{\mathsf{TDR}}
\def\dr{\mathsf{DR}}
\def\mse{\mathsf{MSE}}
\def \hW{\widehat{W}}

\def\myavg{\th}
\def\mytdravg{\hth^{\tdr}}
\def\mydisavg{\th_{\gamma}(p_0)}
\def\mytdrdisavg{\hat{\th}_{\gamma}^{\tdr}(p_0)}

\title{Off-Policy Evaluation in Markov Decision Processes \\ under Weak Distributional Overlap}

\author{Mohammad Mehrabi \\ \texttt{mehrabi@stanford.edu}
\and Stefan Wager \\ \texttt{swager@stanford.edu}}
\date{\today}

\begin{document}


\maketitle

\begin{abstract}
Doubly robust methods hold considerable
promise for off-policy evaluation in Markov decision processes (MDPs) under sequential ignorability:
They have been shown to converge as $1/\sqrt{T}$ with the horizon $T$, to be statistically
efficient in large samples, and to allow for modular implementation where preliminary estimation
tasks can be executed using standard reinforcement learning techniques. Existing results, however,
make heavy use of a strong distributional overlap assumption whereby the stationary distributions of
the target policy and the data-collection policy are within a bounded factor of each other---and
this assumption is typically only credible when the state space of the MDP is bounded. In this paper, we
re-visit the task of off-policy evaluation in MDPs under a weaker notion of distributional overlap,
and introduce a class of truncated doubly robust (TDR) estimators which we find to perform well
in this setting. When the distribution ratio of the target
and data-collection policies is square-integrable (but not necessarily bounded), our approach
recovers the large-sample behavior previously established under strong distributional overlap.
When this ratio is not square-integrable, TDR is still consistent but with a slower-than-$1/\sqrt{T}$-rate; furthermore, this rate of convergence is minimax over a class of
MDPs defined only using mixing conditions. We validate our approach numerically and find that,
in our experiments, appropriate truncation plays a major role in enabling accurate off-policy
evaluation when strong distributional overlap does not hold.
\end{abstract}

\vspace{1cm}



\section{Introduction}

Dynamic policies, i.e., treatment policies that repeatedly interact with a unit
in order to achieve desirable outcomes over time, are of central interest in a
number of application areas such as healthcare 
\citep{murphy2001marginal,kitsiou2017effectiveness,movsisyan2019adapting,liao2021off},
recommendation systems \citep{li2011unbiased}, and education \citep{nahum2019introduction}.
In order to design good dynamic policies, it is important to be able to use samples
collected from an experiment to assess how different candidate policies would
perform if deployed. This task is commonly referred to as off-policy evaluation.

Without assumptions, off-policy evaluation suffers from a curse of dimensionality, and the
best achievable error guarantees blow up exponentially with the number of time periods
over the policy can act \citep{robins1986new,laan2003unified,jiang2016doubly,thomas2016data}.
However, this curse of dimensionality disappears in systems with Markovian dynamics, and
in fact data collected along a long trajectory can improve accuracy. These observations
have led to a recent surge of interest in off-policy evaluation in Markov decision processes
\citep{liu2018breaking,liao2021off,kallus2020double,kallus2022efficiently,liao2022batch,hu2023off}.

Markov Decision Processes (MDPs) \citep{puterman2014markov} model dynamic policies using a set of
states, actions, and reward values under an assumption that, given the current state and taken
action, the next state is independent of the previous history of states and actions. Under this
setting, doubly robust methods that estimate both the $Q$-function of the target policy and
its stationary state-distribution (and then use these estimates to debias each other)
have particularly strong statistical properties \citep{kallus2022efficiently,liao2022batch}.
They are statistically efficient; furthermore, they are first-order robust to errors in the
$Q$-function and stationary distribution estimates, thus enabling a practical and modular
implementation via the double machine learning framework \citep{chernozhukov2018double}.

One limitation of the existing work on doubly robust off-policy evaluation, however, is that
it relies on a {\it strong distributional overlap} assumption. Write $p_b(s)$ for
the stationary distribution of the data-collection (or behavior) policy, and $p_e(s)$
for the stationary distribution of the target (or evaluation) policy. Strong distributional
overlap requires that $\omega(s) := p_e(s) / p_b(s) \leq M$ for some finite constant $M$, uniformly
across all states $s$. This assumption may be acceptable when the state space is finite;
however, it fails in many of even the simplest examples involving unbounded domains.
For example, in a queuing model where the state $s$ measures the queue length, if the
behavior policy induces the state to evolve as in an $M/M/1$ queue and the evaluation
policy succeeds in increasing the arrival rate of the queue, then strong distributional
overlap will not hold.

In this paper, we re-visit off-policy evaluation in MDPs under a {\it weak distributional
overlap} assumption: Instead of requiring the ratio $\omega(s)$ to be bounded, we
only posit a (polynomial) tail bound on its distribution. Under this setting, we find that
standard doubly robust (DR) methods \citep{kallus2022efficiently,liao2022batch} can be brittle.
However, we also find that this brittleness can be addressed via truncation, resulting
in a new estimator which we refer to as the truncated doubly robust (TDR) estimator. We
then show the following:
\begin{itemize}
\item When the distribution ratio $\omega(s)$ is square-integrable (under the behavior
policy), the TDR estimator achieves similar guarantees as were previously established
for the DR method under strong distributional overlap. This includes a $1/\sqrt{T}$ rate
of convergence in the horizon $T$, a central limit theorem, statistical efficiency, and
first-order robustness to estimation error in the algorithm inputs.
\item When the distribution ratio $\omega(s)$ is not square-integrable, the TDR estimator
is consistent but no longer achieves a $1/\sqrt{T}$ rate of convergence. Instead, its
rate of convergence depends on the tail-decay rate of $\omega(s)$.
\item In a class of off-policy evaluation problems with MDPs characterized by
mixing conditions, the rate of convergence of TDR matches the
minimax lower bound for off-policy evaluation derived in \citet{hu2023off}, i.e.,
TDR is minimax rate-optimal.
\end{itemize}
Finally, we validate the numerical performance of TDR and find that,
in our experiments, appropriate truncation is crucial when the distribution ratio
$\omega(s)$ sometimes takes on large values.

\subsection{Related work}

Strong distributional overlap as considered in this
paper is closely related to notions such as the concentrability coefficient and full coverage condition often used in the
reinforcement learning literature. The density-ratio-based concentrability coefficient \citep{zhan2022offline} and
the full coverage condition \citep{uehara2021pessimistic} requires a uniform bound on the density ratio function of 
the joint state and action $(s,a)$ distributions for evaluation and behavior policies, i.e., $\frac{p_e(s,a)}{p_b(s,a)}<B$,
for all states $s$ and actions $a$, and some finite value $B$. This condition is essentially equivalent to
strong distributional overlap together with what we call policy overlap (Assumption \ref{assu: policy}).
There are a number of methods developed in offline and online reinforcement learning under full coverage
and concentrability conditions, e.g., \citet{bennett2021off}, \citet{liu2020provably}, \citet{uehara2021pessimistic},
\citet{xie2022role}, \citet{zhan2022offline},
\citet{xie2019towards}, \citet{yang2022offline}, and
\citet{kallus2020double}.

To highlight a few works for the off-policy evaluation problem under strong distributional overlap conditions, \citet{kallus2022efficiently}, while working under strong distributional overlap, derived efficiency bounds \citep{van2000asymptotic} for off-policy evaluation in discounted average-reward MDPs, characterizing the minimum limit of the square-root-scaled MSE. In addition, they demonstrated that the doubly robust estimator, when deployed in the infinite horizon (a trajectory whose length approaches infinity) for both non-Markovian decision processes \citep{jiang2016doubly} and MDPs \citep{kallus2020double}, will achieve this error bound.

\citet{liao2022batch} derived the efficiency bound for off-policy evaluation in long-run average reward MDPs and provided a doubly robust estimator that achieves efficiency; in doing so, they again made positivity assumptions that imply strong distributional overlap. Moreover, \citet{bennett2021off} considers the off-policy evaluation problem when there exist unobserved confounders. Their proposed solution works under the assumption of state visitation overlap, which implies the strong distributional overlap condition. In addition, the work by \citet{chandak2021universal} explores the challenge of OPE encompassing various statistical measures and functionals such as mean, variance, quantiles, inter-quantile range, and CVaR. The authors establish a comprehensive framework for estimating a broad range of parameters related to the evaluation policy, going beyond the expected return value. This estimation is carried out within the context of a finite-length horizon and under the assumption of strong distributional overlap.

{
The focus of this work is on causal inference in MDPs where strong distributional overlap fails. We would like to highlight that this notion of \textit{distributional} overlap in dynamic treatment regimes is distinct from the classical \textit{policy} overlap assumption in causal inference \citep{imbens2004nonparametric}. Distributional overlap requires that the stationary distributions induced under two policies overlap (see the detailed discussion at the beginning of Section \ref{sec: tdr-estimator}), whereas policy overlap pertains to contiguity of the treatment and control decision rules conditionally on state. Distributional overlap can be violated when policy overlap holds, and vice versa. That said, violations of policy overlap in classical causal inference have been studied extensively---and the truncation-based algorithmic techniques we use to address distributional overlap here are inspired by that literature. \citet{crump2009dealing} consider treatment effect estimation under failures of policy overlap and consider how best to select a subsample over which the average treatment effect can be precisely estimated; \citet{li2018balancing} consider the same task using weighting rather than sample trimming. \cite{yang2018asymptotic} proposes an alternative to the sample trimming procedure of \citet{crump2009dealing} by smoothly reweighting samples based on propensity scores, mitigating the potential instability caused in certain cases by sample trimming ($0/1$ weighting). \citet{d2021overlap} and \citet{mou2023kernel} provide further discussion and recent references.

}

\section{Doubly robust evaluation in MDPs}
We consider the classic reinforcement learning framework where, in periods $t = 0, 1, 2,\ldots$, we observe a state variable
$S_t \in \cS$, an action $A_t \in \cA$, and a reward $R_t \in \mathbb{R}$. We assume
the state space $\cS$ is a measurable space with base measure $\lambda_S$; this measure may
for example be a Lebesgue measure, counting measure, or mixture thereof.
We take the action space $\cA$ to be discrete and finite. A policy $\pi$ is a
mapping from the state $S_t$ to a (potentially randomized) action $A_t$.
We make the following assumptions throughout. The random exploration and
policy overlap assumptions are standard assumptions in causal inference
\citep{hernan2020whatif,imbens2015causal}; one notable dynamic setting where these
assumptions hold is in micro-randomized trials \citep{klasnja2015microrandomized}.
The $\rho$-mixing condition is a standard mixing condition for Markov chains,
and implies the widely used $\alpha$-mixing condition
\citep{bradley2005basic, davidson1994stochastic}.

\begin{assumption}[time-invariant MDP]
\label{assu:MDP}
There exist distributions $p_s$ and $p_r$ such that, for each time $t$,
\begin{equation}
\begin{split}
&R_t \,|\, A_t, \, S_t, \, R_{t-1}, \, A_{t-1}, \, S_{t-1}, \, \ldots \sim p_r(\cdot \,|\, S_t, \, A_t), \\
&S_{t+1} \,|\, R_t, \, A_t, \, S_t, \, R_{t-1}, \, A_{t-1}, \, S_{t-1}, \, \ldots \sim p_s(\cdot \,|\, S_t, \, A_t).
\end{split}
\end{equation}
\end{assumption}

\begin{assumption}[random exploration]
\label{assu:random}
Our data is collected under a known random behavior policy $\pi_b$ such that, for each time $t$,
\begin{equation}
\mathbb{P}\left(A_t = a \,|\, S_t, \, R_{t-1}, \, A_{t-1}, \, S_{t-1}, \, \ldots\right) = \pi_b(a \,|\, S_t).
\end{equation}
\end{assumption}

\begin{assumption}[policy overlap]\label{assu: policy}
We seek to evaluate a policy $\pi_e$ whose actions are consistent with those taken in
the data, in the sense that there exists a constant $C_\eta$ such that
$\pi_e(a \,|\, s) \,/\,\pi_b(a \,|\, s) \leq C_\eta$ for all $a \in \cA$ and $s \in \cS$.
\end{assumption}

\begin{assumption}[$\rho$-mixing]
\label{assu:rho-mix}
The data collected under the behavior policy is $\rho$-mixing: Writing
\begin{equation}
\label{eq:mixings-rho}
\rho_k =\sup_{t\geq 1} \sup_{f, \, g \in L^2} |\corr(f(S_t, \, A_t, \, R_t), \, g(S_{t+k}, \, A_{t+k}, \, R_{t+k}))|,
\end{equation}
we assume that there is a finite constant $C_\rho$ for which \smash{$\sum\limits_{k = 1}^\infty \rho_k \leq C_\rho$}.
\end{assumption}

\begin{assumption}[stationary distribution]
\label{assu:stat}
The behavior policy $\pi_b$ induces a stationary distribution $p_b$ for the
state variable $S_t$, and our data collection is initialized from this stationary
distribution, $S_0 \sim p_b$.
\end{assumption}

When evaluating a target policy $\pi_e$, we consider both the expected discounted
reward with a discount rate $0 < \gamma < 1$, and the expected long-run reward, which are given by
\begin{equation}
\begin{split}
&\mydisavg = (1 - \gamma) \, \mathbb{E}_{\pi_e}\left[\sum_{t = 0}^\infty \gamma^t R_t \ \big| \ S_0 \sim p_0\right], \\
&\myavg = \lim_{T \rightarrow \infty} \mathbb{E}_{\pi_e}\left[\frac{1}{T+1} \sum_{t = 0}^T R_t \right],
\end{split}
\end{equation}
where $\mathbb{E}_{\pi_e}$ samples over trajectories with actions chosen according
to the policy $\pi_e$ and states and rewards generated as in Assumption \ref{assu:MDP},
and $p_0$ is an initial state distribution. The existence of a long-run average reward
$\th$ depends on the evaluation policy also inducing a stationary distribution $p_e$
on the state variable. When such a stationary distribution exists, 
$\lim_{\gamma \rightarrow 1} \mydisavg = \myavg$ for all $p_0$ because
the influence of the initializes washes out thanks to mixing, and so we do not make
the dependence on the initial distribution explicit.

The DR estimator for off-policy evaluation in MDPs was originally introduced for the
discounted setting \citep{kallus2020double}. Define the discounted $Q$-function and
associated value function as
\begin{equation}
\label{eq:qgamma}
q_e^\gamma (s, \, a)= \E_{\pi_e}\left[\sum_{t=0}^{\infty} \gamma^t R_t \ \big| \ S_0=s, \, A_0=a \right], \ \ \ \
v_e^\gamma(s) = \sum_{a \in \cA} \pi_e(a \, | \, s) q_e^\gamma (s,a).
\end{equation}
Furthermore, define the discounted state distribution under the evaluation
policy as
\begin{equation}
p_e^\gamma(\cdot; \, p_0) = (1 - \gamma) \sum_{t=0}^{\infty} \gamma^t \, \mathbb{P}_{\pi_e}\left( S_t = \cdot \ \big| \ S_0 \sim p_0 \right)\,,
\end{equation}
and the induced distribution ratio
\begin{equation}
\label{eq:omega_gamma}
\omega^\gamma(s; \, p_0) = \frac{p_e^\gamma(s; \, p_0)}{p_b(s)}\,.
\end{equation}
Then, given any estimates $\hat{q}_e^\gamma(s, \, a)$ and $\hat{\omega}^\gamma(s; \, p_0)$
and $T$ samples from our data-collection process (collected under behavior policy), the doubly robust estimate for $\mydisavg$ is
\begin{equation}
\label{eq:DRgamma}
\begin{split}
\htheta_\gamma^{\dr}(p_0) &= (1 - \gamma) \int \hat{v}_e^\gamma(s) p_0(s) d\lambda_{\cS}(s) \\
 &\quad\quad +\frac{1}{T} \sum_{t = 0}^{T - 1} \hat{\omega}^\gamma(S_t; \, p_0) \frac{\pi_e(A_t \, | \, S_t)}{\pi_b(A_t \, | \, S_t)} \left(R_t + \gamma \hat{v}_e^\gamma(S_{t+1}) - \hat{q}_e^\gamma(S_t, \, A_t) \right),
 \end{split}
\end{equation}
where $\hat{v}$ is induced by $\hat{q}$ via \eqref{eq:qgamma}. Under the strong
distributional overlap assumption, which here amounts to $\omega^\gamma(s; \, p_0) \leq M$
for all $s$, this doubly robust estimator has been shown to be efficient and first-order
robust to estimation errors in $\hat{q}_e^\gamma(s, \, a)$ and $\hat{\omega}^\gamma(\cdot; \, p_0)$
\citep{kallus2022efficiently}.

Constructing the doubly robust estimator for the long-run average reward requires
some minor modifications \citep{liao2022batch}. Assuming that the quantities below
exist (we will provide assumptions that guarantee this before stating our formal
result), define the differential $Q$-function and induced differential value function
as  \citep{van1998learning, liu2018breaking}:
\begin{align}\label{eq: Q-V}
Q_e(s,a)=\lim\limits_{T\to \infty}^{} \E_{\pi_e}\left[\sum_{t=0}^{T}(R_t-\myavg) \ \Big| \ S_0=s, \, A_0=a \right], \ \ \ 
V_e(s)= \sum_{a \in \cA} \pi_e(a \, | \, s) Q_e (s,a).
\end{align}
Let $p_e(s)$ be the stationary state distribution induced by the evaluation policy,
and let $\omega(s) = p_e(s) / p_b(s)$. Then, given any estimates \smash{$\widehat{Q}_e(s, \, a)$} and
$\homg(s)$, and the induced \smash{$\widehat{V}_e(s)$}, the doubly robust estimate for $\myavg$ is
\begin{equation}
\label{eq:DRlong}
\htheta_e^{\dr} = \frac{\sum\limits_{t = 0}^{T - 1} \hat{\omega}(S_t) \frac{\pi_e(A_t \, | \, S_t)}{\pi_b(A_t \, | \, S_t)} \left(R_t + \widehat{V}_e(S_{t+1}) - \widehat{Q}_e(S_t, \, A_t) \right)}{\sum\limits_{t = 0}^{T - 1} \hat{\omega}(S_t)\frac{\pi_e(A_t \, | \, S_t)}{\pi_b(A_t \, | \, S_t)}} .
\end{equation}
Note that, unlike in \eqref{eq:DRgamma}, we here use a self-normalized form; and this
ends up being required for our desired properties to hold.

\subsection{Review: Why doubly robust methods work}

Before moving to our main results, we here briefly review to motivation
behind the construction of the doubly robust estimator by showing that
it satisfies a ``weak double robustness property'': If either of the estimates
$\homg^{\gamma}$ or $\hq$ is known and correct then the $\dr$ estimator is
unbiased (regardless of misspecification of the other component). This discussion
will also enable us to state some key technical lemmas that will be used throughout.

The doubly robust estimator can be understood in terms of the two Bellman
equations given below. These results are stated in \citet{kallus2022efficiently}
for the discounted case and \citet{liu2018breaking} for the long-run average case; however,
for the convenience of the reader, we also give proofs of these results in
Sections \ref{proof: lemma: bell-q} and \ref{proof: lemma-omega} respectively.

\begin{lemma}[\citet{liu2018breaking, kallus2022efficiently}]\label{lemma: bell-q}
Given the tuple $(S,A,R,S')$ following Assumption \ref{assu:MDP}, then the following Bellman equations hold
\begin{align*}
q^{\gamma}_e(S,A)&=\E[R+\gamma v_e^{\gamma}(S')|S,A]\\
Q_e(S,A)+\myavg&=\E[R+ V_e(S')|S,A]\,.
\end{align*}
\end{lemma}

\begin{lemma}[\citet{liu2018breaking, kallus2022efficiently}]\label{lemma: bell-prob}
For every measurable function $f:\cS\to \reals$, for $(S,A,R,S')$ under Assumptions \ref{assu:MDP} and \ref{assu:random} we have
\[
\E[\omega^{\gamma}(S;p_0)f(S)]=(1-\gamma)\int f(s) p_0(s)\de \lambda_{\cS}(s) +\gamma \E\left[ \omega^{\gamma}(S;p_0)\frac{\pi_e(A|S)}{\pi_b(A|S)}f(S') \right]\,.
\]
In addition, for the density ratio function $\omega(.)$, the following holds
\[
\E_{}\left[\omega(S)f(S)\right]=\E_{}\left[\omega(S)\frac{\pi_e(A|S)}{\pi_b(A|S)}f(S')\right]\,. 
\] 
\end{lemma}

We start by considering the case when $\hq_e=q^{\gamma}_e$, and verify that $\dr$ is
consistent regardless of our choice of $\homg^{\gamma}$:
\begin{align*}
\E\left[\htheta_\gamma^{\dr}(p_0)\right]&= (1 - \gamma) \int {v}_e^\gamma(s) p_0(s) d\lambda_{\cS}(s) \\
 &\quad\quad +\frac{1}{T} \sum_{t = 0}^{T - 1} \E\left[\hat{\omega}^\gamma(S_t; \, p_0) \frac{\pi_e(A_t \, | \, S_t)}{\pi_b(A_t \, | \, S_t)} \left(R_t + \gamma {v}_e^\gamma(S_{t+1}) - {q}_e^\gamma(S_t, \, A_t) \right)\right]\\
&=\mydisavg+\frac{1}{T}\sum\limits_{t=0}^{T-1}\E\left[ \hat{\omega}^\gamma(S_t; \, p_0) \frac{\pi_e(A_t \, | \, S_t)}{\pi_b(A_t \, | \, S_t)} \E\left[R_t + \gamma {v}_e^\gamma(S_{t+1}) - {q}_e^\gamma(S_t, \, A_t)|S_t,A_t\right]\right]\\
&=\mydisavg\,,
\end{align*}
where the last line follows from Lemma \ref{lemma: bell-q}.
The derivation for the long-run average case is analogous. From this calculation, we also
immediately see that if $\hq_e$ is consistent for $q^{\gamma}_e$, the $\dr$ estimator will also
generally be consistent for $\theta^\gamma(p_0)$.

We next consider the second scenario when $\hat{\omega}^{\gamma}(.;p_0)=\omega(.;p_0)$, and
start by noting that
\begin{align*}
\E\left[\htheta_\gamma^{\dr}(p_0)\right]&=\mydisavg+ (1 - \gamma) \int ({\hat{v}}_e^\gamma-v_e^{\gamma})(s) p_0(s) d\lambda_{\cS}(s) \\
 &\quad\quad +\frac{1}{T} \sum_{t = 0}^{T - 1} \E\left[{\omega}^\gamma(S_t; \, p_0) \frac{\pi_e(A_t \, | \, S_t)}{\pi_b(A_t \, | \, S_t)} \left(R_t + \gamma {\hat{v}}_e^\gamma(S_{t+1}) - {\hat{q}}_e^\gamma(S_t, \, A_t) \right)\right]\,.
\end{align*}
We can then apply Lemma \ref{lemma: bell-prob} with $f(s)=\hv_e(s)-v_e^{\gamma}(s)$ and
invoke the relation $\hv(s)=\E_{\pi_e}[\hq_e(s,a)|s]$ to get
\begin{align*}
\E\left[\htheta_\gamma^{\dr}(p_0)\right]&=\mydisavg
+\frac{1}{T} \sum_{t = 0}^{T - 1} \E\left[{\omega}^\gamma(S_t; \, p_0) \frac{\pi_e(A_t \, | \, S_t)}{\pi_b(A_t \, | \, S_t)} \left(R_t + \gamma {{v}}_e^\gamma(S_{t+1}) - {{q}}_e^\gamma(S_t, \, A_t) \right)\right]\\
&=\mydisavg\,,
\end{align*}
where the last line follows from Lemma \ref{lemma: bell-q}. The long-run average
problem is again similar.

\section{The truncated doubly robust estimator}\label{sec: tdr-estimator}

As discussed in the introduction, the main focus of this paper is in extending
our methodological and formal understanding of doubly robust methods in MDPs
to the case where strong distributional overlap may not hold, i.e., where
$\omega(s)$ may grow unbounded. Throughout this paper, we will only assume the tail
bounds as below.

\begin{defi}[weak distributional overlap]
Let $\delta > 0$.\footnote{Note that, by Markov's inequality, $0$-weak distributional
overlap always trivially holds with $C = 1$.}
Given a behavior policy $\pi_b$ with an induced state-stationary distribution $p_b$ and
an evaluation policy $\pi_e$ with an induced state-stationary distribution $p_e$, we
say that long-run $\delta$-weak distributional overlap holds if there is a constant $C$
such that, for all $x > 0$,
\begin{equation}\label{eq: weak-exponent}
\prob_{p_b}\left(\omega(S)^{1+\delta} \ge x\right) \leq \frac{C}{x}, \ \ \ \ \omega(s) = \frac{p_e(s)}{p_b(s)}.
\end{equation}
Similarly, using notation as in \eqref{eq:omega_gamma}, we say that $\gamma$-discounted
$\delta$-weak distributional overlap holds if there is a constant $C$
such that, for all $x > 0$,
\begin{equation}
\prob_{p_b}\left(\omega^\gamma(S;p_0)^{1+\delta} \ge x\right) \leq \frac{C}{x}.
\end{equation}
\end{defi}

The algorithmic adaptation we use to address challenges associated with weak distributional
overlap is extremely simple: We choose a truncation level $\tau_t$, and then we truncate the
$\omega$-estimates used in the DR estimator at this level,

\begin{equation}
\label{eq:TDR}
\begin{split}
&\htheta_\gamma^{\tdr}(p_0) = (1 - \gamma) \int \hat{v}_e^\gamma(s) p_0(s) d\lambda_{\cS}(s) \\
 &\quad\quad\quad\quad\quad\quad +\frac{1}{T} \sum_{t = 0}^{T - 1} \left(\hat{\omega}^\gamma(S_t; \, p_0)\land \tau_t\right) \frac{\pi_e(A_t \, | \, S_t)}{\pi_b(A_t \, | \, S_t)} \left(R_t + \gamma \hat{v}_e^\gamma(S_{t+1}) - \hat{q}_e^\gamma(S_t, \, A_t) \right), \\
&\htheta^{\tdr} = \frac{\sum\limits_{t = 0}^{T - 1} \left(\hat{\omega}(S_t)\land \tau_t\right) \frac{\pi_e(A_t \, | \, S_t)}{\pi_b(A_t \, | \, S_t)} \left(R_t + \widehat{V}_e(S_{t+1}) - \widehat{Q}_e(S_t, \, A_t) \right)}{\sum\limits_{t = 0}^{T - 1} \Big(\hat{\omega}(S_t) \land \tau_t\Big)\frac{\pi_e(A_t|S_t)}{\pi_b(A_t|S_t)}}.
 \end{split}
\end{equation}
The motivation behind this construction is that, under weak distributional overlap,
the weights $\omega(s)$ may sometimes---but not often---take extreme values that
de-stabilize the estimator. Judicious truncation allows us to control this instability
without inducing excessive bias. Conceptually, this strategy is closely related to the
well-known variance-reduction technique of truncating inverse-propensity weights for
treatment effect estimation in causal inference \citep{cole2008constructing,gruber2022data}.

In the following sections, we show that this simple fix is a performant and versatile
solution to the challenges posed by weak distributional overlap. When $\delta$-weak distributional
overlap holds with $\delta > 1$ (and thus $\omega$ is square-integrable), the TDR estimator
can stabilize performance without inducing any asymptotic bias, and recovers properties
proven for the DR estimator under strong distributional overlap. Meanwhile, when $0 < \delta < 1$,
our approach retains consistency, and in fact attains the minimax rate of convergence
for certain classes of off-policy evaluation problems.

\subsection{Guarantees for discounted average reward estimation}

We start by considering mean-squared error guarantees for the $\tdr$ estimator in the
discounted average reward setting. As is standard in the literature on doubly robust
methods, we will allow for the input estimates $\hq_e$ and $\hat{\omega}$ to converge
slower than the target rate for the $\tdr$ estimator; however, we still require
them to be reasonably accurate. Here, for convenience, we assume that $\hq_e$ and $\hat{\omega}$
were learned on a separate training sample; in practice, it may also be of interest
to us a cross-fitting construction if a separate training sample is not available
\citep{chernozhukov2018double}. 

As a convention, we adopt the big $O$ notation to ignore absolute constants, and we let $a_T=o(b_T)$ if $a_T/b_T\to 0$ as $T\to \infty$. In addition, we write $a_T \lesssim b_T$ if $a_T = O(b_T)$.

\begin{assumption}[]\label{assumption: model-estimates}
For estimates $\hq_e$ and $\hw_e$, suppose that there exist parameters $\kappa_T$ and $\xi_T$ such that 
\begin{align}
\E_{p_0\otimes \pi_b}\left[\|\hq_e-q_e^{\gamma}\|_2^2\right] \vee \E_{p_b}[\|\hq_e-q_e^{\gamma}\|_2^2] &\leq \kappa_T^2 \,, \nonumber  \\
\E_{p_b}\left[\|\homg^\gamma(.; \, p_0)-\omega^{\gamma}(.; \, p_0)\|_2^2\right] &\leq \xi_T^2\,. \label{eq: kappa_xi}
\end{align}
\end{assumption}

\begin{thm}
\label{thm: mse}
Under Assumptions \ref{assu:MDP}-\ref{assu:stat}, consider the off-policy evaluation problem for estimating the discounted average reward $\mydisavg$, when the $\gamma$-discounted $\delta$-weak distributional overlap condition for some $\delta\le1$ holds.
In addition, suppose that the estimates $\hq_e$ and $\hat{\omega}^{\gamma}$ satisfy Assumption \ref{assumption: model-estimates} with parameters $\kappa_T, \xi_T$.
Then the $\mathsf{TDR}$ estimator with either truncation rate $\tau_t = t^\alpha$ or $\tau_t = T^\alpha$
admits the following $\mse$ rate:
\begin{equation}
\mathsf{MSE}\Big(\mytdrdisavg \Big)\lesssim 
\begin{cases}
\kappa_T^2\xi_T^2 \vee   T^{-\frac{2\delta}{1+\delta}}\,, & \text{ if $0 < \delta < 1$ and $\alpha = 1/(1 + \delta)$ }\,, \\
\kappa_T^2\xi_T^2 \vee   \frac{\log T}{T}\,, & \text{ if $\delta = 1$ and $\alpha = 1/2$ }\,, \\
\kappa_T^2\xi_T^2 \vee   \frac{1}{T}\,, & \text{ if $\delta > 1$ and {$\alpha \ge {1}/{2}$} }.
\end{cases}
\end{equation}
\end{thm}
\proof

The following two results control the bias and variance of the $\tdr$ estimator.

\begin{lemma}\label{lemma: dis-tdr-var}
Under the conditions of Theorem \ref{thm: mse}, the $\tdr$ estimator with
truncation rate $\tau_t = t^\alpha$ or $\tau_t = T^\alpha$ satisfies
\begin{equation*}
\var(\mytdrdisavg)\lesssim  
\begin{cases}
\frac{1}{T^2}\sum\limits_{t=1}^T\tau_{t}^{1-\delta}+\frac{1}{T}\tau_T^{1-\delta} \,,& \delta<1 \,, \\
\frac{1}{T^2}\sum\limits_{t=1}^{T}\log(\tau_t)+ \frac{1}{T}\log \tau_T\,,& \delta=1\,,\\
\frac{1}{T}\,,& \delta>1\,.
\end{cases}
\end{equation*}
\end{lemma}
The proof of Lemma \ref{lemma: dis-tdr-var} can be seen in Section \ref{proof: lemma: dis-tdr-var}.
\begin{lemma}\label{lemma: dis-tdr-bias}
The following upper bound holds for the bias term for $\delta>0$:
\begin{equation*}
\Big|\E[\rhot_e]-\mydisavg\Big|^2 \lesssim
\begin{cases}
\kappa_T^2\xi_T^2  + \frac{1}{T^2} \left(\sum\limits_{t=1}^T \frac{1}{\tau_t^{\delta}}\right)^2\,, & \delta \le 1\,,\\
\kappa_T^2\xi_T^2  + \frac{1}{T^2} \left(\sum\limits_{t=1}^T \frac{1}{\tau_t}\right)^2\,, & \delta>1\,.
\end{cases}
\end{equation*}
\end{lemma}
The proof of Lemma \ref{lemma: dis-tdr-bias} is provided in Section \ref{proof: lemma: dis-tdr-bias}.

By writing the bias-variance decomposition we get
\begin{equation}\label{eq: tdr-mse}
\mse\Big(\mytdrdisavg\Big)=\E[(\mytdrdisavg-\mydisavg)^2]=\var(\mytdrdisavg)+\Big(\mydisavg-\E\big[\mytdrdisavg\big]\Big)^2\,.
\end{equation}

By combining Lemmas \ref{lemma: dis-tdr-var} and \ref{lemma: dis-tdr-bias} we obtain
\begin{equation*}
\mathsf{MSE}\Big(\rhot_e \Big)\lesssim
\begin{cases}
 \max\left\{ \kappa_T^2\xi_T^2, \frac{1}{T^2}\Big(\sum\limits_{t=1}^{T}\tau_t^{-\delta}\Big)^2  ,\frac{1}{T^2}\sum\limits_{t=1}^{T}\tau_t^{1-\delta},\frac{\tau_{T}^{1-\delta} }{T} \right\}\,, & \delta<1\,, \\
 \max\left\{ \kappa_T^2\xi_T^2, \frac{1}{T^2}\Big(\sum\limits_{t=1}^{T}\tau_t^{-1}\Big)^2  ,\frac{1}{T^2}\sum\limits_{t=1}^{T}\log \tau_t,\frac{1}{T}\log(\tau_T) \right\}\,,  & \delta=1\,, \\ 
 \max\left\{ \kappa_T^2\xi_T^2,
 \frac{1}{T^2}\Big(\sum\limits_{t=1}^{T}\tau_t^{-1}\Big)^2,\frac{1}{T}, 
 \right\}\,,  & \delta>1\,.
\end{cases}
\end{equation*}
To strike a balance between the bias and variance term, \eqref{eq: tdr-mse} reads as for $\delta<1$, the truncation rate can be chosen as either $\tau_t=t^{\frac{1}{1+\delta}}$ or $\tau_t=T^{\frac{1}{1+\delta}}$. Both options result in an $\mse$ rate of $T^{-\frac{2\delta}{1+\delta}}\vee  \kappa_T^2\xi_T^2$. For the case of $\delta=1$, the truncation policies $\tau_t=\sqrt{t}$ or $\tau_t=\sqrt{T}$ can be applied, leading to an $\mse$ rate of $\frac{\log T}{T} \vee   \kappa_T^2\xi_T^2$.

For the case of $\delta>1$, we note that the variance is upper bounded by the leading term $\frac{1}{T}$. For the bias, by using $\tau_t=t^\alpha$, it will be upper bounded by $\kappa_T^2\xi_T^2 +T^{-2 \alpha}$, given that $\alpha\ge \frac{1}{2}$, we get that the bias term is upper bounded by $\kappa_T^2\xi_T^2 +\frac{1}{T}$, this completes the proof.  
\qed

The above result gives our first encouraging finding about $\tdr$. When $\delta > 1$,
i.e., when the density ratio is square integrable, our estimator recovers the rate
of convergence of the $\dr$ estimator previously only established under strong distributional
overlap. When $\delta \leq 1$, $\tdr$ cannot achieve a $1/T$ rate of convergence using; however,
a good choice of $\alpha$ can still ensure a reasonably fast (polynomial) rate of convergence.

The proof of Theorem \ref{thm: mse} also shows that, when $\delta > 1$, the variance of the
$\tdr$ estimator dominates its bias. The following result goes further and establishes a
central limit theorem. We emphasize that the asymptotic variance $\sigma_b^2$ given in
\eqref{eq: sigma_b} matches the efficiency bound of \citet{kallus2022efficiently}.

\begin{thm}
\label{thm: truncated}
Suppose that Assumptions \ref{assu:MDP}-\ref{assu:stat} hold,
that $\gamma$-discounted $\delta$-weak distributional overlap holds for some $\delta>1$, 
and that we have an estimate $\homg^{\gamma}(.; \, p_0)$ that also satisfies
\[
\prob_{p_b}\left(\homg^{\gamma}(S; \, p_0)^{1+\delta}\ge x\right)\le \frac{C'}{x}\,, \quad \text{for all } x>0
\]
and some $C' < \infty$. In addition, suppose that  estimates $\hq_e$ and $\homg^{\gamma}(.; \, p_0)$ satisfy
Assumption \ref{assumption: model-estimates} with $\kappa_T \xi_T=o_T\big(1/\sqrt{T}\big)$, that
\begin{equation}\label{eq: mu}
\E_{p_b}\left[\|\hq_e-q^{\gamma}_e\|_2^{\frac{2(1+\delta)}{\delta-1}}\right]^{\frac{\delta-1}{1+\delta}}\le \mu_T^2\,,
\end{equation}
and that $\kappa_T\vee \xi_T \vee \mu_T=o_T(1)$.
Then, the $\tdr$ estimator with truncation rate $\tau_t=t^{\frac{\alpha}{2}}$ for some
$\alpha$ in the range $\frac{2}{1+\delta}< \alpha <\frac{1+\delta}{2}$ satisfies
\begin{equation}
\label{eq: sigma_b}
\begin{split}
&\sqrt{T}(\mytdrdisavg-\mydisavg) \overset{(d)}{\to} \normal(0,\sigma_b^2), \\
&\sigma_b^2=\E\left[\omega^\gamma (S; \, p_0)^2\frac{\pi_e(A|S)^2}{\pi_b(A|S)^2}\Big(R+\gamma v^{\gamma}_e(S') -q^{\gamma}_e(S,A)\Big)^2\right],
\end{split}
\end{equation}
where $(S,A,R,S')$ denotes a state-action-reward-state tuple generated under the behavior policy.
\end{thm}
\proof
We start with the following approximation result:

\begin{lemma}\label{lemma: martingale-o-one}
Under the conditions of Theorem \ref{thm: truncated}, we have
\begin{equation}
\begin{split}
&\sqrt{T}\left(\mytdrdisavg-\mydisavg\right)=\sum_{t=0}^{T-1} X_{t,T} +o_P(1), \\
&X_{t,T}\overset{\Delta}{=}\frac{1}{\sqrt{T}}\left({\omega}^\gamma(S_t; \, p_0)\land \tau_t\right) \frac{\pi_e(A_t \, | \, S_t)}{\pi_b(A_t \, | \, S_t)} \left(R_t + \gamma {v}_e^\gamma(S_{t+1}) - {q}_e^\gamma(S_t, \, A_t) \right)\,.
\end{split}
\end{equation}
\end{lemma}

Now, by Lemma \ref{lemma: bell-q}, we immediately see that the $X_{t, T}$
form a martingale difference sequence. The following lemmas further verify
a Lindeberg-type tail bound and a predictable variance condition for this
sequence.

\begin{lemma}\label{lemma: Lindeberg}
Under the conditions of Theorem \ref{thm: truncated},
$G_{t,T}=\sum_{\ell\le t }X_{\ell,T}$ is a squared-integrable triangular martingale array with respect to filteration $\cF_t=\{(S_\ell,A_\ell,R_\ell)_{\ell\le t+1}\}$. In addition, it satisfies Lindeberg's condition, where for every positive $\eps$ we have
\begin{equation*}
\sum_{t=1}^{T}\E[X_{t,T}^2\ind(|X_{t,T}|>\eps)|\cF_{t-1}]\overset{(p)}{\to} 0.
\end{equation*}
\end{lemma}

\begin{lemma}\label{lemma: var-convergence}
Under the conditions of Theorem \ref{thm: truncated}
and writing $\sigma_b^2$ as in \eqref{eq: sigma_b}, we have
\[
\sum_{t=1}^{T} X_{t,T}^2\overset{(p)}{\to} \sigma_b^2\,.
\]
\end{lemma}

The claimed result then follows from the martingale central limit theorem, e.g.,
Corollary 3.1 of \citet{hall2014martingale}
\qed

\subsection{Guarantees for long-run average reward estimation}
\label{sec:long_term}

We next proceed to focus on off-policy evaluation for the average reward value $\myavg$ when the weak distributional overlap assumption holds for the density ratio function $\omega_e(s)$.  In this case, we rely on a stronger mixing assumptions known as \textit{geometric mixing}, which ensures that the average value $\myavg$ is finite. This assumption has frequently been used in reinforcement learning to, e.g., study methods for differential value functions and temporal difference learning \citep[][Chapter 7]{van1998learning}; more closely related to us, \citet{liao2022batch} and \citet{hu2023off} use this assumption in the context of long-run average reward estimation.

\begin{assumption}[geometric mixing]\label{assumption: exp-mix}
For policy  $\pi\in \{\pi_e,\pi_b\}$, with $P^{\pi}$ being its corresponding state transition dynamics, we suppose that the MDP process is mixing with parameter $\exp(-1/t_0)$ for some $t_0>0$. More precisely, for $f_1,f_2$ as two distributions on MDP states, the following holds
\[
d_{\tv} \left(f_1P^{\pi}, f_2P^{\pi}\right) \le \exp\Big(-\frac{1}{t_0}\Big)d_{\tv}(f_1, f_2 )\,,
\]
where $fP^{\pi}$ stands for the state distribution after one step of MDP dynamics (under policy $\pi$) starting from the distribution $f$.
\end{assumption}

We would like to highlight that in the above definition, the state transition dynamics $P^{\pi}$ for two states $s,s'$ is given by
\[
P^{\pi}(s'|s)=\sum\limits_{a}^{} p_s(s'|s,a)\pi(a|s)\,.
\]
In the next proposition, we show that the differential value functions $V_e$, and $Q_e$ are uniformly bounded.   

\begin{proposition}\label{prop-unif-bound}
Under Assumptions \ref{assu:MDP}-\ref{assu:stat}, and the geometric mixing Assumption \ref{assumption: exp-mix} the differential value function $Q_e$ given in \eqref{eq: Q-V} is uniformly bounded, i.e., there exists a finite value constant $C_Q$ such that almost surely $Q_e(s,a)\le C_Q$ for all $s \in \cS$, and $a\in \cA$. 
\end{proposition}

By noting that $V_e(S)=\E_{\pi_e}[Q_e(S,A)|S]$, the above upper bound must work for $V_e(.)$ as well. An immediate consequence of this result implies that $\myavg\le C_Q$, given that $\myavg=\E_{p_e}[V_e(S)]$. 

\begin{assumption}\label{assumption: model-estimates-gamma-1}
For estimates $\hQ_e, \homg$, we suppose that there exists parameters $K_T$ and $\Xi_T$ such that 
\begin{align}
\E_{p_b\otimes \pi_b}[\|\hQ_e-Q_e\|_2^2] &\le K_T^2 \,, \nonumber  \\
\E_{p_b}\left[\|\homg-\omega\|_2^2\right] &\le \Xi_T^2\nonumber\,.
\end{align}
\end{assumption}

In the next theorem, we characterize the convergence rate for estimating $\myavg$ by $\tdr$ under $\delta$-weak distributional overlap.  
\begin{thm}\label{thm: mse-gamma-1}
Consider the off-policy evaluation for estimating the long-run average reward $\myavg$ under Assumptions \ref{assu:MDP}-\ref{assu:stat} and the long-run $\delta$-weak distributional overlap condition.  In addition, let model estimates $\hQ_e,\homg$ satisfy Assumption \ref{assumption: model-estimates-gamma-1} with parameters $K_T$ and $\Xi_T$. Then, the $\tdr$ estimator with either truncation rates $\tau_t=T^{\alpha}$ or $\tau_t=t^{\alpha}$, has the following convergence rate in the sense of mean-squared error:
{
\begin{equation}
\mathsf{MSE}\Big(\mytdravg \Big)\lesssim 
\begin{cases}
K_T^2\Xi_T^2 \vee   T^{-\frac{2\delta}{1+\delta}}\,, & \text{ if $0 < \delta < 1$ and $\alpha = 1/(1 + \delta)$ }\,, \\
K_T^2\Xi_T^2 \vee   \frac{\log T}{T}\,, & \text{ if $\delta = 1$ and $\alpha = 1/2$ }\,, \\
K_T^2\Xi_T^2 \vee   \frac{1}{T}\,, & \text{ if $\delta > 1$ and {$\alpha \ge {1}/{2}$} }.
\end{cases}
\end{equation}

}

\end{thm}

We refer to Section \ref{proof: thm: mse-gamma-1} for the Proof of Theorem \ref{thm: mse-gamma-1}.

In the next theorem, we establish CLT result for $\tdr$ with $1/\sqrt{T}$ rate of convergence. The limiting variance $\Sigma_b$ again matches the efficiency bound for this setting \citep{liao2022batch}.

\begin{thm}
\label{thm: truncated-gamma-1-clt}
Suppose that Assumptions \ref{assu:MDP}-\ref{assu:stat} hold, that long-run $\delta$-weak distributional overlap holds for some $\delta>1$, and that we have an estimate $\homg(.)$ that also satisfies
\begin{equation*}
\prob_{p_b}\left(\homg(S)^{1+\delta}\ge x\right)\le \frac{C'}{x}\,, \quad \text{for all } x>0\,,
\end{equation*}
for some positive finite-value $C'$. Consider $\tdr$ estimator with truncation rates $\tau_t=t^{\frac{\alpha}{2}}$ for some $\alpha$ in the range $\frac{2}{1+\delta}< \alpha <\frac{1+\delta}{2}$. In addition, suppose that model estimates $\hQ_e$ and $\homg$ satisfy Assumption \ref{assumption: model-estimates-gamma-1} with $K_T \Xi_T =o_T\big(1/\sqrt{T}\big)$, that
\begin{equation}\label{eq: M}
\E_{p_b\otimes \pi_b}\left[\|\hQ_e-Q_e\|_2^{\frac{2(1+\delta)}{\delta-1}}\right]^{\frac{\delta-1}{1+\delta}}\le M_T^2\,,
\end{equation}
and $ \Xi_T\vee M_T=o_T(1)$. 
We then have
\begin{align}
&\sqrt{T}\left(\mytdravg-\myavg\right) \overset{(d)}{\to} \normal(0,\Sigma_b^2)\,,\nonumber\\
&\Sigma_b^2=\E\left[\omega(S)^2\frac{\pi_e(A|S)^2}{\pi_b(A|S)^2}\Big(R+V_e(S)-Q_e(S,A)-\th\Big)^2\right]\label{eq: Sigma-b-2}\,,
\end{align}
where $(S,A,R,S')$ denotes a state-action-reward-state tuple generated under the behavior policy. 

\end{thm}

We refer to Section \ref{proof: thm: truncated-gamma-1-clt} for the proof of Theorem \ref{thm: truncated-gamma-1-clt}.


\section{Mixing-implied distributional overlap}

In the previous sections, we have found that the $\tdr$ estimator can achieve a
$T^{-\frac{2\delta}{\delta+1}}$ rate of convergence for off-policy evaluation in
MDPs under $\delta$-weak distributional overlap with $0<\delta < 1$. Two questions
left open by this result, however, are
\begin{enumerate}
\item When should we expect $\delta$-weak distributional overlap to hold; and
\item Can the rate of convergence of the $\tdr$ estimator be improved?
\end{enumerate}
To shed light on these questions, we here show that $\delta$-weak distributional
overlap can in some cases be derived from mixing assumptions; and the resulting
rates of convergence for $\tdr$ match the minimax rate of convergence under these
mixing assumptions.

The following result shows that under assumptions made in Section \ref{sec:long_term},
i.e., policy overlap (Assumption \ref{assu: policy}) and geometric mixing
(Assumption \ref{assumption: exp-mix}), $\delta$-weak distributional overlap
automatically holds---and so Theorems \ref{thm: mse-gamma-1} and \ref{thm: truncated-gamma-1-clt}
can directly be applied to characterize the MSE of the $\tdr$ estimator. The proof
of Theorem \ref{thm: mse-general-MDP} is given at the end of this section.

\begin{thm}\label{thm: mse-general-MDP}
 Consider an MDP under Assumptions \ref{assu:MDP}, \ref{assu:random}, \ref{assu:stat}, and geometric mixing Assumption \ref{assumption: exp-mix} with parameter $t_0$. In addition, we suppose that policy overlap Assumption \ref{assu: policy} holds with parameter $C_\eta=\exp(\zeta_\pi)$.  In this setting, the long-run $\delta$-weak distributional overlap is satisfied with $\delta={1}/{(\zeta_{\pi} t_0)}$.  
\end{thm}

\begin{coro}\label{coro:mix-tdr}
Consider MDP off-policy evaluation problem under Assumption \ref{assu:MDP}-\ref{assu:stat}, and geometric mixing Assumption \ref{assumption: exp-mix} with parameter $t_0$. In addition, we suppose that the policy overlap Assumption \ref{assu: policy} holds with parameter $C_\eta=\exp(\zeta_\pi)$. Depending on the magnitude of $\zeta_\pi t_0$, we  consider the following three settings:
\begin{enumerate}
\item For $\zeta_{\pi} t_0 < 1$, consider model estimates $\hQ_e,\homg$ satisfying Assumption \ref{assumption: model-estimates-gamma-1} with parameters $K_T, \Xi_T$. If we run $\tdr$ with either truncation rates $\tau_t = t^{\frac{\alpha}{2}}$ or $\tau_t = T^{\frac{\alpha}{2}}$, for some $\alpha \geq 1$, then the following mean-squared error rate holds:
        \[
        \mse(\mytdravg) \lesssim K_T^2 \Xi_T^2 \vee \frac{1}{T}.
        \]

If furthermore  $K_T \Xi_T = o_T(1/\sqrt{T})$ and for the parameter $M_T$ given in \eqref{eq: M} we have $\Xi_T \vee M_T = o_T(1)$, and we run $\tdr$ with truncation rate $\tau_t = t^{\frac{\alpha}{2}}$ and
        \[
        \frac{2\zeta_{\pi} t_0}{1 + t_0 \zeta_{\pi}} < \alpha < \frac{1 + t_0 \zeta_{\pi}}{2\zeta_{\pi} t_0},
        \]  
then the following Central Limit Theorem (CLT) convergence holds:
        \[
        \sqrt{T}\left(\mytdravg - \myavg\right) \overset{(d)}{\to} \normal(0, \Sigma_b^2),
        \]
        where $\Sigma_b^2$ is given in \eqref{eq: Sigma-b-2}.

\item For $\zeta_{\pi} t_0 =1$, consider model estimates $\hQ_e,\homg$ satisfying Assumption \ref{assumption: model-estimates-gamma-1} with parameters $K_T,$ and $\Xi_T$. We run $\tdr$ with either truncation rates $t^{\frac{1}{2}}$, or  $T^{\frac{1}{2}}$, then the following $\mse$ rate holds for the $\tdr$ estimator:
\[
\mse(\mytdravg)\lesssim K_T^2 \Xi_T^2 \vee \frac{\log T}{T}\,.
\]

\item For $\zeta_{\pi} t_0 >1$, consider model estimates $\hQ_e,\homg$ satisfying Assumption \ref{assumption: model-estimates-gamma-1} with parameters $K_T,$ and $\Xi_T$. We run $\tdr$ with either truncation rates  $\tau_t=t^{\frac{\zeta_{\pi} t_0}{1+\zeta_{\pi} t_0}} $ or $\tau_t=T^{\frac{\zeta_{\pi} t_0}{1+\zeta_{\pi} t_0}}$, then the following $\mse$ convergence rate holds:
\[
\mse(\mytdravg)\lesssim K_T^2 \Xi_T^2 \vee T^{-\frac{2}{1+\zeta_{\pi} t_0}}\,.
\]  
\end{enumerate}
\end{coro}

The above result is already encouraging, as it shows that guarantees for $\tdr$
can be derived by mixing and policy-overlap assumptions alone, without having
to ever explicitly make assumptions about the tail-decay rate of $\omega(\cdot)$.

Furthermore, it turns out that the polynomial exponent in the rate of convergence
is optimal in a minimax sense. To see this, we draw from recent work by \citet[Section 3.2]{hu2023off},
who establish the following minimax-error bound for MDPs under mixing
assumptions. A comparison of the rates in Corollary \ref{coro:mix-tdr} and
Theorem \ref{thm:lb} immediately implies that the $\tdr$ attains the minimax
rate of convergence when $\delta \neq 1$ (and is within a log-factor of minimax
when $\delta = 1$).

\begin{thm}[\citet{hu2023off}]
\label{thm:lb}
There exists a set of MDP off-policy evaluation tasks satisfying
Assumptions \ref{assu:MDP}--\ref{assu:stat} and \ref{assumption: exp-mix}
with policy overlap parameter  $C_\eta = \exp(\zeta_{\pi})$ and mixing parameter $t_0$
for which no estimator can achieve a rate of convergence faster than
\smash{$T^{-{2}/{(1+\zeta_{\pi} t_0)}}\vee T^{-1}$} in mean-squared error.
\end{thm}

Interestingly, \citet{hu2023off} proved this lower bound for a much smaller
class of MDPs, namely strongly regenerative MDPs that return to a ``reset''
state with probability uniformly bounded from below by $1 - e^{-1/t_0}$
(this strongly regenerative assumption implies Assumption \ref{assumption: exp-mix},
and so their lower bound directly also applies to our setting).
They also proved that this bound was attained for strongly regenerative MDPs
using a simple inverse-propensity weighted estimator that also resets itself
each time the MDP returns to its reset state. Given this context, our results
imply that, as long as we can estimate $\hQ_e$ and $\homg$ at reasonably
fast rates, off-policy estimation in MDPs under geometric mixing is
no harder than off-policy evaluation in strongly regenerative MDPs---and optimal
rates can be achieved using the $\tdr$ estimator.

\subsection{Proof of Theorem \ref{thm: mse-general-MDP}}

Let $p$ and $q$ denote stationary distributions under the behavior and evaluation policies $\pi_b$ and $\pi_e$, respectively.  We want to find the largest value $\delta$ such that the long-run $\delta$-weak distributional overlap assumption holds for the density ratio function $\omega(s)=\frac{q(s)}{p(s)}$. In particular, we must find the largest positive $\delta$ such that there exists a positive $C$ value for which the following holds:
 \begin{align*}
 \prob_{p}\left(\omega(S)^{1+\delta} \ge x \right)\le \frac{C}{x}\,, \quad \forall x>0\,.
 \end{align*}
By rewriting the the above tail inequality we arrive at
 \begin{align}
 \prob_p(\omega(S)^{1+\delta} \ge x)&=\int \ind\left(\omega(s)^{1+\delta}\ge x\right) p(s) \de \lambda_s(s) \nonumber\\
 &=\int \frac{\ind\left(\omega(s)^{1+\delta}\ge x\right)}{\omega(s)} q(s) \de \lambda_s(s)\nonumber\\
 &\le x^{-\frac{1}{1+\delta}} \int \ind\left(\omega(s)^{1+\delta}\ge x\right) q(s) \de\lambda_s(s)\nonumber\\
 &= x^{-\frac{1}{1+\delta}} \prob_q\left(\omega(S)^{1+\delta} \ge x\right)\,,\label{eq: mdp-tmp1}
 \end{align}
 where in the penultimate relation we used the inequality $\frac{\ind(x\ge t)}{x} \le \frac{\ind(x\ge t)}{t}$ for positive $x$ and $t$. We next try to connect $\prob_q\left(\omega(S)^{1+\delta} \ge x\right)$  to $\prob_p\left(\omega(S)^{1+\delta} \ge x\right)$, which enables us to employ \eqref{eq: mdp-tmp1} and establish an upper bound tails of $\omega(s)$ under the behavior policy. 
For this end, we try to use another distribution which we know is in $\tv$ distance close to $q$. Specifically, we consider distribution $q_k$ which is the distribution on states when we start from $p$ and proceed $k$ steps under the evaluation policy $\pi_e$. From the geometric mixing assumption, we know that $q_k$ must approach to $q$ as $k$ grows to infinity. More precisely, by using Assumption \ref{assumption: exp-mix}, we obtain 
\begin{equation}\label{eq: tmp-tv}
d_{\tv}(q_k,q)\leq \exp(-k/t_0)\,,
\end{equation}
where this holds due to the fact that $q$ is the stationary distribution on states under evaluation policy.  We next consider the dataset $\{S'_1,A'_1,S'_2,A'_2,\dots,S'_{k},A'_k,S'_{k+1}\}$ which is collected under the evaluation policy $\pi_e$ starting from $S'_1\sim p$. Given that $q$ is the stationary distribution under the evaluation policy, we expect that the state distribution of $S'_\ell$ gets closer to $q$ as $\ell$ grows. We also consider another dataset collected under behavior policy. Specifically, we consider 
$\{S_1,A_1,S_2,A_2,\dots,S_{k},A_k,S_{k+1}\}$ which starts from $S_1\sim p$ and proceeds under the behavior policy $\pi_b$, thereby the state distributions will not change, as the process is invariant under the stationary distribution $p$. Let $p^{(k+1)}_{\mathbf{S},\mathbf{A}}$ denote the joint density function of $(S_1,A_1,S_2,A_2,\dots,S_{k},A_k,S_{k+1})$. More precisely, we have
\[
p^{(k+1)}_{\mathbf{S},\mathbf{A}}(s_1,a_1,\dots,s_k,a_k,s_{k+1})=p(s_1)\pi_b(a_1|s_1)p_s(s_2|a_1,s_1)\dots p_s(s_{k}|s_{k-1},a_{k-1})\pi_b(a_k|s_k)p_s(s_{k+1}|s_k,a_k)\,.
\]
Similarly, we let$p^{(k+1)}_{\mathbf{S}',\mathbf{A}'}$ denote the distribution of $(S'_1,A'_1,S'_2,A'_2,\dots,S'_{k},A'_k,S'_{k+1})$, this implies that
\[
p^{(k+1)}_{\mathbf{S}',\mathbf{A}'}(s_1,a_1,\dots,s_k,a_k,s_{k+1})=p(s_1)\pi_e(a_1|s_1)p_s(s_2|a_1,s_1)\dots p_s(s_{k}|s_{k-1},a_{k-1})\pi_e(a_k|s_k)p_s(s_{k+1}|s_k,a_k)\,.
\]

In the next step, for an optional measurable function $g:\cS\to \reals$ we have
\begin{align}\label{eq: tmp-q_k}
\E_{q_k}[g(S)]&=\E[g(S'_{k+1})]\nonumber\\
&=\int g(s_{k+1})p_{\mathbf{S}',\mathbf{A}'}(a_1,s_1,\dots,s_k,a_k,s_{k+1})\de \lambda_s(s_1)\de \lambda_a(a_1)\dots\de \lambda_s(s_{k})\de \lambda_a(a_k)\de \lambda_s(s_{k+1})\nonumber \\
&=\int g(s_{k+1})p_{\mathbf{S},\mathbf{A}}(a_1,s_1,\dots,s_k,a_k,s_{k+1})\prod\limits_{i=1}^{k} \frac{\pi_e(a_i |s_i)}{\pi_b(a_i|s_i)}~ \de \lambda_s(s_1)\de \lambda_a(a_1)\dots\de \lambda_s(s_{k})\de \lambda_a(a_k)\de \lambda_s(s_{k+1}) \nonumber  \\
&=\E\left[g(S_{k+1})^{} \prod\limits_{i=1}^{k} \frac{\pi_e(A_i |S_i)}{\pi_b(A_i|S_i)} \right]\nonumber \\
&\leq \exp(\zeta_\pi k) \E_{p}\left[g(S)^{}\right]\,,
 \end{align}
where the last inequality follows policy overlap Assumption \ref{assu: policy}, and the fact that induced state distribution on $S_{k+1}$ has density function $p$, as stationary distribution $p$ is invariant under the behavior policy. 
In addition, for every positive $x$ value, by using the total variation distance definition we have
 \[
   \prob_q\left(\omega(S)^{1+\delta}\ge x\right) \le d_{\tv}(q,q_k)+\prob_{q_k}\left(\omega(S)^{\delta}\ge x\right)\,.
 \]
 Using \eqref{eq: tmp-tv} in the above yields
  \begin{align*}
  \prob_q(\omega(S)^{1+\delta}\geq x)& \le \exp(-k/t_0)+\prob_{q_k}\left(\omega(S)^{1+\delta}\geq x\right)\nonumber \,.
  \end{align*}
Plugging this in \eqref{eq: mdp-tmp1} gives us
 \begin{align*}
  \prob_p(\omega(S)^{1+\delta} \ge x) &\le  x^{-\frac{1}{1+\delta}} \left(  \exp(-k/t_0)+\prob_{q_k}(\omega(S)^{1+\delta}\geq x) \right) \\
  &= x^{-\frac{1}{1+\delta}} \left(  \exp(-k/t_0)+\E_{q_k}\Big[\ind\big(\omega(S)^{1+\delta}\ge x\big)\Big] \right)\\
  &\le x^{-\frac{1}{1+\delta}} \left(  \exp(-k/t_0)+\exp(k \zeta_\pi)\E_{p}\Big[\ind\big(\omega(S)^{1+\delta}\ge x\big)\Big] \right)\,,
 \end{align*}
 where the last relation follows \eqref{eq: tmp-q_k} for the function $g(s)=\ind(\omega(s)^{1+\delta}\geq x)$. This can be written as
 \begin{align*}
 \prob_p\left(\omega(S)^{1+\delta} \ge x\right) \le x^{-\frac{1}{1+\delta}} \left(  \exp(-k/t_0)+\exp(k \zeta_\pi)\prob_p\Big(\omega(S)^{1+\delta}\ge x\Big) \right)\,.
 \end{align*}
 In the next step, using the above relation when $x^{1+\delta} \ge \exp(k \zeta_\pi)$ it is easy to arrive at the following
\begin{align}\label{eq: mdp-tmp2}
 \prob_p\left(\omega(S)^{1+\delta} \ge x\right) \le  \frac{ \exp(-k/t_0) } {x^{\frac{1}{1+\delta}} -\exp(k\zeta_\pi) }\,.
 \end{align}
 Since this relation holds for all positive integer values of $k$ that $x^{1+\delta} \ge \exp(k \zeta_\pi)$, and we want to establish a small upper bound for $ \prob\left(\omega(s)^{1+\delta} \ge x\right)$, we can take the infimum for all positive integers. In particular, we have
 \begin{align*}
 \prob_p\left(\omega(S)^{1+\delta} \ge x\right) \le\inf\limits_{\substack{k\in \mathbb{Z}^+, \\ k(1+\delta)\zeta_\pi \le \log (x)}}^{}
\frac{ \exp(-k/t_0) } {x^{\frac{1}{1+\delta}} -\exp(k\zeta_\pi) }\,.
 \end{align*}
  Guided by the above optimization problem, we consider $k(x)=\lfloor k^* \rfloor$ for 
 \begin{equation}\label{eq: mdp-tmp3}
 k^*=\frac{1}{\zeta_\pi}\log \frac{x^{\frac{1}{1+\delta}}}{1+\zeta_\pi t_0}\,.
 \end{equation}
 To ensure that $k(x) \ge 1$, we focus on $x\ge x_0$ where 
 \begin{align}\label{eq: x-lower-start}
x_0=\Big(\exp(\zeta_\pi)(\zeta_{\pi} t_0+1)\Big)^{1+\delta}\,.
 \end{align}
 It is also obvious to see that $k(x)\zeta(1+\delta) \le \zeta_{\pi}$ and the condition in the above optimization problem is satisfied. We then plug $k(x)$ in \eqref{eq: mdp-tmp2} and use $k^*-1< k(x) \le k^*$ to get
\begin{align*}
 \prob_p\left(\omega(S)^{1+\delta} \ge x\right) &\le  \frac{ \exp(-k^*/t_0)\exp(1/t_0) } {x^{\frac{1}{1+\delta}} -\exp(k^*\zeta_\pi) }\\
 &=\frac{ \exp(1/t_0)  \Big(\frac{x^{\frac{1}{1+\delta}}}{\zeta_\pi t_0+1} \Big)^{-\frac{1}{\zeta_\pi t_0}} } {x^{\frac{1}{1+\delta}}\Big(1-\frac{1}{1+\zeta_\pi t_0} \Big)  }\,,
\end{align*}
where in the last relation we used \eqref{eq: mdp-tmp3}.
This yields
\[
 \prob_p\left(\omega(S)^{1+\delta} \ge x\right)  \le \frac{ \exp(1/t_0)  \Big(\frac{1}{\zeta_\pi t_0+1} \Big)^{-\frac{1}{\zeta_\pi t_0}} } {x^{\frac{1}{1+\delta}\Big(1+\frac{1}{\zeta_\pi t_0 } \Big ) }\Big(1-\frac{1}{1+\zeta_\pi t_0} \Big)  }\,.
\]
Putting everything together, by considering $\delta=\frac{1}{\zeta_\pi t_0}$ we get $\prob_p\left(\omega(S)^{1+\delta}\ge x\right)\le \frac{C'}{x}$, for all $x\ge x_0$ for $x_0$ given in \eqref{eq: x-lower-start}, with $\delta=\frac{1}{\zeta t_0}$, and the following value of $C'$:
\[
C'=\frac{ \exp(1/t_0)  \Big({\zeta_\pi t_0+1} \Big)^{1+\frac{1}{\zeta_\pi t_0}} }
 { \zeta_\pi t_0  }\,.
\] 
In summary, we get
\[
\prob_{p}\left(\omega(S)^{1+\frac{1}{\zeta_{\pi}t_0 }} \ge x\right) \le \frac{C'}{x}\,, \quad \forall x\ge \Big(\exp(\zeta_\pi)(\zeta_{\pi} t_0+1)\Big)^{1+\frac{1}{\zeta_{\pi} t_0}}\,.
\]  
By considering $C=\max\left\{\Big(\exp(\zeta_\pi)(\zeta t_0+1)\Big)^{1+\frac{1}{\zeta_\pi t_0}}, C' \right\}$ we finally arrive at
\[
\prob_{p}\left(\omega(S)^{1+\frac{1}{\zeta_{\pi}t_0 }} \ge x\right) \le \frac{C}{x}\,,\quad \forall x>0\,.
\]  
\qed

{
\section{Data-driven choice of truncation rates}

In the theorems discussed above, we theoretically characterized the optimal truncation rate that balances the variance and bias of the TDR estimator. Specifically, this optimal truncation depends on the mixing rate and the weak-distributional-overlap exponent of the decision process. Because these parameters can be computationally difficult to estimate (unless they are clear from context), in this section we provide a data-driven method to balance the variance and bias of the TDR estimator. To do so, we adopt Lepski’s method \citep{lepskii1992asymptotically}, a common approach for data-driven selection of regularization parameters in nonparametric statistics.  

For the off-policy evaluation problem, \citet{su2020adaptive} propose a Lepski-based procedure for stationary processes, and we adopt their method. At the core of this approach, for each estimate, a statistically valid confidence interval (for a fixed probability level) is constructed—we will describe this process in detail later. Then, estimators are sorted in ascending order of bias, so that $\mathsf{bias}(i) \le \mathsf{bias}(i+1)$. Finally, the estimator with the largest bias that \textit{still} overlaps with all previous confidence intervals is selected, i.e.,
\[
\hat{i} = \max\big\{ i : \cap_{j=1}^i \neq \emptyset \big\}\,.
\]
In our setup, to achieve an increasing bias order, we consider truncation rates in descending order—since higher truncation implies lower bias. To construct a confidence interval for each truncation rate, we use bootstrap resampling to estimate the mean and standard error, and then transform these values into confidence intervals.  

One challenge here is that the samples in each decision trajectory are not independent, and so the classic bootstrap resampling \citep{efron1994introduction} which discards the temporal order of samples cannot be used for inference. To address this, we use the block bootstrap procedure for stationary processes outlined in \cite{kunsch1989jackknife}. In this method, consecutive samples of length $\ell$ are grouped into blocks (preserving order), and bootstrap resampling is performed on these blocks. Concatenating the resampled blocks produces a new trajectory. \cite{kunsch1989jackknife} show that for stationary processes, if the block length for some positive $0<\alpha<1/2$ satisfies $\ell=O(T^{1/2-\alpha})$ (where $T$ is the total sample length), then the resulting estimator is unbiased and has convergence in probability to the ground-truth parameters. The resulting method for selecting the truncation level is summarized in Algorithm \ref{alg: lepski}.

\begin{algorithm}[t]
\caption{Data-driven truncation selection for $\tdr$}
\label{alg: lepski}
\begin{center}

{\begin{minipage}{0.98\linewidth}

\medskip
\textbf{Input:}
\begin{enumerate}
\item Stationary trajectory $\mathcal{D}=\{(S_t,A_t,R_t)\}_{t=1}^T$ under behavior policy $\pi_b$

\item Ordered truncation grid in descending order $\mathcal{T}=\{M_1,\ldots,M_G\}$-- \emph{low bias} to \emph{high bias} 
\item 
Block lengths $\ell$ such that $\ell=O(T^{1/2-\eps})$ for $0<\eps<1/2$ (e.g., $\ell=\lfloor T^{1/3}\rfloor$);
\item Number of bootstrap draws $B$ (e.g., $B=100$) and standard normal quantile $c$ (e.g., $c=1.96$ for $95$-th percentile).
\end{enumerate}
\textbf{Output:} selected truncation $\widehat{\tau}=\tau_{i^\star}$ 

\medskip
\begin{enumerate}
\item[\textbf{1.}] \textbf{Bootstrap CIs for each truncation.} For $g=1,\ldots,G$:
\begin{enumerate}
\item Generate $\{\mathcal{D}^{*(b)}\}_{b=1}^B$ by the moving block bootstrap with block length $\ell$:
form all length-$\ell$ overlapping blocks; draw $k=\lceil T/\ell\rceil$ blocks uniformly random with replacement; concatenate and truncate to length $T$.
\item Compute $\widehat{\mu}^{*(b)}_g \leftarrow \mathrm{TDR}(\mathcal{D}^{*(b)},M_g)$ for $b=1,\ldots,B$; (TDR with truncation $M_g$).
\item Let $\bar{\mu}_g=\frac{1}{B}\sum_{b=1}^B \widehat{\mu}^{*(b)}_g$ and
$\widehat{\sigma}_g=\Big(\frac{1}{B-1}\sum_{b=1}^B(\widehat{\mu}^{*(b)}_g-\bar{\mu}_g)^2\Big)^{1/2}$.
\item Define the confidence interval $\mathrm{CI}_g=[\,\bar{\mu}_g- c\,\widehat{\sigma}_g,\,\bar{\mu}_g+ c\,\widehat{\sigma}_g ]$.
\end{enumerate}

\item[\textbf{2.}] \textbf{Lepski's method selection.}
Initialize $\mathcal{I}\gets \mathrm{CI}_1$ and $i^\star\gets 1$.
For $g=2,\ldots,G$:
\[
\mathcal{I}\gets \mathcal{I}\cap \mathrm{CI}_g;\quad
\text{if }\mathcal{I}\neq\emptyset\text{ then } i^\star\gets g\ \text{ else break.}
\]
(Return $i^\star$ as the largest index whose CI still overlaps all previous ones.)

\item[\textbf{3.}] \textbf{Final chosen truncation.}
Set $\widehat{M}=M_{i^\star}$ 
\end{enumerate}
\end{minipage}}
\end{center}
\end{algorithm}

}

\section{Numerical experiments}
\label{sec:simu}
\subsection{Background}
In this section, we compare the empirical performance of the $\tdr$ estimator to that of the standard (untruncated) $\dr$ estimator  \citep{kallus2022efficiently,liao2022batch} for the off-policy evaluation problem. 
We estimate $Q$-functions required to form these estimators via temporal difference learning \citep{sutton2018reinforcement} using samples distinct from those for the later OPE problem.
We estimate the discounted $q$-function ($q^\gamma_e$) via an iterative learning
algorithm with learning rate $\nu$, and updates
\begin{equation}\label{eq: td}
\hq_e(s,a) \leftarrow \hq_e(s,a)+\nu \left(r+\gamma \big(\pi_e \hq_e(s',1)+ (1-\pi_e)\hq_e(s',0)\big)-\hq_e(s,a)\right),
\end{equation}
where the tuple $(s,a,r,s')$ denotes a state, action, and reward, together with the next state.
We estimate the long-run $Q$-function ($\hQ_e$)  via an iterative learning algorithm
learning rate parameters $\nu_1, \nu_2$ and updates
\begin{align}\label{eq: td-long-run}
Q_e(s,a)&\leftarrow Q_e(s,a)+\nu_1 \left(r-\hth+\pi_e Q_e(s',1)+ (1-\pi_e)Q_e(s',0)-Q_e(s,a)\right)\,, \nonumber\\
\hth&\leftarrow \hth+\nu_2 \left(r-\hth+\pi_e Q_e(s',1)+ (1-\pi_e)Q_e(s',0)-Q_e(s,a) \right),
\end{align}
{
where $(s,a,r,s')$ has the same meaning as above.
We estimate the stationary density ratio function $\omega^\gamma(s;, p_0)$ via empirical moment matching using a stationary version of the Bellman given in Lemma \ref{lemma: bell-prob} \citep{uehara2020minimax}. The estimator is given in Algorithm \ref{alg: omega-estimate}; see Appendix \ref{sec: omega-estimation} for further details.
}

\begin{algorithm}[t]
\caption{Density-ratio $\omega^{\gamma}(.;p_e)$ estimation by moment matching}
\label{alg: omega-estimate}
\begin{center}
\begin{minipage}{0.98\linewidth}
{
\medskip
\textbf{Input:}
\begin{enumerate}
\item Policies $\pi_b,\ \pi_e$ and finite state space $\cS=\{1,\dots,m\}$
\item Behavior trajectory $\mathcal{D}_b=\{(S_t,A_t,R_t)\}_{t=1}^{T}$ under $\pi_b$

\end{enumerate}
\textbf{Output:} $\hat\omega(j)\approx \omega^\gamma(j;p_e)=\frac{p_e(j)}{p_b(j)}$ for all $j\in[m]$

\medskip
\textbf{Steps:}
\begin{enumerate}
\item \textbf{Compute importance weights.} For $t=1,\dots,T-1$ set $\eta_t \leftarrow \frac{\pi_e(A_t\mid S_t)}{\pi_b(A_t\mid S_t)}.
$

\item \textbf{State visit counts.} For each $j\in[m]$ set $N_j \leftarrow \sum_{t=1}^{T-1}\ind(S_t=j)$.

\item \textbf{Weighted transition matrix.} For all $i,j\in[m]$ set  $M^{(T)}_{j,i} \leftarrow \sum_{t=1}^{T-1}\ind(S_t=i,\,S_{t+1}=j)\,\eta_t.$

\item \textbf{Build linear system.} Let $H \leftarrow \mathsf{diag}([N_1,\dots,N_m]) - M^{(T)}, 
\qquad
c \leftarrow \frac{1}{T-1}\big[N_1,\dots,N_m\big]^\sT.
$ 
\item \textbf{Estimate $\beta$.} Solve the constrained least squares
\[
\hat\beta \leftarrow \arg\min_{\beta\in\reals^m}\ \|H\beta\|_2^2
\quad \text{s.t.}\quad c^\sT \beta = 1\,,\quad \beta_i\ge 0\,,~~\forall i\in[m]\,.
\]

\item \textbf{Return.} Set $\hat\omega(j)\leftarrow \hat\beta_j$ for all $j\in[m]$.
\end{enumerate}
}
\end{minipage}
\end{center}
\end{algorithm}

We consider the following two data-generating designs, and use these two run 3 experiments.

\vspace{0.5\baselineskip}
\noindent\textbf{Setup 1.} We adopt the MDP setup outlined in \citet{hu2023off}, where a Markov process with $Q$ states and two actions, $a \in \{0,1\}$, is considered. In each state, selecting action $0$ leads to a transition back to state $1$ with probability one. Alternatively, being in state $i$ and choosing treatment $1$ results in transitioning to the state $(i+1) \wedge Q$ with probability $1-\beta$. However, with probability $\beta$, there is a return to state $1$ if treatment $1$ is taken. For the treatment probability $u$ (the probability of selecting action $1$), we consider $v=u(1-\beta)$ then the stationary distribution is given by
\begin{equation}\label{eq: stationary-ex-1}
p^{\mathsf{stationary}}(v)=(1-v) \left[1,v, v^2,\dots,v^{Q-2},\frac{v^{Q-1}}{1-v}\right]\,.
\end{equation}
In this setting, when the treatment likelihood is larger under the evaluation policy, the upper bound in strong distributional overlap grows larger, potentially leading to a large $\mse$ of $\dr$ estimator. Rewards are generated as follows:
\[
r(s)=10-\frac{5}{\sqrt{s}}+w\,,\quad w \sim \mathsf{Unif}\Big(\big[-\frac{1}{2},\frac{1}{2}\big]\Big)\,,\quad \forall s\in [Q]\,.
\]   

\noindent\textbf{Setup 2.} We consider a queuing system where $X_t$
represents the number of users in the queue, and we assume that the system maintains a constant processing rate of 1. At time $t$, a random amount $B_t$ of users joins the system, where the mean value of $B_t$ depends on the action taken in state $X_t$. More precisely, if action $a \in \{0,1\}$ is done, then $B_t$ follows a Poisson distribution with parameter $\lambda_a$. The state dynamics of this process, given that we are in state $X_t$ and action $a_t$ is taken, are represented as follows:

\[
X_{t+1}=\left(X_t-1+a_t B_t^{(1)}+(1-a_t) B_t^{(0)} \right)_+\,,\quad  B_t^{(0)}\sim \mathsf{Poiss}(\lambda_0)\,,\quad B_t^{(1)} \sim \mathsf{Poiss}(\lambda_1)\,.
\]
This stochastic process is MDP with the state space being non-negative integers. We consider the following reward system depending on the state $X_t$:
\[
R_t=10-\frac{5}{\sqrt{X_t+1}}+w_t\,, \quad w_t\sim \mathsf{Unif}\left(\Big[-\frac{1}{2},\frac{1}{2}\Big]\right)\,.
\]
We focus on a class of policies, where the treatment probability is independent of states and it is always equal to $q$.
\subsection{Experiments}
{Before presenting the experiments, we provide a brief summary of each to clarify their setup and characterization.
We consider two approaches to setting the threshold level. For the first set of experiments (1--3) we assume a-priori knowledge of the actual values of $\omega(.)$ in order to isolate the role of truncation in the $\tdr$ estimator. For these experiments, we use this information for two purposes: first, to form the doubly robust estimator; and second, to appropriately select the truncation level with $\tdr$. Specifically, we choose $\delta$ to be as large as possible while ensuring that \smash{$\E_{p_b}[\omega^{1+\delta}]$} remains sufficiently small, motivated by applying Markov’s inequality to control the constant term in the weak distributional overlap definition given in \eqref{eq: weak-exponent}. With this chosen $\delta$, and guided by Theorems \ref{thm: mse} and \ref{thm: mse-general-MDP}, we consider truncation levels of the form \smash{$t^{\frac{1}{1+\delta}}$} or \smash{$T^{\frac{1}{1+\delta}}$}. Then, in subsequent experiments, we use Lepski’s method to select the truncation rate to achieve end-to-end evaluation. The full set of experiments is summarized in Table \ref{tab:exp-summary-compact}.

\begin{table}[t]
\centering
\footnotesize
\setlength{\tabcolsep}{4pt}
\renewcommand{\arraystretch}{1.1}
\begin{tabular}{c c c c c}
\toprule
\textbf{Experiment} & \textbf{Setup} & \textbf{OPE type} & Distributional ratio \boldmath$\omega(.)$ & \textbf{Truncation} \\
\midrule
1 & 1 & Discounted & Known & Theory-guided \\
2 & 2 & Long-run average & Known & Theory-guided\\
3 & 2 & Discounted & Known & Theory-guided \\
4 & 1 & Discounted  & Known & {Lepski's} method \\
5 & 1 & Discounted & Estimated & {Lepski's} method\\
6 & 1 & Discounted  & Estimated & Broad grid (robustness study) \\
\bottomrule
\end{tabular}
\caption{Compact summary of experiments setups. “Known” implies that $\omega^{\gamma}(\cdot)$ is treated as oracle; “Estimated” uses the moment-matching procedure given in Algorithm \ref{alg: omega-estimate}.}
\label{tab:exp-summary-compact}
\end{table}

}
\vspace{0.5\baselineskip}
\noindent\textbf{Experiment 1.}
We adopt MDP Setup 1 (outlined above) for estimating the discounted average reward with $\gamma=0.5$ where the initial distribution $p_0=p_e$, with $p_e$ being the stationary distribution under the evaluation policy. Specifically, we consider an evaluation policy with a treatment probability of $u_e=1$, while the behavior policy adopts a treatment probability of $u_b=0.2$. Given that $u_e>u_b$, it becomes evident that the strong distributional overlap escalates towards very large values as we progress to states with higher labels. In this experiment, we consider the number of states $Q=20$, and $\beta=0.5$. We consider $\tdr$ estimator with two truncation rates $\tau_t=t^{0.7}$, and $\tau_t=T^{0.7}$.
 We use temporal-difference learning given in \eqref{eq: td} with the learning rate $\nu = 0.03$ and a trajectory length of $T = 10{,}000$. For the OPE problem, as collected data under the behavior policy, we consider a set of trajectory lengths $T\in \{50,600,7200,86400\}$. For each fixed $T$, we average results over 50000 experiments. We plot the log of $\mse$ to the log of $T$ for both $\tdr$ and $\dr$ estimators.

\begin{figure}[t]
  \centering
  \begin{minipage}{.48\textwidth}
    \centering
    \includegraphics[scale=0.39]{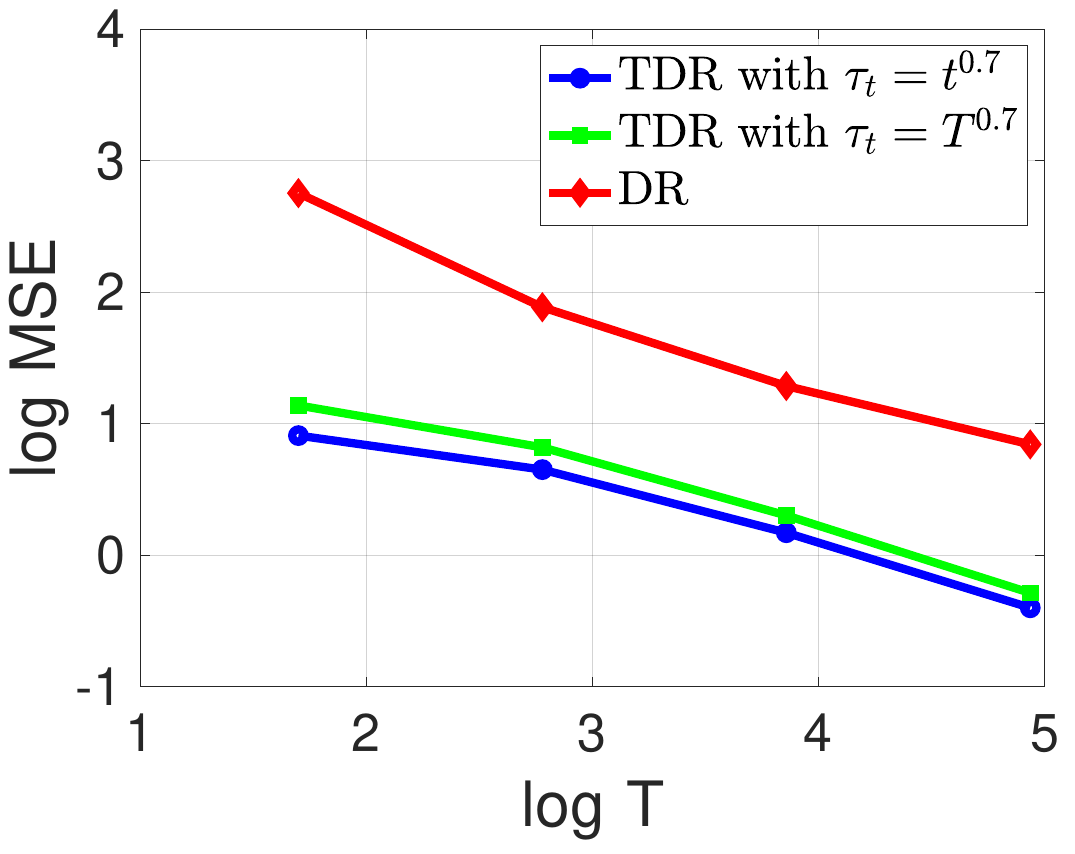}
    \caption{ The  $\log$ of $\mse$ to the $\log$ of trajectory length $T$ for $\tdr$ and $\dr$ estimators under setup outlined in Experiment 1.}
    \label{fig: tdr-1}
  \end{minipage}
   \hfill
     \begin{minipage}{.48\textwidth}
    \centering
    \includegraphics[scale=0.39]{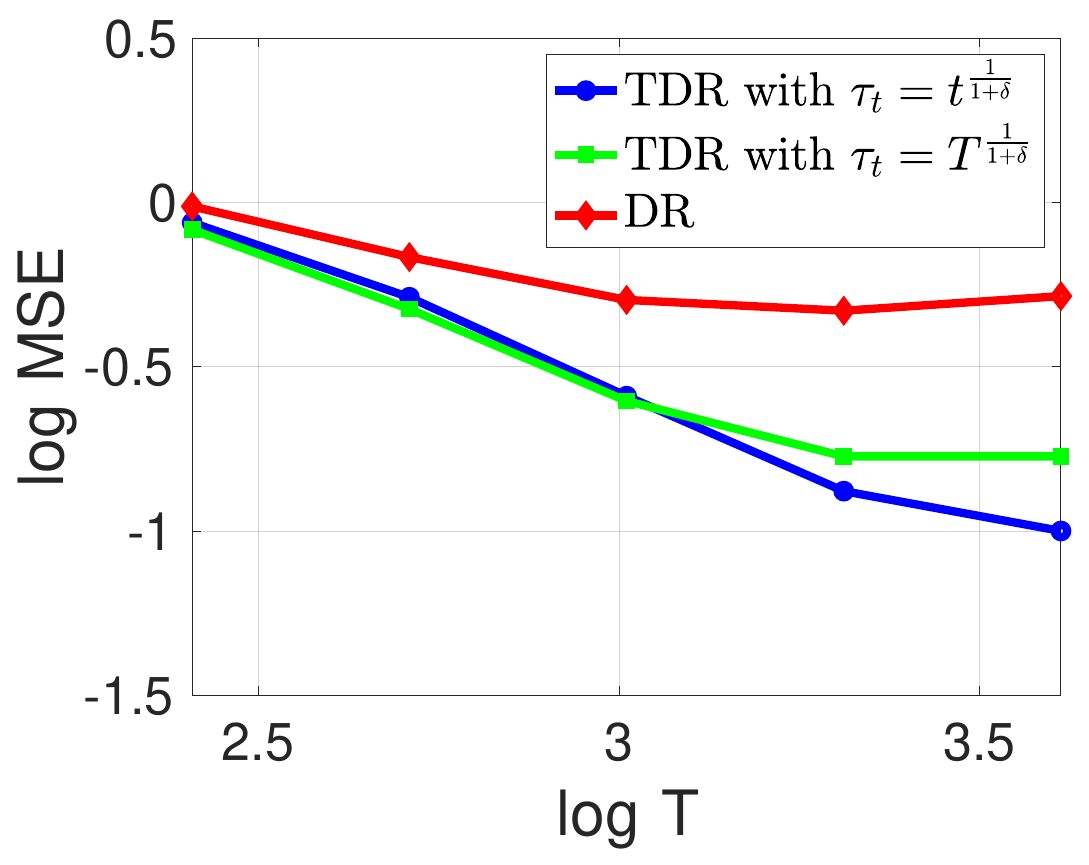}
    \caption{ The  $\log$ of $\mse$ to the $\log$ of trajectory length $T$ for $\tdr$ and $\dr$ under setup given in Experiment 2.}
    \label{fig: tdr-self-normalization}
  \end{minipage}

\end{figure}

The results are shown in Figure \ref{fig: tdr-1}. As the trajectory length increases, the mean-squared error for both estimators steadily decreases. However, the $\dr$ estimator consistently exhibits mean squared error values approximately 10 times higher than the $\tdr$ estimator. In addition, the truncated estimator demonstrates similar performance across both truncation policies, highlighting the robustness of $\tdr$ estimator with respect to truncation policies. 

\vspace{0.5\baselineskip}
\noindent\textbf{Experiment 2.}
We follow data generation Setup 2 for $\lambda_0=0.1$ and $\lambda_1=0.9$, and we consider the treatment probabilities of $q_b=0.1$, and $q_e=1$ for behavior and evaluation policies, respectively. In this experiment, higher treatment probability under the evaluation policy drives the random walk towards higher states and violates the strong distributional overlap assumption. We follow the temporal difference learning method for the long-run average reward given in \eqref{eq: td-long-run} to estimate $\hQ_e$, where a single trajectory of length $T=5000$ with parameters $\nu_1=\nu_2=0.05$ are used. For the OPE problem, we run $\tdr$ with two truncation rates, $\tau_t = t^{0.7}$ and $\tau_t = T^{0.7}$. Specifically, we consider the trajectory lengths $T \in \{256, 512, 1024, 2048, 4096\}$. For each $T$ value and for each of the three estimators ($\tdr$ with its two truncation rates and $\dr$), we compute the average $\mse$ over 10000 experiments.

The results are presented in Figure \ref{fig: tdr-self-normalization}. In this experiment, for small values of $T$, the impact of truncation is not significant, and both estimators perform almost similarly. However, as $T$ increases, the significance of proper truncation becomes more pronounced, where at $\log T \approx 3.6$ the $\dr$ estimator has mean squared error approximately 10 times higher than the $\dr$ estimator.  In addition, it can be observed that as the trajectory length increases, the decreasing trend for both $\tdr$ and $\dr$ estimators halts at around $\log T \approx 3.3$, likely due to estimation error of the $Q$-function.

\begin{figure}[t]
\begin{minipage}{.48\textwidth}
    \centering
\centering
    \includegraphics[scale=0.38]{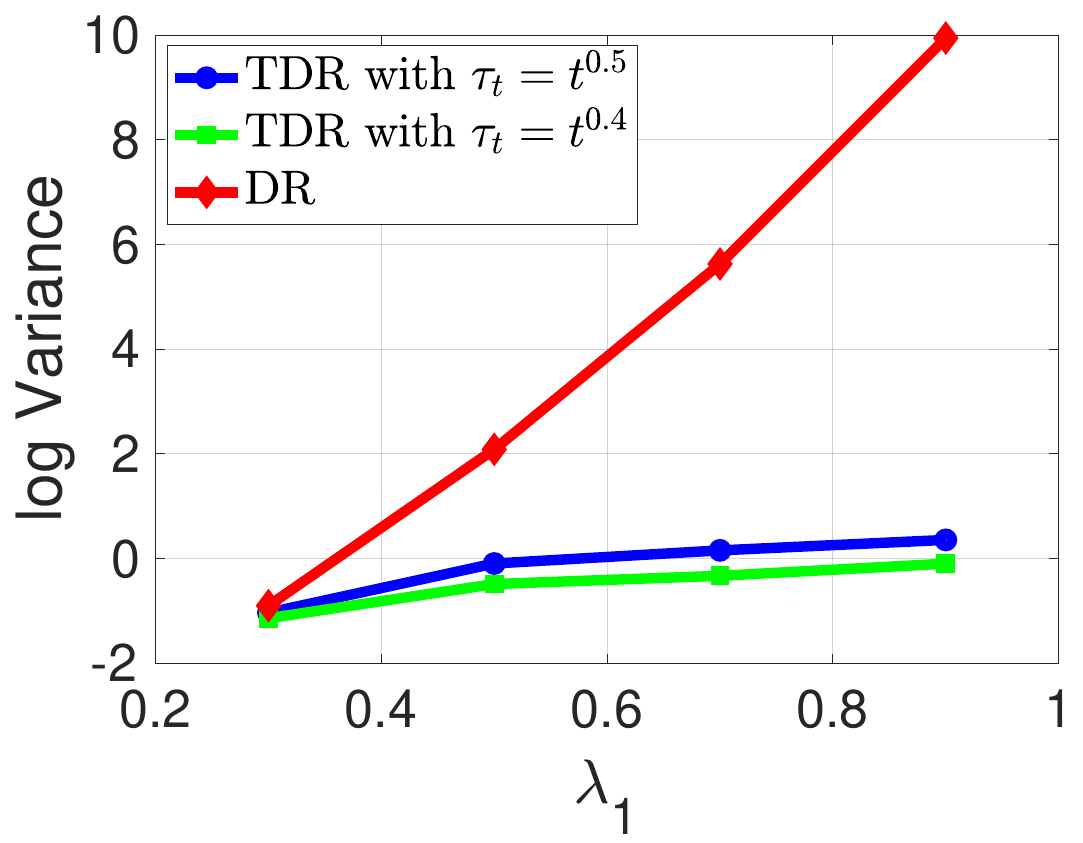}
    \caption{The  $\log$ of variance to treatment arrival rate $\lambda_1$ for the random walk on non-negative integers setup outlined in Experiment 3.}\label{fig: var}
  \end{minipage}
    \hfill
 \begin{minipage}{.48\textwidth}
    \centering
    \includegraphics[scale=0.38]{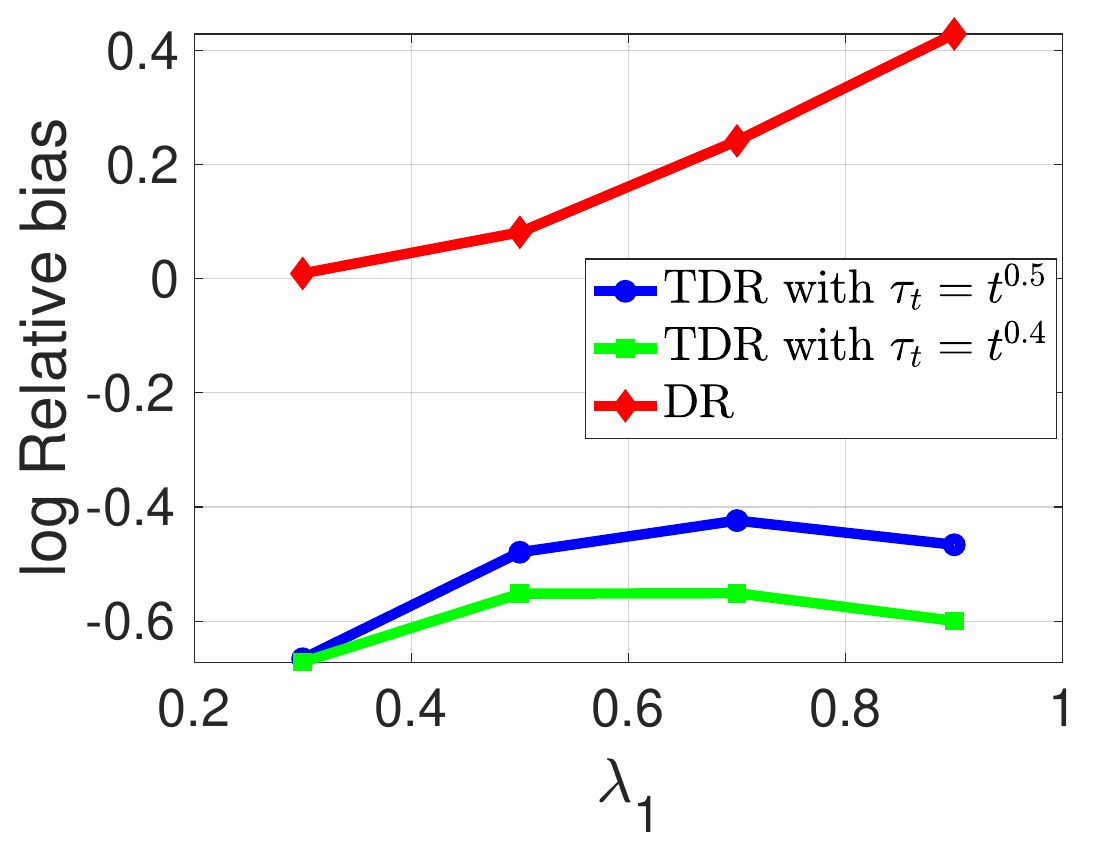}
    \caption{The log of relative bias value to treatment arrival rate $\lambda_1$ for the random walk on non-negative integers setup outlined in Experiment 3.}\label{fig: bias}
  \end{minipage}

\end{figure}

\vspace{0.5\baselineskip}
\noindent\textbf{Experiment 3.}
In this experiment, we compare the empirical performance of $\dr$ and $\tdr$ estimators in terms of bias and variance. Specifically, we adopt the MDP setting given in Setup 2, a random walk on non-negative integers. We set treatment probabilities under evaluation and behavior policies $q_e=1$ and $q_b=0.3$, respectively. In addition, we fix $\lambda_0=0.1$, and focus on estimating the discounted average reward $\mydisavg$ with $\gamma=0.5$. We let the initial distribution be the stationary distribution under the evaluation policy, i.e., $p_0=p_e$.  

We consider a range of values for $\lambda_1$, where it is selected from the set $\{0.3,0.5,0.7,0.9\}$. As we increase the values of $\lambda_1$, this leads to a shift towards states with larger values. Given that the probability of treatment is larger under the evaluation policy and the arrival rate $\lambda_1$ under treatment is much larger than the arrival rate $\lambda_0$ under control, this will drive the random walk to explore further states (larger integer values), compared to the MDP state visitation under the behavior policy. Thereby, as $\lambda_1$ increases, the strong distributional overlap assumption will be violated and result in extremely large values for $\omega(s)$. 

To estimate $\hat{q}^\gamma (s,a)$, we follow the temporal difference learning procedure given in \eqref{eq: td} with the learning rate $\alpha=0.03$ for a single trajectory of the length $T=10000$ collected under the behavior policy. For the OPE problem, we use data collected under a behavior policy of length $T = 500$ and employ the $\tdr$ estimator $\mytdrdisavg$ with two truncation rates: $\tau_t = t^{0.4}$ and $\tau_t = t^{0.5}$. For each value of $\lambda_1$, we compute the values of the three estimators ($\dr$ and $\tdr$ with two truncation rates) and report the average bias and variance across 20,000 independent experiments. 

The logarithms of the variances are shown in Figure~\ref{fig: var}, while the logarithms of relative biases, calculated as $\smash{{|{\mytdrdisavg} - \mydisavg|}/{\mydisavg}}$, are presented in Figure~\ref{fig: bias}. As $\lambda_1$ increases, violating strong distributional overlap more severely, both the variance and bias of the $\dr$ estimator grow significantly larger compared to the $\tdr$ estimator. This trend-- characterized by increasing variance and bias, persists as $\lambda_1$ continues to grow. In contrast, the truncated estimator maintains significantly smaller bias and variance, even at high values of $\lambda_1$.

\begin{figure}[t]
  \centering
  \begin{subfigure}[t]{0.48\textwidth}
    \centering
    \includegraphics[width=\linewidth]{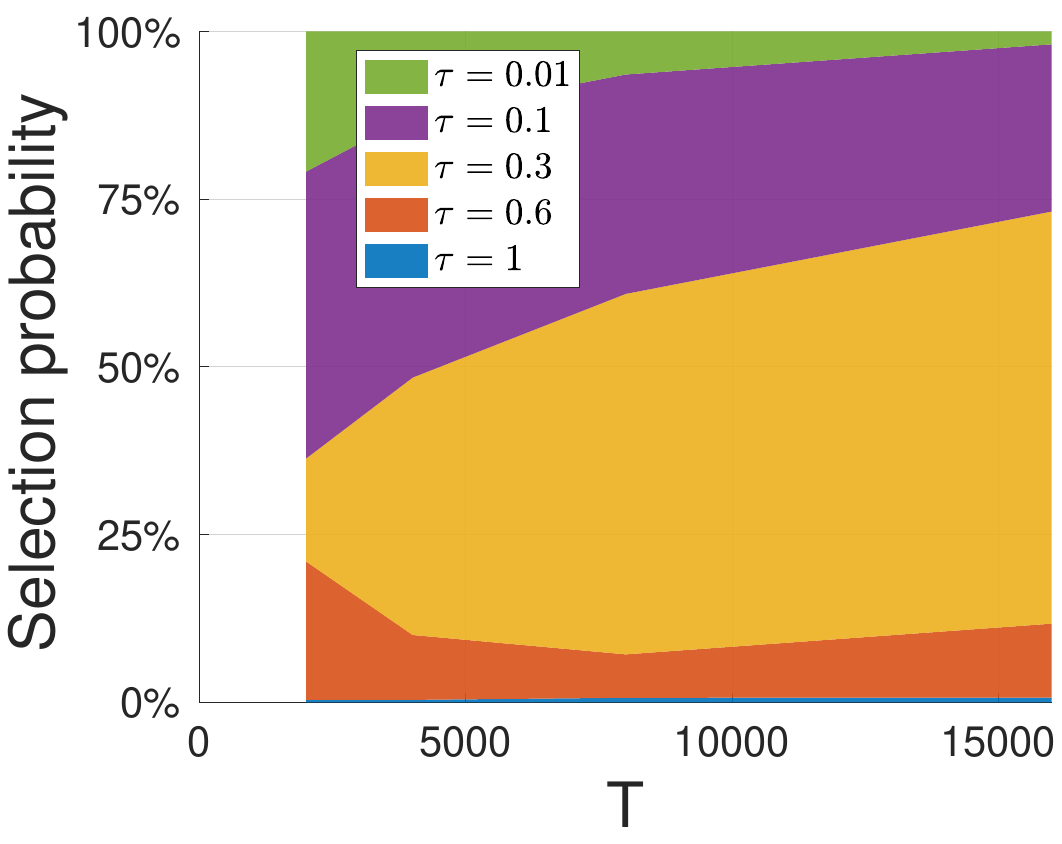}
    \caption{Percentage of times each truncation level is selected under Lepski’s method (Setup 1, 10{,}000 experiments).}
    \label{fig: selection-epski}
  \end{subfigure}
  \hfill
  \begin{subfigure}[t]{0.48\textwidth}
    \centering
    \includegraphics[width=\linewidth]{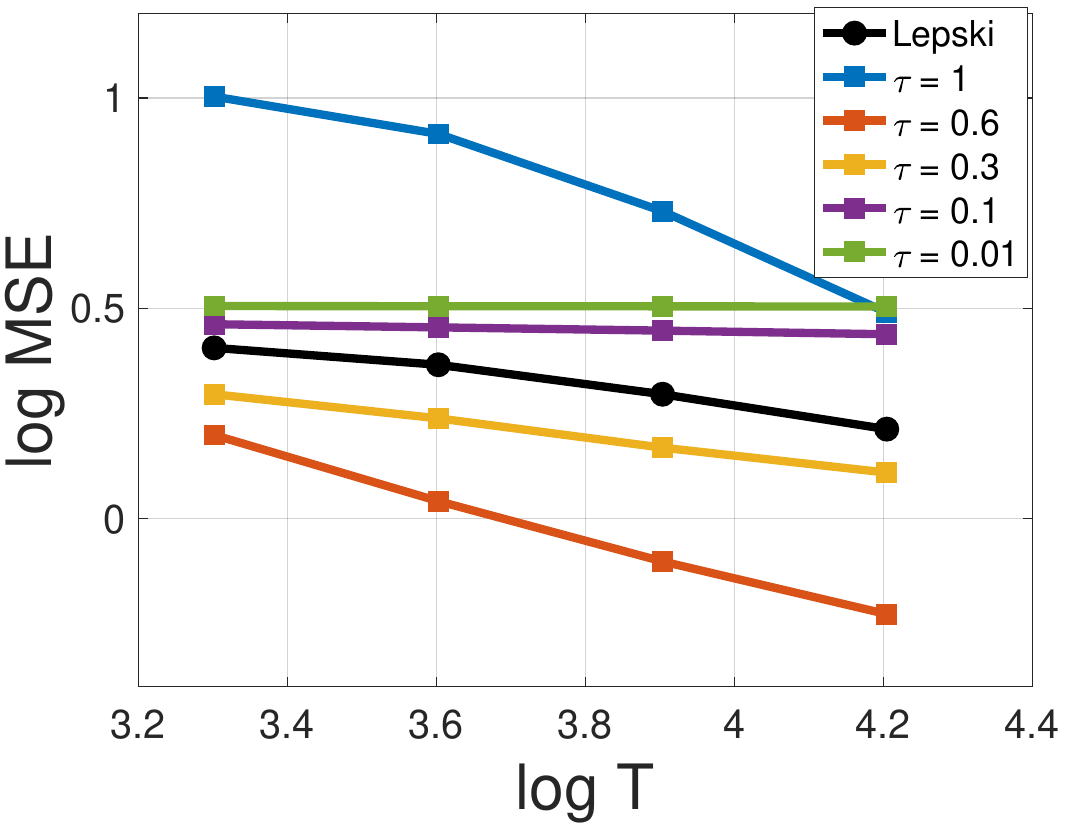}
    \caption{MSE for the OPE problem compared across truncation rates, including Lepski’s data-driven choice.}
    \label{fig: mse-Lepski}
  \end{subfigure}
  \caption{(a) Truncation-level selection frequencies and (b) resulting MSEs.}
  \label{fig:sidebyside}
\end{figure}

{
\vspace{0.5\baselineskip}
\noindent\textbf{Experiment 4.} 
In this experiment, we follow the same setup and experimental description as Experiment 1, but choose the truncation level via a data-driven procedure using Lepski’s method (Algorithm \ref{alg: lepski}). Specifically, we select the truncation rate for the TDR estimator from the grid of vlaues $t^\delta$ with $\delta \in \{1,0.6,0.3,0.1,0.01\}$. We use $B=100$ blocks for bootstrap sampling, and constructing confidence intervals of the form $[\bar{\mu}_g-c\cdot \hat{\sigma}_g,\bar{\mu}_g+c\cdot \hat{\sigma}_g]$ with $c=1$ (see Algorithm \ref{alg: lepski}). In addition, for each $T$ value we consider $10{,}000$ independent experiments to report mean-squared error rates.  First, in Figure~\ref{fig: selection-epski} we plot the percentage of times each truncation level is selected (from the $10{,}000$ experiments). It can be observed that, as the number of samples grow, Lepski’s approach successfully balances bias and variance by avoiding both overly aggressive and overly weak truncation rates; mostly moderate values of truncations are selected.

Finally, in Figure~\ref{fig: mse-Lepski}, we plot the MSE for this OPE problem across all truncation rates. The MSE achieved using Lepski’s data-driven selection is competitive, even though the method is completely oblivious to the mixing and distributional-overlap exponents. Specifically, the truncation rates selected by Lepski’s procedure outperform extreme truncation choices, i.e., $\delta\in\{1,0.1,0.01\}$.

\vspace{0.5\baselineskip}
\noindent\textbf{Experiment 5.}   In this experiment, we follow the same setup and details as in Experiment 4, with the only difference that we estimate the stationary distribution ratio $\omega^{\gamma}(\cdot; p_0)$ using the model in Algorithm~\ref{alg: omega-estimate}. We consider the OPE setting with $p_0=p_e$ and estimate the $q$-functions and $\omega$ from an auxiliary dataset of length $T=2\times 10^5$ collected under the behavior policy (both models are trained on the same auxiliary dataset, which is distinct from the dataset later used for policy evaluation). After model estimation, we follow the exact same truncation-selection setup as in Experiment 4. We report the mean-squared error averaged over $10{,}000$ experiments, as depicted in Figure~\ref{fig: all-data-driven-lepski}. It can be observed that Lepski’s method successfully balances the truncation rate between large and small values and achieves meaningful truncation.

\vspace{0.5\baselineskip}
\noindent\textbf{Experiment 6.} In the last experiment, we analyze the robustness of the TDR estimator across a range of truncation values when both nuisance components are estimated via data-driven models. Using the same setup as in Experiment 5, we run TDR with truncation rates of the form $t^{\delta}$ for $\delta \in \{0.1,0.2,0.3,0.4,0.5,0.6,0.7,0.8,0.9,1.0\}$. As it can be seen in Figure \ref{fig: mse-Lepski-robust}, for truncation exponents below $0.7$, TDR’s performance is very similar across choices. The reported numbers are averaged over $100{,}000$ experiments.

\begin{figure}[t]
  \centering
  \begin{subfigure}[t]{0.48\textwidth}
    \centering
    \includegraphics[width=\linewidth]{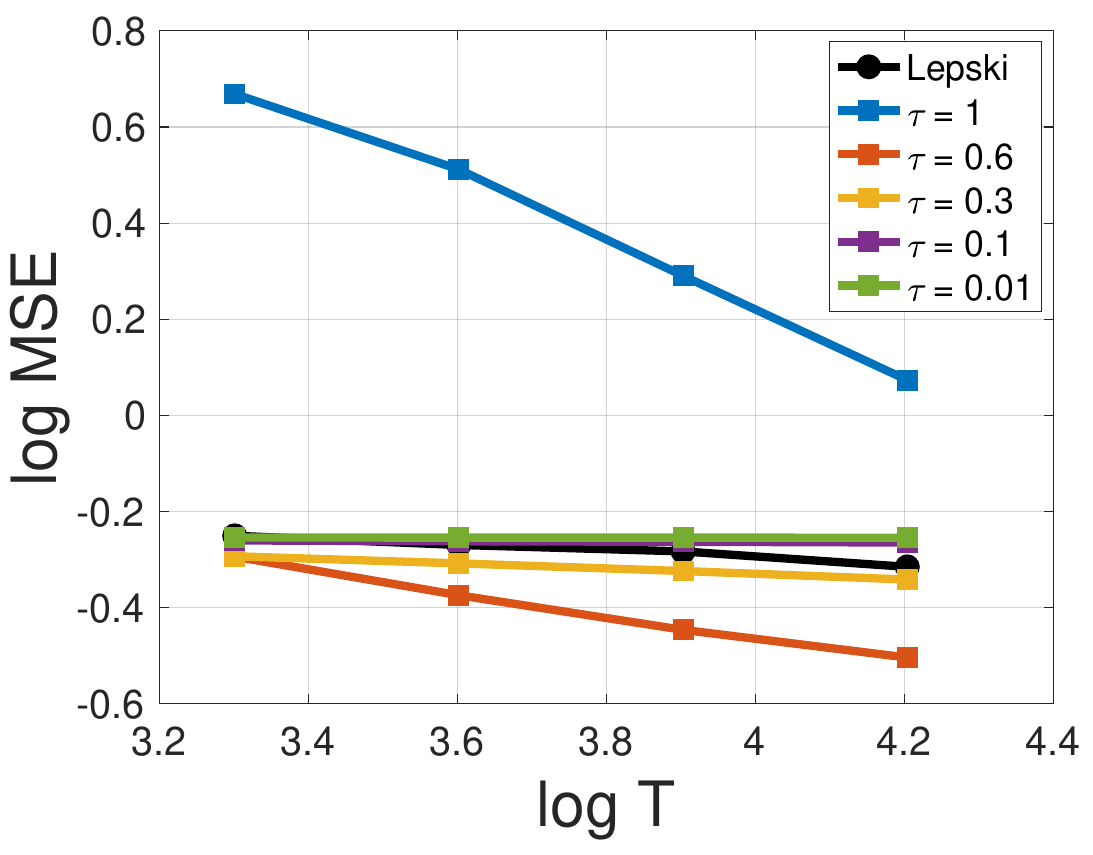}
    \caption{MSE for the OPE problem by TDR estimator, where all TDR components ($q,\omega$-functions, and truncation rates) are chosen data-driven. The performance of Lepski's method for data-driven choice of trunation rate is compared to other truncations }
    \label{fig: all-data-driven-lepski}
  \end{subfigure}
  \hfill
  \begin{subfigure}[t]{0.48\textwidth}
    \centering
    \includegraphics[width=\linewidth]{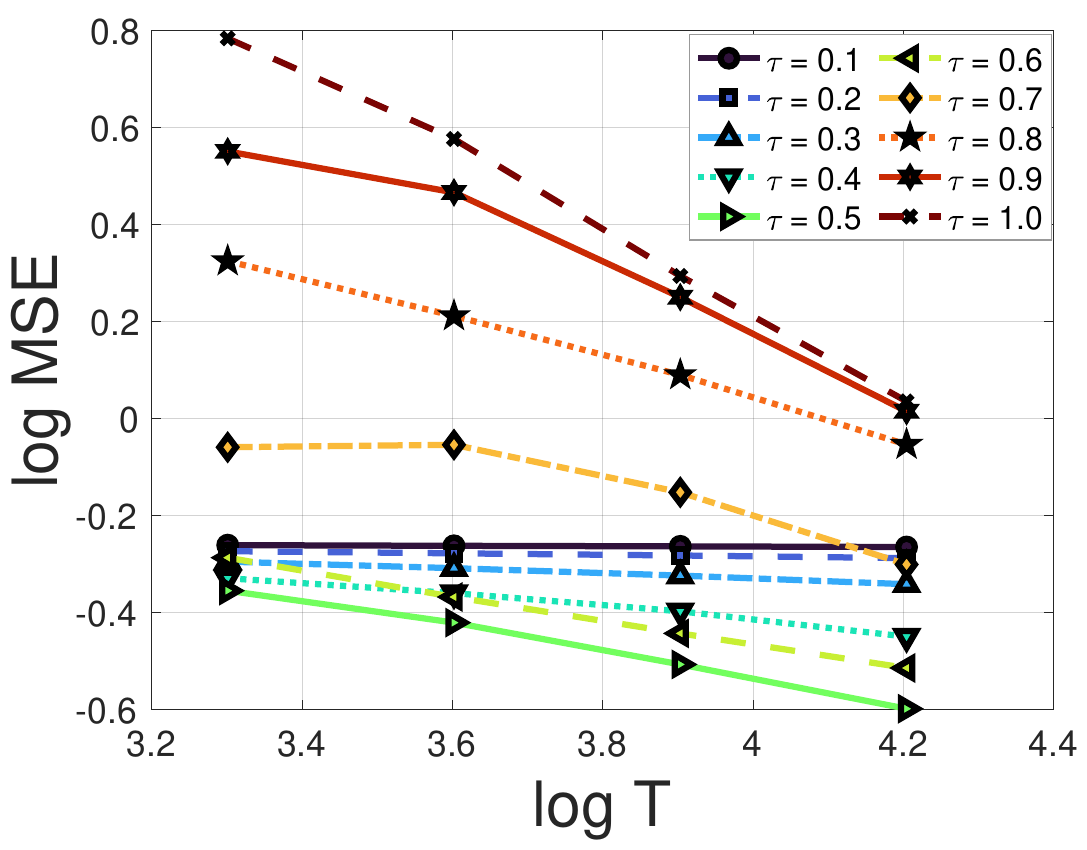}
    \caption{MSE for the OPE problem compared across a variety of truncation rates with model estimates for nuisance components $q,\omega$. }
    \label{fig: mse-Lepski-robust}
  \end{subfigure}
  \caption{(a) Lepski's selected truncation level for TDR estimate with model estimates for $q,\omega(.)$, and (b) resulting MSEs for a broad range of truncations.}
  \label{fig: data-driven-omega}
\end{figure}

}

\vspace{0.5cm}
To conclude, throughout the experiments, we observe that as the strong distributional overlap assumption weakens, the $\dr$ estimator's error rate can grow to very large values, with both the bias and variance components contributing significantly to this increase. In contrast, the $\tdr$ estimator achieves a significantly smaller mean-squared error rate by effectively controlling extreme distributional ratio values.

\section*{Acknowledgment}
This research was supported by grant N000142412091 from the Office of Naval Research.


\newpage

\bibliographystyle{plainnat}
\bibliography{mybib}

\appendix

\newpage
\section{Background results}
For reference and easier access to notation, we provide the following table. 
\subsection{Table of notations}
\begin{table}[h]
\centering
\caption{Selected notation}
\label{tab:notation}
\small
\begin{tabularx}{\textwidth}{@{} l l X l @{}}
\toprule
\textbf{Symbol} & \textbf{Type} & \textbf{Meaning} & \textbf{Where used}\\
\midrule
$\mathcal{S},\ \mathcal{A}$ & set & State space; action space & Assump.~\ref{assu:MDP} \\
$S_t,\ A_t,\ R_t$ & random variable & State, action, reward at time $t$ & Assump.~\ref{assu:MDP} \\
$\pi_b,\ \pi_e$ & policy (mapping) & Behavior policy; evaluation policy & Assump.~\ref{assu:random}, \ref{assu: policy} \\
$p_b,\ p_e$ & distribution & Stationary state distributions under $\pi_b$, $\pi_e$ & Assump.~\ref{assu:stat}, Sec.~\ref{sec:long_term} \\
$\omega^\gamma(s; p_0)$ & density ratio & Discounted state ratio $p_e^\gamma/p_b$ & Eq.~\eqref{eq:omega_gamma} \\
$\omega(s)$ & density ratio & Long-run state ratio $p_e/p_b$ & Sec.~\ref{sec:long_term} \\
$\eta(s,a)$ & ratio & Action-probability ratio $\pi_e(a\mid s)/\pi_b(a\mid s)$ & Assump.~\ref{assu: policy} \\
$C_\eta$ & constant & Policy-overlap bound for $\eta(s,a)$ & Assump.~\ref{assu: policy} \\
$\rho_k,\ C_\rho$ & coefficient, constant & $\rho$-mixing coefficient, its summability bound & Eq.~\eqref{eq:mixings-rho}, Assump.~\ref{assu:rho-mix} \\
$t_0$ & constant & Geometric-mixing parameter & Assump.~\ref{assumption: exp-mix} \\
$\delta$ & exponent & Weak-overlap exponent & Eq.~\eqref{eq: weak-exponent} \\
$\gamma$ & scalar & Discount factor ($0<\gamma<1$) & Before Eq.~\eqref{eq:qgamma} \\
$T$ & scalar & Trajectory length & Throughout \\
$p_0$ & distribution & Initial state distribution & Before Eq.~\eqref{eq:qgamma} \\
$q_e^\gamma(s,a),\ v_e^\gamma(s)$ & function & Discounted $Q$- and value functions under $\pi_e$ & Eq.~\eqref{eq:qgamma} \\
$Q_e(s,a),\ V_e(s)$ & function & Average-reward $Q$ and value functions & Eq.~\eqref{eq: Q-V} \\
$\widehat{q}_e^\gamma,\ \widehat{\omega}^\gamma(\cdot;p_0)$ & estimator & Learned models for discounted $Q$ and ratio & Eq.~\eqref{eq:DRgamma} \\
$\widehat{Q}_e,\ \widehat{\omega}$ & estimator & Learned models for $Q$ and long-run ratio & Eq.~\eqref{eq:DRlong} \\
$\widehat{\theta}^{\dr}_\gamma(p_0)$ & estimator & DR estimator (discounted reward) & Eq.~\eqref{eq:DRgamma} \\
$\widehat{\theta}^{\dr}$ & estimator & DR estimator (long-run average reward) & Eq.~\eqref{eq:DRlong} \\
$\tau_t$ & scalar & Truncation level used inside TDR & Sec.~\ref{sec: tdr-estimator}, Eq.~\eqref{eq:TDR} \\
$\widehat{\theta}^{\tdr}_\gamma(p_0),\ \widehat{\theta}^{\tdr}$ & estimator &Discounted and long-run TDR & Eq.~\eqref{eq:TDR} \\
$\kappa_T,\ \xi_T$ & rate parameter & Error rates for $\widehat{q}_e^\gamma$ and $\widehat{\omega}^\gamma$ & Assump.~\ref{assumption: model-estimates} \\
$\mu_T$ & rate parameter & Higher-moment control for $\widehat{q}_e^\gamma$ & Eq.~\eqref{eq: mu} \\
$K_T,\ \Xi_T$ & rate parameter & Error rates for $\widehat{Q}_e$ and $\widehat{\omega}$ & Assump.~\ref{assumption: model-estimates-gamma-1} \\
$M_T$ & rate parameter & Higher-moment control for $\widehat{Q}_e$ & Eq.~\eqref{eq: M} \\
$\sigma_b^2$ & scalar & Asymptotic variance (discounted case) & Eq.~\eqref{eq: sigma_b} \\
$\Sigma_b^2$ & scalar & Asymptotic variance (long-run case) & Eq.~\eqref{eq: Sigma-b-2} \\
$\nu$ & scalar & TD learning rate for discounted $q$-learning & Eq.~\eqref{eq: td} \\
$\nu_1,\ \nu_2$ & scalar & TD learning rates for $Q$ & Eq.~\eqref{eq: td-long-run} \\
$\ell$ & integer & Block length for moving-block bootstrap & Alg.~\ref{alg: lepski} \\
$B$ & integer & Number of bootstrap draws & Alg.~\ref{alg: lepski} \\
\bottomrule
\end{tabularx}
\end{table}

\subsection{Proof of Lemma \ref{lemma: bell-q}}\label{proof: lemma: bell-q}

First, note that we have
\begin{align*}
q_e^{\gamma}(s,a)&=\E_{\pi_e}\left[\sum_{t=0}^{\infty} R_t \gamma^t|S_0=s,A_0=a\right]\\
&=\E[R_0|S_0=s,A_0=a]+ \gamma \E\left[\sum_{t=1}^{\infty} R_t \gamma^{t-1}|S_0=s,A_0=a\right]\\
&=\E[R_0|S_0=s,S_0=a]+ \gamma \int \E\left[\sum_{t=1}^{\infty} R_t \gamma^{t-1}\Big |S_1=s'\right]p_s(s'|s,a)\de \lambda (s') \\
&=\E[R_0|S_0=s,A_0=a]+ \gamma \int v_e^{\gamma}(s') p_s(s'|s,a)\de \lambda (s') \\
&=\E[R_0|S_0=s,A_0=a]+\gamma \E [v_e^{\gamma}(S_1)|S_0=s,A_0=a]\,.
\end{align*}
This gives us
\[
\E[-q_e^{\gamma}(S_0,A_0)+R_0 + \gamma v_e^{\gamma}(S_1)|S_0,A_0]=0\,.
\]
This completes the proof.  We next prove the result for the long-run differential value functions. 
\begin{align*}
Q_e(s,a)&= \E_{\pi_e}\left[\sum_{t=0}^{\infty}(R_t-\myavg)\Big|S_0=s,A_0=a\right]\\
&=\E_{\pi_e}[R_0|S_0=s,A_0=a]-\myavg+ \E\left[\sum_{t=1}^{\infty}(R_t-\myavg)\Big|S_0=s,A_0=a \right]\\
&=\E[R_0|S_0=s,A_0=a]-\myavg+ \int V_e(s') p_s(s'|s,a) \de \lambda_s(s')\\
&=\E[R_0|S_0=s,A_0=a]-\myavg+ \int V_e(s') p_s(s'|s,a) \de \lambda_s(s')\,.
\end{align*}
This completes the proof. 


\subsection{Proof of Lemma \ref{lemma: bell-prob} }\label{proof: lemma-omega}

From the definition of the discounted average frequency visits $\pinf{.}$ we have
\begin{align}\label{eq: bell-omega-1}
\pinf{s'}&=(1-\gamma)p_0(s')+(1-\gamma)\gamma\sum_{t=1}^{\infty}\gamma^{t-1} \prob_{\pi_e}(S_t=s'|s_0\sim p_0)\,.
\end{align}
Let $p_{\pi_e}(s'|s)$ denote the marginal state transition under the evaluation policy $\pi_e$, specifically let
\begin{equation}\label{eq: bell-omega-0}
p_{\pi_e}(s'|s)=\int p_s(s'|s,a)\pi_e(a|s) \de \lambda_a(a)\,.
\end{equation}
Using this for $t \ge 1$ brings us
\begin{equation}\label{eq: bell-omega-2}
\prob_{\pi_e}(S_t=s'|s_0\sim p_0)=\int \prob_{\pi_e}(S_{t-1}=s|s_0\sim p_0) p_{\pi^e}(s'|s)\de \lambda_s(s)\,.
\end{equation}
Employing \eqref{eq: bell-omega-2} in \eqref{eq: bell-omega-1} yields
\begin{align*}
\pinf{s'}&=(1-\gamma)p_0(s')+\gamma (1-\gamma)\sum_{t=1}^{\infty} \gamma^{t-1} \int \prob_{\pi_e}(S_{t-1}=s|s_0\sim p_0) p_{\pi^e}(s'|s)\de \lambda_s(s) \\
&=(1-\gamma)p_{0}(s')+\gamma \int \Big(p_{\pi^e}(s'|s)\cdot (1-\gamma) \sum_{t=1}^{\infty}  \gamma^{t-1} \prob_{\pi_e}(S_{t-1}=s|s_0\sim p_0) \Big) \de \lambda_s(s)  \\
&=(1-\gamma)p_{0}(s')+\gamma \int \pinf{s} p_{\pi^e}(s'|s)\de \lambda_s(s) \,,
\end{align*}
where in the last relation we used the definition of $\pinf(.)$. In summary, the above gives us
\begin{equation}\label{eq: bell-omega-3}
\pinf{s'}= (1-\gamma)p_0(s')+\gamma \int \pinf{s} p_{\pi^e}(s'|s)\de \lambda_s(s)\,.
\end{equation}
In the next step, using $\omega^{\gamma}(s;p_0)=\frac{\pinf{s}}{p_b(s)}$ yields

\begin{align*}
\E_{p_b}[f(S')\omega^{\gamma}(S';p_0)]&=\E_{\pinf{.}}[f(S')]\\
&=\int f(s')\pinf{s'}\de \lambda_s(s')\,.
\end{align*}
Employing \eqref{eq: bell-omega-3} in the above gives us
\begin{align*}
\E_{p_b}[f(S')\omega^{\gamma}(S';p_0)]&=(1-\gamma)\int f(s')p_{0}(s')\de \lambda_s(s') + \gamma \iint f(s') p_{\pi_e}(s'|s)\pinf{s}\de \lambda_s(s) \de \lambda_s(s')\,.
\end{align*}
We next use $\pinf{s}=\omega^{\gamma}(s;p_0) p_b(s)$ to get
\begin{align*}
\E_{p_b}[f(S')\omega(S')]
&=(1-\gamma)\int f(s')p_{0}(s')\de \lambda_s(s') + \gamma \iint f(s') p_{\pi_e}(s'|s)\omega^{\gamma}(s;p_0) p_b(s)\de \lambda_s(s) \de \lambda_s(s')\\
&=(1-\gamma)\int f(s')p_{0}(s')\de \lambda_s(s') \,\\
&\,~ + \gamma \iint f(s') \Big( \int p_s(s'|s,a)\pi_e(s|a) \de \lambda_a(a) \Big) \omega^{\gamma}(s;p_0) p_b(s) \de \lambda_s(s) \de \lambda_s(s')\,,
\end{align*}
where in the last relation we deployed \eqref{eq: bell-omega-0}. Finally, by using $\eta(s,a)=\frac{\pi_e(a|s)}{\pi_b(a|s)}$ we arrive at 
\begin{align*}
\E_{p_b}[f(S')\omega^{\gamma}(S';p_0)]&=(1-\gamma)\E_{p_{0}}[f(S)]\\
\,~&+\gamma \iiint f(s') \omega^{\gamma}(s;p_0)\eta(s,a) p_b(s)p_s(s'|s,a)\pi_b(a|s)\de\lambda_s(s)\de \lambda_a(a)  \de\lambda_s(s')\,.
\end{align*}
Put all together, we get
\[
\E_{p_b}[f(S)\omega^{\gamma}(S;p_0)]=(1-\gamma)\E_{p_0}[f(S)]+ \gamma \E_{(S,A,S')\sim p_b(S)\pi_b(A|S)p(S'|S,A)}\left[ \omega^{\gamma}(S;p_0) \eta(S,A) f(S') \right]\,.
\]
This completes the proof. 



We next move to the proof for the long-run density ratio function $\omega(s)$. Similar to the previous part, we let $\eta(s,a)=\frac{\pi_e(a|s)}{\pi_b(a|s)}$.  We have 
\begin{align*}
\E[\omega(S)\eta(S,A)f(S')]&=\iiint \omega(S)\eta(S,A)f(S')p_b(S)\pi_b(A|S)p_s(S'|S,A) \de \lambda_a(A) \de \lambda_s(S)\de \lambda_s(S')\,.
\end{align*}
We next use the definition for density ratio function $\omega(s)$ and $\eta(s,a)$ to arrive at
\begin{align*}
\E[\omega(S)\eta(S,A)f(S')]&=\iiint f(s') p_e(s) \pi_e(a|s)p_s(s'|s,a) \de \lambda_a(a) \de \lambda_s(s)\de \lambda_s(s')\\
&=\int f(s') \left( \iint  p_e(s) \pi_e(a|s)p_s(s'|s,a)  \de \lambda_a(a)\de \lambda_s(s)  \right)   \de \lambda_s(s')\,.
\end{align*}
We next use the invariance of stationary state distribution $p_e$ under evaluation policy, which reads as
\[
 \iint p_e(s)\pi_e(a|s)p_s(s'|s,a)  \de \lambda_a(a) \de \lambda_s(s)=p_e(s')\,.
\] 
Using this in the above yields
\begin{align*}
\E[\omega(S)\eta(S,A)f(S')]&=\int f(s') p_e(s')  \de \lambda_s(s')\\
&=\int f(s')\omega(s') p_b(s')  \de \lambda_s(s')\\
&=\E_{p_b}[\omega(S)f(S)]\,.
\end{align*}
This completes the proof.


\section{Preliminaries and technical lemmas}

We let $\eta(s,a)=\frac{\pi_e(a|s)}{\pi_b(a|s)}$.  For states $s,s'\in \cS$, an action $a\in \cA$, reward value $r\in \cR$, and functions $f_{\omega}:\cS\to \reals$, $f_{q}: \cS\times \cA\to \reals$, and $f_v(S)=\E_{\pi_e}[f_q(S,A)|S]$ we let $\psi(s,a,r,s';f_{\omega},f_q)$ represents the following
\[
\psi(s,a,r,s';f_\omega,f_q)=(1-\gamma) \E_{S_0\sim p_0}\left[f_v(S_0)\right] + f_{\omega}(s)\eta(s,a)(r+\gamma f_v(s')-f_q(s,a))\,.
\]
For the tuple $(S_t,A_t,R_t,S_{t+1})$, we introduce the following shorthand:
\[
\psi_t=\psi\left(S_t,A_t,R_t,S_{t+1};\omega^{\gamma}(.;p_0),q_e^{\gamma}\right)\,.
\]
We also let $\homg_t=\hat{\omega}^{\gamma}(S_t;p_0)$ (for the discounted estimation problem) or $\homg_t=\hat{\omega}(S_t)$ (for the long-run average estimation problem), $\eta_t=\eta(S_t,A_t)$, $\hq_t=\hq_e(S_t)$ and $\hv_t=v^{\gamma}_e(S_t)$. In addition, for model estimates $\hat{\omega}^{\gamma}(.;p_0)$ and $\hq_e$ we let 
\[
\hpsi_t=\psi\Big(S_t,A_t,R_t,S_{t+1};\hat{\omega}^{\gamma}(.;p_0),\hat{q}_e^{\gamma}\Big)\,.
\]
It is easy to observe that the $\dr$ estimator for the discounted average reward can be written as the following:
\begin{align*}
\hat{\theta}_\gamma^{\dr}(p_0)&=\frac{1}{T}\sum\limits_{t=1}^{T}\hpsi_t\,.
\end{align*}


Similar to discussed above, for the long-run model estimates $\hQ_e$ and $\hV_e$, we also let $V_t=V_e(S_t)$, $\hV_t=\hV_e(S_t)$, $Q_t=Q_e(S_t,A_t)$, and $\hQ_t=\hQ_e(S_t,A_t)$. In addition, for the long-run average reward $\myavg$ estimation problem, we adopt the following shorthands:
\begin{align}
c_t&=\omega_t\eta_t\,,\nonumber\\
\hc_t&=\hat{\omega}_t \eta_t \,, \nonumber\\
b_t&=(R_t+V_{t+1}-Q_t-\myavg)\,,\nonumber\\
\hb_t&=(R_t+\hV_{t+1}-\hQ_t-\myavg)\label{eq: a-b}\,.
\end{align}

We end this section by stating lemmas that we will use later in the proof of Theorems. 

\begin{lemma}\label{lemma-first-little-o}

Under Assumptions \ref{assu:MDP}-\ref{assu:stat}, and $\gamma$-discounted $\delta$-weak distributional overlap for some $\delta>1$, suppose that estimates $\hq_e(.)$ and $\homg^{\gamma}(.;p_0)$ satisfy Assumption \ref{assumption: model-estimates}, and equation \eqref{eq: mu} for parameters $\kappa_T, \xi_T,$ and $\mu_T$ such that $\kappa_T\vee \xi_T \vee \mu_T=o_T(1)$.  Then the following holds
 \[
 \sqrt{T}\Big(\frac{1}{T}\sum_{t=1}^T (\hpsi_t - \psi_t)- \big(\hat{\th}^{\dr}_\gamma(p_0)-\mydisavg\big)\Big)=o_T(1)\,.
 \]

\end{lemma}
The proof of this Lemma can be seen in Section \ref{proof: lemma-first-little-o}.
\begin{lemma} [Doubly robustness]\label{lemma-second-little-o}
Under Assumptions \ref{assu:MDP}-\ref{assu:stat}, for estimates  $\hq_e(.)$ and $\homg^{\gamma}(.;p_0)$ satisfying Assumption \ref{assumption: model-estimates} with parameters $\kappa_T$ and $\xi_T$, we have
\[
\left|\hat{\th}_{\gamma}^{\dr}(p_0)-\mydisavg\right| \lesssim \kappa_T\xi_T\,.
\]
\end{lemma}

We refer to Section \ref{proof: lemma-second-little-o} for the proof of Lemma \ref{lemma-second-little-o}.

\begin{lemma}\label{lemma-first-little-o-gamma-1}

We consider the off-policy evaluation problem under Assumptions \ref{assu:MDP}-\ref{assu:stat}, geometric mixing assumption \ref{assumption: exp-mix}, and the long-run $\delta$-weak distributional overlap condition for $\delta>1$. In addition, we suppose that estimates $\hQ_e$ and $\homg$ satisfy Assumption \ref{assumption: model-estimates-gamma-1} and equation \eqref{eq: M} for parameters $\Xi_T,K_T$, and $M_T$ such that $\Xi_T \vee M_T=o_T(1)$.  Then we have
 \[
 \frac{1}{\sqrt{T}}\sum_{t=1}^T \left(\hc_t\hb_t-c_tb_t-\E[\hc_t\hb_t]+\E[c_t b_t]\right)=o_T(1)\,.
 \]
\end{lemma}
The proof of this Lemma can be seen in Section \ref{proof: lemma-first-little-o-gamma-1}.
\begin{lemma} \label{lemma-second-little-o-gamma-1}
Under Assumptions \ref{assu:MDP}-\ref{assu:stat}, for estimates $\hQ_e,\hat{\omega}$ satisfying Assumption \ref{assumption: model-estimates-gamma-1} with parameters $K_T$ and $\Xi_T$, we have $\E[c_tb_t]=0$, and also the following holds
\[
\left|\E[\hc_t\hb_t]\right| \le  C_\eta K_T\Xi_T\,,
\]
where $C_\eta$ is the policy overlap parameter given in Assumption \ref{assu: policy}.

\end{lemma}

We refer to Section \ref{proof: lemma-second-little-o-gamma-1} for the proof of Lemma \ref{lemma-second-little-o-gamma-1}.


\subsection{Proof of Lemma \ref{lemma-first-little-o}}\label{proof: lemma-first-little-o}
By using the identity $(\widehat{a} \widehat{b}-ab)=(\widehat{a}-a)(\widehat{b}-b)+a(\widehat{b}-b)+(\widehat{a}-a)b$ we get
\begin{align*}
\hpsi_t-\psi_t&=\eta_t(\homg_t-\omega_t)(\gamma\hv_{t+1}-\gamma v_{t+1}+q_t-\hq_t)+(1-\gamma) \E_{p_0}[\hv_e(S)-v^{\gamma}_e(S)]\\
&~\,+ \eta_t \omega_t (q_t-\gamma v_{t+1} -\hq_t+\gamma \hv_{t+1})+ \eta_t(\homg_t-\omega_t)(R_t+\gamma v_{t+1}-q_t)\,.
\end{align*}
In the next step, using $\var(\hpsi_t-\psi_t)\leq \E[(\hpsi_t-\psi_t)^2]$ in conjunction with Cauchy-Schwartz inequality we arrive at
\begin{align*}
\var(\hpsi_t-\psi_t)&\le 4\E[\eta_t^2(\homg_t-\omega_t)^2(\gamma\hv_{t+1}-\gamma v_{t+1}+q_t-\hq_t)^2]+4(1-\gamma)^2 \E_{p_0}[(\hv_e(S)-v_e^{\gamma}(S))^2]\\
&~\,+ 4\E[\eta_t^2 \omega_t^2 (q_t-\gamma v_{t+1} -\hq_t+\gamma \hv_{t+1})^2]+ 4\E[\eta_t^2(\homg_t-\omega_t)^2(R_t+\gamma v_{t+1}-q_t)^2]\,.
\end{align*}
We next deploy $\eta_t\le C_\eta$ (policy overlap assumption) and $\max(\hv_e,v_e^{\gamma}, q_e^{\gamma},\hq_e)\le \frac{1}{1-\gamma}R_{\max}$ to get
\begin{align*}
\var(\hpsi_t-\psi_t)&\lesssim \E_{p_b}\left[\|\homg^{\gamma}(S;p_0)-\omega^{\gamma}(S;p_0)\|_2^2\right]+ \E_{p_0}\left[\|\hv_e(S)-v_e^{\gamma}(S)\|_2^2\right]\\
&\,+\E_{p_b}\left[\omega(S)^2\|\hq_e(S)-q_e^{\gamma}(S)\|_2^2\right] + \E_{p_b}\left[\|\homg^{\gamma}(S;p_0)-\omega^{\gamma}(S;p_0)\|_2^2\right]\,.
\end{align*}
We then use Holder's inequality $\E[|xy|]\le \E[|x|^p]^{\frac{1}{p}} \E[|x|^q]^{\frac{1}{q}}$ for $p=\frac{1+\delta}{2}$ and $q=\frac{1+\delta}{\delta-1}$. This yields
\begin{align*}
\var(\hpsi_t-\psi_t)&\lesssim \E_{p_b}\left[\|\homg^{\gamma}(S)-\omega^{\gamma}(S)\|_2^2\right]+ \E_{p_0}\left[\|\hv_e(S)-v_e^{\gamma}(S)\|_2^2\right]\\
&\,+\E_{p_b}\left[\omega^{\gamma}(S;p_0)^{1+\delta}\right]^{\frac{2}{1+\delta}}\E\left[\|\hq_e(S)-q_e^{\gamma}(S)\|_2^{\frac{2(1+\delta)}{\delta-1}}\right]^{\frac{\delta-1}{\delta+1}}\,.
\end{align*}

Finally, employing the $\gamma$-discounted $\delta$-weak distributional overlap condition in along with estimates Assumption \ref{assumption: model-estimates} and equation \eqref{eq: mu} yields
\begin{equation}\label{eq: var_uppeR_tmp}
\var(\hpsi_t-\psi_t)=O\left(\kappa_T^2 \vee \xi_T^2\vee \mu_T^2\right)\,.
\end{equation}
 In the next step, we have
\begin{align*}
\var\left(\sqrt{T}\Big(\frac{1}{T}\sum_{t=1}^T (\hpsi_t - \psi_t)\Big)\right)&=\frac{1}{T}\sum_{t=1}^T \var(\hpsi_t-\psi_t)+ \frac{1}{T}\sum_{\ell\neq k }^T \cov(\hpsi_k-\psi_k, \hpsi_\ell-\psi_\ell)\\
&=\var(\psi_1-\hpsi_1)+  \frac{\var(\psi_1-\hpsi_1)}{T}\sum_{\ell\neq k }^T \corr(\hpsi_k-\psi_k, \hpsi_\ell-\psi_\ell))\\
&\leq \var(\psi_1-\hpsi_1)\left(1+ \frac{2}{T}\sum_{\ell<k}^{T} \rho_{k-\ell}\right)\,,
\end{align*}
where in the last relation we used the definition of $\rho_n$ mixing. Finally, using Assumption \ref{assu:rho-mix} that $\sum \rho_n <+\infty$ in along with \eqref{eq: var_uppeR_tmp} yields
\[
\var\left(\sqrt{T}\Big(\frac{1}{T}\sum_{t=1}^T (\hpsi_t - \psi_t)\Big)\right)=O\left(\kappa_T^2 \vee \xi_T^2\vee \mu_T^2\right)\,.
\]
Given that $\E[ (\hpsi_t - \psi_t)]=\widehat{\th}_\gamma^{\dr}(p_0)-\mydisavg$, by an application of Chebyshev's inequality we get for every $\eps>0$ we have

 \[
 \lim_{T\to \infty}\prob\left(\sqrt{T}\Big(\frac{1}{T}\sum_{t=1}^T (\hpsi_t - \psi_t)- (\widehat{\th}_\gamma^{\dr}(p_0)-\mydisavg)\Big) \geq \eps\right) \leq \lim_{T\to \infty} \frac{O\left(\kappa_T^2 \vee \xi_T^2\vee \mu_T^2\right)}{\eps^2}=0\,,
 \]
 where we used the fact that $\kappa_t \vee \xi_T\vee \mu_T=o_T(1)$. 
 This completes the proof that $\sqrt{T}\Big(\frac{1}{T}\sum_{t=1}^T (\hpsi_t - \psi_t)- (\widehat{\th}_\gamma^{\dr}(p_0)-\mydisavg)\Big)=o_T(1)$. 


\subsection{Proof of Lemma \ref{lemma-second-little-o}} \label{proof: lemma-second-little-o}
By using the identity $(\widehat{a} \widehat{b}-ab)=(\widehat{a}-a)(\widehat{b}-b)+a(\widehat{b}-b)+(\widehat{a}-a)b$ we get
\begin{align*}
\hpsi_t-\psi_t&=\eta_t(\homg_t-\omega_t)(\gamma\hv_{t+1}-\gamma v_{t+1}+q_t-\hq_t)+(1-\gamma) \E_{p_0}[\hv_e(S)-v^{\gamma}_e(S)]\\
&~\,+ \eta_t \omega_t (q_t-\gamma v_{t+1} -\hq_t+\gamma \hv_{t+1})+ \eta_t(\homg_t-\omega_t)(R_t+\gamma v_{t+1}-q_t)\,.
\end{align*}
This gives us
\begin{align}
\hat{\th}^{\dr}_\gamma(p_0)-\mydisavg&=\frac{1}{{T}}\E\left[\sum_{t=1}^{T}(\hpsi_t-\psi_t)\right] \nonumber\\
&=\frac {1}{{T}} \sum_{t=1}^T \E\left[\eta_t(\homg_t-\omega_t)(\gamma\hv_{t+1}-\gamma v_{t+1}+q_t-\hq_t)+(1-\gamma) \E_{p_0}[\hv_e(S)-v^{\gamma}_e(S)]\right]\nonumber \\
&~\,+ \frac{1}{{T}}\sum_{t=1}^T \E\left[\eta_t \omega_t (q_t-\gamma v_{t+1} -\hq_t+\gamma \hv_{t+1})+ \eta_t(\homg_t-\omega_t)(R_t+\gamma v_{t+1}-q_t)\right]\label{eq: bias-tmp-1} \,.
\end{align}
By using Jensen's inequality we obtain
\begin{align*}
\E\left[(\homg_t-\omega_t)(\gamma\hv_{t+1}-\gamma v_{t+1}+q_t-\hq_t)\right]&\le \E\left[(\homg_t-\omega_t)^2\right]^{1/2} \E\left[(\gamma\hv_{t+1}-\gamma v_{t+1}+q_t-\hq_t)^2\right]^{1/2}\\
&\lesssim \kappa_T\xi_T\,.
\end{align*}
Using this in \eqref{eq: bias-tmp-1} yields
\begin{align}
\Big|\hat{\th}^{\dr}_\gamma(p_0)-\mydisavg\Big|&\lesssim \kappa_T\xi_T+ \Big| \frac {1}{{T}} \sum_{t=1}^T  \left( \E[\eta_t \omega_t (q_t-\gamma v_{t+1} -\hq_t+\gamma \hv_{t+1})]+(1-\gamma) \E_{p_0}[\hv_{e}(S)-v_{e}^{\gamma}(S)]\right)\Big|\nonumber\\
&\,+ \Big|\frac{1}{{T}} \sum_{t=1}^{T}\E[\eta_t(\homg_t-\omega_t)(R_t+\gamma v_{t+1}-q_t)]\Big|\label{eq: bias-tmp-2}\,.
\end{align}
In the next step, we deploy Lemma \ref{lemma: bell-q} to get
\[
\E[\eta_t(\homg_t-\omega_t)(R_t+\gamma v_{t+1}-q_t)]=\E[\eta_t(\homg_t-\omega_t) \E[(R_t+\gamma v_{t+1}-q_t)|S_t,A_t]]=0\,.
\]
By employing this in \eqref{eq: bias-tmp-2} we obtain
\begin{align}\label{eq: bias-tmp-3} 
|\hat{\th}^{\dr}_\gamma(p_0)-\mydisavg|\lesssim \kappa_T\xi_T +  \Big| \frac {1}{{T}} \sum_{t=1}^T  \left( \E[\eta_t \omega_t (q_t-\gamma v_{t+1} -\hq_t+\gamma \hv_{t+1})]+(1-\gamma) \E_{p_0}[\hv_{}(S)-v_{}(S)]\right)\Big|
\end{align}
 We next use Lemma \ref{lemma: bell-prob} with the measurable function $f(s)=-v_e^{\gamma}(s) +\hv_e(s)$ to get
\begin{equation}\label{eq: bias-tmp-5}
\gamma \E[\eta_t \omega_t (-v_{t+1} +\hv_{t+1})]+(1-\gamma) \E_{p_0}[\hv_{e}(S)-v_{e}^{\gamma}(S)]=\E[\omega_t(\hv_t-v_t)]\,.
\end{equation}
In addition, we have
\begin{align}
\E[\eta_t\omega_t(q_t-\hq_{t})]&=\E[\E[\eta_t\omega_t(q_t-\hq_{t})|S_t]]\nonumber\\
&=\E[\omega_t \E_{\pi^e}[q_t-\hq_t|S_t] ]\nonumber\\
&=\E[\omega_t (v_t-\hv_t)]\,. \label{eq: bias-tmp-4}
\end{align}
Combining \eqref{eq: bias-tmp-4} and \eqref{eq: bias-tmp-5} brings us
\[
\E[\eta_t \omega_t (q_t-\gamma v_{t+1} -\hq_t+\gamma \hv_{t+1})]+(1-\gamma) \E_{p_0}[\hv_{e}(S)-v_{e}^{\gamma}(S)]=0\,.
\]
Plugging this into \eqref{eq: bias-tmp-3} yields
\[
|\hat{\th}^{\dr}_\gamma(p_0)-\mydisavg| \lesssim  \kappa_T\xi_T\,.
\]


\subsection{Proof of Lemma \ref{lemma-first-little-o-gamma-1}}\label{proof: lemma-first-little-o-gamma-1}
Following shorthands given in \eqref{eq: a-b} we obtain
\[
\var(\hc_t\hb_t-c_tb_t)\le 3\E\left[c_t^2(\hb_t-b_t)^2\right]+3\E\left[(\hc_t-c_t)^2 b_t^2\right]+3\E\left[(\hc_t-c_t)^2(\hb_t-b_t)^2\right]\,. 
\]
This gives us
\[
\var(\hc_t\hb_t-c_tb_t) \lesssim \E\left[\omega(S)^2(\hQ(S,A)-Q(S,A))^2\right] + \E_{p_b}\left[(\hat{\omega}(S)-\omega(S))^2\right]\,.
\]
We then use Holder's inequality $\E[|X||Y|]\le \E[|X|^p]^{\frac{1}{p}} \E[|Y|^q]^{\frac{1}{q}}$ for $p= \frac{1+\delta}{2}$ and $q=\frac{\delta+1}{\delta-1}$. This yields
\begin{align*}
\var(\hc_t\hb_t-c_tb_t)&\lesssim \E_{p_b}\left[\|\omega-\hat{\omega}\|_2^2\right]+ \E_{p_b}\left[\omega(S)^{1+\delta}\right]^{\frac{2}{1+\delta}} \E_{p_b}\left[\|\hQ-Q\|_2^{\frac{2(\delta+1)}{\delta-1}}\right]^{\frac{\delta-1}{(\delta+1)}}\,.
\end{align*}

Finally, employing the $\delta$-weak distributional overlap for $\delta>1$ gives us that  $\E_{p_b}[\omega(S)^{1+\delta}]$ is bounded. This in conjunction with Assumption \ref{assumption: model-estimates-gamma-1} and equation \eqref{eq: M} brings us
\begin{equation}\label{eq: var_uppeR_tmp-adjusted}
\var(\hc_t\hb_t-c_tb_t)=O\left(M_T^2 \vee \Xi_T^2\right)\,.
\end{equation}
In the next step, we have
\begin{align*}
\var\left(\frac{1}{\sqrt{T}}\sum_{t=1}^T (\hc_t\hb_t-c_tb_t) \right)&=\frac{1}{T}\sum_{t=1}^T \var(\hc_t\hb_t-c_tb_t)+ \frac{1}{T}\sum_{\ell\neq k }^T \cov(\hc_k\hb_k-c_kb_k,\hc_\ell\hb_\ell-c_\ell b_\ell )\\
&=\var(\hc_1\hb_1-c_1b_1)+  \frac{\var(\hc_1\hb_1-c_1b_1)}{T}\sum_{\ell\neq k }^T \corr(\hc_k\hb_k-c_kb_k,\hc_\ell\hb_\ell-c_\ell b_\ell )\\
&\le \var(\hc_1\hb_1-c_1b_1)\left(1+ 2\sum_{\ell=1}^{T} \rho_{\ell}\right)\,,
\end{align*}
where in the last relation we used the definition of $\rho_n$-mixing coefficients. Using Assumption \ref{assu:rho-mix} that $\sum \rho_n \le C_\rho$ in along with \eqref{eq: var_uppeR_tmp-adjusted} yields
\[
\var\left(\frac{1}{\sqrt{T}}\sum_{t=1}^T (\hc_t\hb_t-c_tb_t)\right)=O\left(M_T^2 \vee \Xi_T^2\right)\,.
\]
Finally, by an application of Chebyshev's inequality we obtain that for every $\eps>0$ we must have
\begin{align*}
 \lim_{T\to \infty}\prob\left(\frac{1}{\sqrt{T}}\sum_{t=1}^T \Big(\hc_t\hb_t-c_tb_t-\E[\hc_t\hb_t]+\E[c_tb_t]\Big) \ge \eps\right) &\le \lim_{T\to \infty} \frac{\var\left(\frac{1}{\sqrt{T}}\sum_{t=1}^T (\hc_t\hb_t-c_tb_t)\right)}{\eps^2}  \\
 &=  \lim_{T\to \infty} \frac{O(M_T^2 \vee \Xi_T^2)}{\eps^2}=0\,,
 \end{align*}
 where we used the assumption that $M_t \vee \Xi_T=o_T(1)$. Put all together, this shows the following
  \[
\frac{1}{\sqrt{T}}\sum_{t=1}^T \Big(\hc_t\hb_t-c_tb_t-\E[\hc_t\hb_t]+\E[c_tb_t]\Big)=o_T(1)\,.
  \]


\subsection{Proof of Lemma \ref{lemma-second-little-o-gamma-1}}\label{proof: lemma-second-little-o-gamma-1}
We recall definitions of $c_t,b_t,\hc_t$, and $\hb_t$:
\[
c_t=\omega_t\eta_t,\quad \hc_t=\hat{\omega}_t \eta_t\,, \quad b_t=(R_t+Q_t-V_{t+1}-\myavg)\,, \quad \hb_t=(R_t+\hQ_t-\hV_{t+1}-\myavg)\,.
\]
We first note that 
\begin{align*}
 \E\left[\omega_t \eta_t(R_t+V_{t+1}-Q_t)\right]&= \E\left[\omega_t \eta_t\E[(R_t+V_{t+1}-Q_t)|S_t,A_t]\right]\\
 &= \E\left[\omega_t \eta_t \right]\myavg\,.
\end{align*}
where the second equations follows Lemma \ref{lemma: bell-q} for the long-run density ratio function $\omega(.)$. This implies that $\E[c_tb_t]=0$.  
We next use the identity $(\widehat{c_t} \widehat{b_t}-c_tb_t)=(\widehat{c_t}-c_t)(\widehat{b_t}-b_t)+c_t(\widehat{b_t}-b_t)+(\widehat{c_t}-c_t)b_t$, using this brings us
\begin{align}
\E[\hc_t\hb_t]&=\E[\hc_t\hb_t-c_tb_t]\nonumber\\
&=\frac{1}{T}\sum_{t=1}^T \E\left[ \eta_t(\homg_t-\omega_t)(\hV_{t+1}- V_{t+1}+Q_t-\hQ_t)\right] \label{eq: bias-gamma-1-a}\\
&~\,+ \frac{1}{T}\sum_{t=1}^T \E\left[\eta_t \omega_t (Q_t- V_{t+1} -\hQ_t+\hV_{t+1})\right] \label{eq: bias-gamma-1-b}\\
&~\,+ \frac{1}{T}\sum_{t=1}^T\E[\eta_t(\hat{\omega}_t-\omega_t)(R_t+ V_{t+1}-Q_t-\myavg)]\label{eq: bias-gamma-1-c}\,.
\end{align}
We start with expression \eqref{eq: bias-gamma-1-a}, by an application of Jensen's inequality and policy overlap assumption we obtain
\begin{align}
\E\left[ \eta_t(\hat{\omega}_t-\omega_t)(\hV_{t+1}- V_{t+1}+Q_t-\hQ_t)\right] &\le C_\eta \E\Big[ (\hat{\omega}_t-\omega_t)^2 \Big]^{\frac{1}{2}} \E\Big[(\hV_{t+1}- V_{t+1}+Q_t-\hQ_t)^2\Big]^{\frac{1}{2}} \nonumber\\
&\le C_\eta K_T\Xi_T\,, \label{eq: worked-a}
\end{align}
where in the last relation we used Assumption \ref{assumption: model-estimates-gamma-1}.  For the term \eqref{eq: bias-gamma-1-b}, we have
\begin{align}
 \E\left[\eta_t \omega_t (Q_t- V_{t+1} -\hQ_t+\hV_{t+1})\right]&= \E\left[\eta_t \omega_t (Q_t-\hQ_t)\right]+  \E\left[\eta_t \omega_t (- V_{t+1}+\hV_{t+1})\right]\nonumber\\
 &= \E\left[\eta_t \omega_t (Q_t-\hQ_t)\right]+  \E\left[\omega_t (- V_{t}+\hV_{t})\right]\nonumber\\
 &=\E\left[ \omega_t (V_t-\hV_t)\right]+ \E\left[\omega_t (- V_{t}+\hV_{t})\right]=0\label{eq: worked-b}\,,
\end{align}
where we used Lemma \ref{lemma: bell-prob} for density ratio function $\omega(.)$.  Finally, we focus on \eqref{eq: bias-gamma-1-c} to obtain
\begin{align}
\E[\eta_t(\hat{\omega}_t-\omega_t)(R_t+ V_{t+1}-Q_t-\myavg)]&=\E[\eta_t(\hat{\omega}_t-\omega_t)\E[(R_t+ V_{t+1}-Q_t-\myavg)|S_t,A_t]]\nonumber \\
&=0\label{eq: worked-c} \,,
\end{align}
where we used Lemma \ref{lemma: bell-q} for $\omega(.)$. 
By combining \eqref{eq: worked-a}, \eqref{eq: worked-b}, and \eqref{eq: worked-c} we arrive at
\begin{equation}\label{eq: DR-gamma-1}
\left|\E[\hc_t\hb_t]\right| \le C_\eta K_T \Xi_T\,.
\end{equation}


\section{Proof of Lemma \ref{lemma: dis-tdr-var}}\label{proof: lemma: dis-tdr-var}
We let $\eta(s,a)=\frac{\pi_e(s,a)}{\pi_b(s,a)}$, and $\homg_t=\hat{\omega}^{\gamma}(S_t;p_0)$ (for the discounted estimation problem), $\eta_t=\eta(S_t,A_t)$, $\hq_t=\hq_e(S_t)$ and $\hv_t=\hat{v}^{\gamma}_e(S_t)$. Similarly, $v_t=v^{\gamma}_r(S_t)$, and $q_t=q^{\gamma}_e(S_t)$.
We first show that for $\delta\le 1$, given the finite value of $\E_{p_b}[\|\homg^{\gamma}(S;p_0)-\omega^{\gamma}(S;p_0)\|_2^2]$, the same tail inequality as in the weak distributional overlap can be established for $\homg^{\gamma}(.;p_0)$. In particular, using the triangle's inequality we have
\begin{align*}
\prob_{p_b}\left(\homg^{\gamma}(S;p_0)^{1+\delta}\ge x\right) &=\prob_{p_b}\left(\homg^{\gamma}(S;p_0)\ge x^{\frac{1}{1+\delta}}\right)  \\
&\le \prob_{p_b}\left(\omega^{\gamma}(S;p_0)\ge \frac{x^{\frac{1}{1+\delta}}}{2}\right) +  \prob_{p_b}\left(\Big|\homg^{\gamma}(S;p_0)-\omega^{\gamma}(S;p_0)\Big|\ge \frac{x^{\frac{1}{1+\delta}}}{2}\right)\,.
\end{align*}
We next use the $\delta$-weak distributional overlap assumption in along with Markov's inequality to obtain
\begin{align*}
\prob_{p_b}\left(\homg^{\gamma}(S;p_0)^{1+\delta}\ge x\right) &\le \frac{C 2^{1+\delta}}{x}+\frac{2^{1+\delta}\E\left[|\homg^{\gamma}(S;p_0)-\omega^{\gamma}(S;p_0)|^{1+\delta}\right]}{x}\,.
\end{align*} 
Given that we are in the setting of $\delta \le 1$, from Assumption \ref{assumption: model-estimates} we have $\E\left[|\homg^{\gamma}(S;p_0)-\omega^{\gamma}(S;p_0)|^{2}\right]<+\infty$, therefore $\E\left[|\homg^{\gamma}(S;p_0)-\omega^{\gamma}(S;p_0)|^{1+\delta}\right]<+\infty$. Using this in the above implies that there exists constant $C'$ with finite value such that 
\begin{align}\label{eq: mse-homg-tail}
\prob_{p_b}\left(\homg^{\gamma}(S;p_0)^{1+\delta}\ge x\right) \le \frac{C'}{x}\,.
\end{align}
Let $\lambda_t = \eta_t(\hat{\omega}_t \wedge \tau_t)(R_t + \gamma \hat{v}_{t+1} - \hat{q}_t)$. For the variance term, we have
\begin{align}
\var(\rhot_e)&=\var\left(\frac{1}{T}\sum_{t=1}^T \lambda_t  \right)\nonumber\\
&=\frac{1}{T^2}\sum_{t=1}^T \var(\lambda_t)+ \frac{2}{T^2}\sum\limits_{1\le \ell <t\le T}^{} \cov(\lambda_\ell,\lambda_t)\nonumber\\
&= \frac{1}{T^2}\sum_{t=1}^T \var(\lambda_t) + \frac{2}{T^2}\sum\limits_{\ell<t}^{} {\corr(\lambda_\ell,\lambda_t)}{\var(\lambda_\ell)^{1/2}\var(\lambda_t)^{1/2}}\nonumber\,,
\end{align}
where in the last relation we used the correlation function definition. Given that $\E[\lambda_t^2]\lesssim \tau_t^2$, so $\lambda_t,\lambda_{k+t}$ belong to the space of square integrable functions and we have $\corr(\lambda_\ell, \lambda_t)\le \rho_{|t-\ell|}$, based on the $\rho$-mixing coefficient $\rho_n$. We use this to arrive at
\begin{align}
\var(\rhot_e)&\leq \frac{1}{T^2}\sum_{t=1}^T \var(\lambda_t) + \max\limits_{\ell<t}\left\{\var(\lambda_\ell)^{1/2}\var(\lambda_t)^{1/2}\right\} \frac{2}{T^2}\sum\limits_{\ell<t}^{} {\corr(\lambda_\ell,\lambda_t)}\nonumber \\
&\leq  \frac{1}{T^2}\sum_{t=1}^T \var(\lambda_t) + \max\limits_{\ell<t}\left\{\var(\lambda_\ell)^{1/2}\var(\lambda_t)^{1/2}\right\} \frac{2}{T^2}\sum\limits_{\ell<t}^{} \rho_{t-\ell}\nonumber \\
&\le \frac{1}{T^2}\sum_{t=1}^T \var(\lambda_t) +\frac{2}{T} \max\limits_{\ell<t}\left\{\var(\lambda_\ell)^{1/2}\var(\lambda_t)^{1/2}\right\} \sum_{m=1}^{T}\rho_m \nonumber \\
&\le  \frac{1}{T^2}\sum_{t=1}^T \var(\lambda_t) +\frac{2C_\rho}{T} \max\limits_{\ell<t}\left\{\var(\lambda_\ell)^{1/2}\var(\lambda_t)^{1/2}\right\}\,.
\label{eq: tmp-var-upper}
\end{align}
The last inequality follows $\rho$-mixing Assumption \ref{assu:rho-mix}. From \eqref{eq: tmp-var-upper} we need to upper bound $\var(\lambda_t)$,  we use $\var(\lambda_t)\le \E[\lambda_t^2]$. Specifically,  we have $\var(\lambda_t)\leq C_1 \E[(\homg_t\wedge \tau_t)^2]$, where we used the fact that $\eta_t$ and reward values $R_t$ are almost surely bounded, we consider the absolute constant $C_1$ such that  $\eta_t(R_t+\gamma v_{t+1}-q_{t}) \le C_1$.  Put all together, we focus on upper bounding $\E[(\homg_t\wedge \tau_t)^2]$.

For the case $\delta>1$, from the distributional overlap assumption we get that $\E[\omega_t^2]<\infty$, and then using triangle's inequality yields $\E[\homg_t^2]<+\infty$. Thereby, when $\delta>1$,  there exists $C'_V<+\infty$ such that $\var(\lambda_t)\le C'_V$, using this in \eqref{eq: tmp-var-upper} gives us that for $\delta>1$ we have $\var(\mytdrdisavg)\lesssim \frac{1}{T}$. 

We now focus on the case when $\delta\le 1$. By rewriting this expectation in terms of probability density functions we get

\begin{equation}\label{eq: var-tmp}
 \E\Big [(\homg_t\wedge \tau_t)^2\Big]=\int\limits_{0}^{\tau_t} 2x \prob(\homg_t\geq x)\de x\,.
\end{equation}
We next choose $\beta_t$ such that  $\beta_t<\tau_t$, we arrive at
\begin{align}
\int\limits_{0}^{\tau_t} 2x \prob(\homg_t\ge x)&= \int\limits_{0}^{\beta_t} 2x \prob(\homg_t\ge x)\de x+ \int\limits_{\beta_t}^{\tau_t} 2x \prob(\homg_t \ge x)\de x  \nonumber \\
&\leq \beta_t^2+ \int\limits_{\beta_t}^{\tau_t} 2x \prob(\homg_t\ge x)\de x \nonumber \,.
\end{align}
We next employ the tail bound on $\homg$ given in \eqref{eq: mse-homg-tail} to obtain
\begin{align}
\int\limits_{0}^{\tau_t} 2x \prob(\homg_t\ge x)& \leq \beta_t^2+ 
\int\limits_{\beta_t}^{\tau_t}  2x\frac{C'}{x^{1+\delta}}\nonumber\\
&=  \beta_t^2+ 2C' \int \limits_{\beta_t}^{\tau_t} \frac{1} {x^\delta}  \de x\,. \label{eq: general-delta}
\end{align}
 In the next step, by using \eqref{eq: general-delta} we get
 \begin{equation}\label{eq: tmp-upper-integration}
\int\limits_{0}^{\tau_t} 2x \prob(\homg_t\ge x)  \le 
\begin{cases}
\beta_t^2+2C'\frac{\tau_t^{1-\delta}}{1-\delta} & \delta<1\,,\\
\beta_t^2+ 2C'\log(\tau_t/\beta_t)  & \delta=1\,.
\end{cases}
\end{equation}
By an appropriate choice of $\beta_t$, specifically  $\beta_t=\tau_t^{\frac{1-\delta}{2}}$ for $\delta<1$ and $\beta_t=\log(\tau_t)^{\frac{1}{2}}$  in \eqref{eq: tmp-upper-integration}, and then combining \eqref{eq: var-tmp} and \eqref{eq: tmp-upper-integration} we arrive at
\begin{equation}\label{eq: var-truncated}
\var(\lambda_t) \lesssim 
\begin{cases}
\tau_t^{1-\delta} & \delta<1\,,\\
\log(\tau_t) & \delta=1\,.
\end{cases}
\end{equation}
We suppose that $\tau_t$ is an non-decreasing sequence in $t$. Using \eqref{eq: var-truncated} in \eqref{eq: tmp-var-upper} gives us 
\begin{equation}\label{eq: var-tdr}
\var(\rhot_e)\lesssim  
\begin{cases}
\frac{1}{T^2}\sum\limits_{t=1}^T\tau_{t}^{1-\delta}+\frac{1}{T}\tau_T^{1-\delta} \,,& \delta<1 \,, \\ 
\frac{1}{T^2}\sum\limits_{t=1}^{T}\log(\tau_t)+ \frac{1}{T}\log \tau_T\,,& \delta=1\,. 
\end{cases}
\end{equation}

\section{Proof of Lemma \ref{lemma: dis-tdr-bias}}\label{proof: lemma: dis-tdr-bias}
We follow the same shorthands introduced in Section \ref{proof: lemma: dis-tdr-var}. We focus on upper bounding the bias term. This gives us

\begin{align}
\E[\rhot_e]-\mydisavg&=(1-\gamma)\E_{p_0}[\hv(S)]+\frac{1}{T}\sum\limits_{t=1}^T  \E[\eta_t (\homg_t\wedge \tau_t)(R_t+\gamma \hv_{t+1}-\hq_t)]-(1-\gamma)\E_{p_0}[v(S)]\nonumber\\
&=\frac{1}{T}\sum\limits_{t=1}^{T}\E[ ((\homg_t\wedge \tau_t)-\homg_t)\eta_t(R_t+\gamma \hv_{t+1}-\hq_t)]\nonumber\\
&\, + (1-\gamma)\E_{p_0}[\hv(S)]+\frac{1}{T}\sum\limits_{t=1}^T  \E[\eta_t \homg_t(R_t+\gamma \hv_{t+1}-\hq_t)]-(1-\gamma)\E_{p_0}[v(s)]\nonumber\\
&=  \frac{1}{T}\sum\limits_{t=1}^{T}\E[ ((\homg_t\wedge \tau_t)-\homg_t)\eta_t(R_t+\gamma \hv_{t+1}-\hq_t)+ \hat{\th}^{\dr}_{\gamma}(p_0)-\mydisavg \label{eq: second-exp-tmp}\,.
\end{align}

We then utilize the inequality $\eta_t (R_t + \gamma \hat{v}_{t+1} - \hat{q}_t) \le C_1$ in conjunction with the doubly robustness Lemma \ref{lemma-second-little-o} to establish an upper bound for the second expression given in \eqref{eq: second-exp-tmp}. This result provides us with
\begin{equation}\label{eq: tmp-bias-upper-bound}
|\E[\rhot_e]-\mydisavg|\lesssim \kappa_T\xi_T +  \frac{1}{T}\sum\limits_{t=1}^T \E[\homg_t-\homg_t\wedge \tau_t]\,.
\end{equation}
We next provide an upper bound on $\E[\homg_t-\homg_t\wedge \tau_t]$. For the case when $\delta>1$, using the finite value of $\E[\homg_t^2]$, we arrive at

\begin{align*}
\E[\homg_t-\homg_t\wedge \tau_t]&=\int\limits_{\tau_t}^{\infty}\prob(\homg_t\geq u)\de u\\
&\le \int\limits_{\tau_t}^{\infty}\frac{C'}{u^2} \de u\,.
\end{align*}
This implies that when $\delta>1$, we have
\begin{equation}\label{eq: bias-bound-tmp-2}
\E[\homg_t-\homg_t\wedge \tau_t] \lesssim \frac{1}{\tau_t}\,,
\end{equation}
In addition, when $\delta \le 1$, using the tail inequality of $\homg$ given in \eqref{eq: mse-homg-tail} we obtain:
\begin{align*}
\E[\homg_t-\homg_t\wedge \tau_t]&=\int\limits_{\tau_t}^{\infty}\prob(\homg_t\geq u)\de u\\
&\le \int\limits_{\tau_t}^{\infty}\frac{C'}{u^{1+\delta}} \de u\,.
\end{align*}
This brings us
\begin{align}\label{eq: bias-bound-tmp}
\E[\homg_t-\homg_t\wedge \tau_t]&\le \frac{C'}{\delta\tau_t^\delta}\,.
\end{align}
By employing \eqref{eq: bias-bound-tmp} and \eqref{eq: bias-bound-tmp-2} in \eqref{eq: tmp-bias-upper-bound} and utilizing the identity $(a+b)^2\le 2(a^2+b^2)$, we obtain the following result:
 \begin{equation}\label{eq: bias}
\Big|\E[\rhot_e]-\mydisavg\Big|^2 \lesssim
\begin{cases}
\kappa_T^2\xi_T^2  + \frac{1}{T^2} \left(\sum\limits_{t=1}^T \frac{1}{\tau_t^{\delta}}\right)^2\,, & \delta \le 1\,,\\
\kappa_T^2\xi_T^2  + \frac{1}{T^2} \left(\sum\limits_{t=1}^T \frac{1}{\tau_t}\right)^2\,, & \delta>1\,.
\end{cases}
\end{equation}




\section{Proof of Lemma \ref{lemma: martingale-o-one}}\label{proof: lemma: martingale-o-one}
We follow the same shorthands introduced in Section \ref{proof: lemma: dis-tdr-var}.
We start the proof of Theorem \ref{thm: truncated} by expanding $\rhot-\mydisavg$:

\begin{align}
\sqrt{T}(\rhot-\mydisavg)&=\sqrt{T}\left((1-\gamma) \E_{S\sim p_0}\Big[\hv(S)\Big]+\frac{1}{T}\sum_{t=1}^T  (\homg_t \wedge \tau_t)\eta_t(R_t+\gamma \hv_{t+1}-\hq_t) - {\mydisavg}\right)\nonumber\\
&=\sqrt{T}\left(\frac{1}{T}\sum_{t=1}^T \Big[ (\homg_t\wedge \tau_t)\eta_t(R_t+\gamma \hv_{t+1}-\hq_t)- \homg_t\eta_t(R_t+\gamma \hv_{t+1}-\hq_t)\Big]     \right) \label{eq: term-1}\\
&\,~+ \sqrt{T}\left(  \frac{1}{T}\sum_{t=1}^T \Big[\widehat{\psi}_t-\psi_t\Big]-(\hat{\th}_\gamma^{\dr}(p_0)-\mydisavg)   \right) \label{eq: term-2}\\
&\,~+\sqrt{T}(\hat{\th}_\gamma^{\dr}(p_0)-\mydisavg) \label{eq: term-3}\\
&\,~+ \sqrt{T}\left( \frac{1}{T} \sum_{t=1}^T \omega_t\eta_t(R_t+\gamma v_{t+1}-q_t) - (\omega_t\wedge \tau_t)\eta_t(R_t+\gamma v_{t+1}-q_t)   \right) \label{eq: term-4}\\
&\,~+ \frac{1}{\sqrt{T}}\sum_{t=1}^T (\omega_t\wedge \tau_t)\eta_t(R_t+\gamma v_{t+1}-q_t) \,. \label{eq: term-5}
\end{align}

By employing Lemmas \ref{lemma-first-little-o} and \ref{lemma-second-little-o} we get that terms \eqref{eq: term-2} and \eqref{eq: term-3} are $o_P(1)$, respectively.  For the rest, we also show that terms \eqref{eq: term-4} and \eqref{eq: term-1} are also $o_P(1)$. Define
\begin{equation}
\beta\overset{\Delta}{=}\delta-1-\frac{1}{2}\cdot \min\left\{ \delta-1,\frac{1+\delta}{2}-\alpha,\alpha-\frac{2}{1+\delta}\right\}\,.
\end{equation}
It is easy to check that $\delta-1>\beta>0$ and $\frac{2}{2+\beta}< \alpha < \frac{2+\beta}{2}$, given that $\frac{2}{1+\delta}< \alpha < \frac{1+\delta}{2}$. In addition, since $2+\beta<1+\delta$ and from $\delta$-weak distributional overlap assumption we have $\prob\big(\omega^{\gamma}(S;p_0)^{1+\delta}>x\big)\lesssim \frac{1}{x}$ and model estimate assumption $\prob\big(\hat{\omega}^{\gamma}(S;p_0)^{1+\delta}>x\big)\lesssim \frac{1}{x}$, we get 
\begin{equation}\label{eq: E_2_beta}
\E[{\omega}^{\gamma}(S;p_0)^{2+\beta}]<+\infty\,,\quad \E[\hat{\omega}^{\gamma}(S;p_0)^{2+\beta}]<+\infty\,. 
\end{equation}
We first show that \eqref{eq: term-4} is $o_P(1)$.  To show this, let 
\[
z_t=\omega_t\eta_t(R_t+\gamma v_{t+1}-q_t)\,, \quad z'_t=(\omega_t\wedge \tau_t)\eta_t(R_t+\gamma v_{t+1}-q_t)\,.
\]
We will show that $\frac{1}{\sqrt{T}}\sum_{t=1}^T (z_t-z'_t)=0$, almost surely. For this end, first not that we have 
\begin{align}\label{eq: Borel-condition}
\sum_{t=1}^\infty \prob(z_t\neq z'_t)&=\sum_{t=1}^\infty \prob(\omega_t > \tau_t)\nonumber\\
&\le \sum_{t=1}^\infty \frac{\E\Big[\omega_t^{2+\beta}\Big]}{\tau_t^{2+\beta}}\nonumber\\
&=\E\Big[\omega_t^{2+\beta}\Big] \sum_{t=1}^{+\infty}\frac{1}{t^{\alpha(2+\beta)}}<+\infty\,,
\end{align}

where in the last relation we used \eqref{eq: E_2_beta} in along with $\alpha (2+\beta)>1$. In the next step, using \eqref{eq: Borel-condition} in along with Borel–Cantelli lemma we arrive at
\[
\prob\left(\lim\inf_{t\to \infty }\left \{s\in \cS: z_t(s)=z'_t(s)\right\}\right)=1\,.
\]
This brings us that if we define  $B=\bigcup\limits_{t=1}^{\infty} \bigcap\limits_{r\geq t}^{\infty} \{s\in \cS: z_r(s)=z'_r(s)\}$, then $\prob(B)=1$.  Concretely, for every  $s\in B$, there exists an integer $\Lambda(s)$ such that if $t\geq \Lambda(s)$ then $Z_t(s)=Z'_t(s)$. This implies that for each $s\in B $ the following holds
 \begin{align*}
 \lim\sup_{T\to \infty} \frac{1}{\sqrt{T}}\sum_{t=1}^T (Z_t(s)-Z'_t(s))&= \lim\sup_{T\to \infty} \frac{1}{\sqrt{T}}\sum_{t=1}^{\Lambda(s)} (Z_t(s)-Z'_t(s))=0\,.
 \end{align*}

Given that $\prob(B)=1$, this gives us the almost sure convergence of $\frac{1}{\sqrt{T}}\sum_{t=1}^T (Z_t-Z'_t)$ to zero, and therefore \eqref{eq: term-4} is $o_P(1)$.


In addition, given that for model estimate $\homg$ we have $\E[\homg_t^{2+\beta}]<\infty$, by a similar argument used above for \eqref{eq: term-4}, it can be shown that expression $\eqref{eq: term-1}$ is also $o_P(1)$. This completes the proof.

\section{Proof of Lemma \ref{lemma: Lindeberg}}\label{proof: lemma: Lindeberg}

We follow the same shorthands introduced in Section \ref{proof: lemma: dis-tdr-var}. For the filtration $\cF_{t}$ generated by $\{(S_\ell,A_\ell,R_\ell)\}_{\ell\leq {t+1}}$ we show that $G_{t,T}$ is a triangular martingale array with respect to $\cF_t$. As a first step, it is easy to observe that $G_{t,T}$ is measurable with respect to $\cF_t$ (adaptivity). We next show that $\E[G_{t,T}|\cF_{t-1}]=G_{t-1,T}$ and establish the martingale property:
\begin{align*}
\E[G_{t,T}|\cF_{t-1}]&=G_{t-1,T}+\frac{1}{\sqrt{T}} \E[(\omega_t \wedge \tau_t)\eta_t(R_t+\gamma v_{t+1}-q_t)|\cF_{t-1}]\\
&=G_{t-1,T}+\frac{1}{\sqrt{T}}\E\Big[\E[(\omega_t \wedge \tau_t)\eta_t(R_t+\gamma v_{t+1}-q_t)|S_t,A_t]\Big]\\
&=G_{t-1,T}+\frac{1}{\sqrt{T}} \E[(\omega_t \wedge \tau_t)\E[\eta_t(R_t+\gamma v_{t+1}-q_t)|S_t,A_t]]\\
&=G_{t-1,T}+\frac{1}{\sqrt{T}}\E\left[(\omega_t \wedge \tau_t) \E_{\pi_e}[(R_t+\gamma v_{t+1}-q_t)|S_t,A_t]\right]\\
&=G_{t-1,T}\,,
\end{align*}
where the last relation follows Lemma \ref{lemma: bell-q}. 
Let $X_{\ell,T}=\frac{1}{\sqrt{T}}(\omega_\ell\wedge \tau_\ell)\eta_\ell(r_\ell+\gamma v_{\ell+1}-q_\ell)$, so $G_{t,T}=\sum_{\ell\le t }X_{\ell,T}$. We next show that $\{G_{t,T}\}$ is square-integrable.  We have

\begin{align*}
\E[G_{t,T}^2]=\sum_{\ell\le t}\E[X_{\ell,T}^2] +\sum_{\ell_1\neq \ell_2}\E[X_{\ell_1,T}X_{\ell_2,T}]\,.
\end{align*}
For $\ell_1<\ell_2$, again by using a similar argument used above with Lemma \ref{lemma: bell-q} we get $\E[X_{\ell_1,T}X_{\ell_2,T}]=\E[X_{\ell_1,T}\E[X_{\ell_2,T}|\cF_{\ell_2-1}]]=0$. This give us
\begin{align*}
\E[G_{t,T}^2]&=\sum_{\ell\le t}\E[X_{\ell,T}^2]\\
&\le \frac{t}{T}\E[\omega_t^2]\\
& \le \sigma_b^2 <+\infty.  
\end{align*}

So far, we have shown that $G_{t,T}$ is a centered square-integrable triangular Martingale array. We then use \citep{hall2014martingale}, Corollary 3.1 to establish the CLT. For this purpose, we first show that the Lindeberg's condition holds. Formally, for every $\eps>0$ we have to prove
\begin{equation}\label{eq: lindeberg's conditon}
\sum_{t=1}^{T}\E[X_{t,T}^2\ind(|X_{t,T}|>\eps)|\cF_{t-1}]\overset{(p)}{\to} 0.
\end{equation}

This means that for every $c>0$ we must have
\begin{equation}\label{eq: Lind}
\lim_{T\to \infty} \prob\left(\sum_{t=1}^{T}\E[X_{t,T}^2\ind(|X_{t,T}|>\eps)|\cF_{t-1}]>c \right)=0
\end{equation}

We start by using Markov's inequality to obtain
\begin{align*}
 \prob\left(\sum_{t=1}^{T}\E[X_{t,T}^2\ind(|X_{t,T}|>\eps)|\cF_{t-1}]>c \right)&\leq \frac{\E\left[\sum_{t=1}^T X_{t,T}^2 \ind(|X_{t,T}|> \eps) \right]}{c}\\
 &\leq \frac{1}{c}\E\left[\max_{1\le t\le T} |X_{t,T}|^2 \sum_{t=1}^T \ind(|X_{t,T}|>\eps)\right]\,,
 \end{align*}
 We next use $X_t\le \tau_t C_\eta \frac{R_{\max}}{\sqrt{T}(1-\gamma)}$ to get
 \begin{align}\label{eq: linden-1}
 \prob\left(\sum_{t=1}^{T}\E\Big[X_{t,T}^2\ind(|X_{t,T}|>\eps)|\cF_{t-1}\Big]>c \right) &\lesssim \frac{\max\limits_{1\le t\le T}^{}{\tau_t}^2}{c T}\sum_{t=1}^{T}\prob(|X_{t,T}|>\eps)\,.
 \end{align}
 We next use $X_{t,T}\lesssim \frac{\omega_t}{\sqrt{T}}$ and by an application of Markov's inequality we get 
 \[
 \prob(|X_{t,T}|\ge \eps)\lesssim \frac{\E_{p_b}[\omega^{\gamma}(S;p_0)^{2+\beta}]}{\eps\, T^{\frac{2+\beta}{2}}}\,.
 \]
 Using this in \eqref{eq: linden-1} yields
 \begin{align*}
\prob\left(\sum_{t=1}^{T}\E[X_{t,T}^2\ind(|X_{t,T}|>\eps)|\cF_{t-1}]>c\right) & \lesssim \Big({\max\limits_{1\le t\le T}^{}{\tau_t}^2}\Big) \frac{\E_{p_b}[\omega^{\gamma}(S;p_0)^{2+\beta}]}{c\eps\, T^{1+\frac{\beta}{2}}}\,.
\end{align*}
Having $\tau_t=t^{\frac{\alpha}{2}}$ give us $\frac{\max\limits_{1\le t\le T}{\tau_t}^2}{T^{1+\frac{\beta}{2}}}=T^{\alpha- \frac{2+\beta}{2} }$. Using this in the above, in along with  the fact that $\alpha< \frac{2+\beta}{2}$ brings us
\[
\lim_{T\to \infty}\prob\left(\sum_{t=1}^{T}\E[X_{t,T}^2\ind(|X_{t,T}|>\eps)|\cF_{t-1}]>c\right)=0\,.
\]
This completes the proof for \eqref{eq: Lind}.
\section{Proof of Lemma \ref{lemma: var-convergence}}\label{proof: lemma: var-convergence}
We follow the same shorthands introduced in Section \ref{proof: lemma: dis-tdr-var}. We want to show that
 \begin{equation}\label{eq: var-stability}
 \sum_{t=1}^{T} X_{t,T}^2\overset{(p)}{\to} \sigma_b^2\,.
 \end{equation}
 We have
 \begin{align*}
 \E\Big[\omega_t^2-(\omega_t\wedge \tau_t)^2\Big]= \int\limits_{\tau_t}^{+\infty} 2x \prob(\omega_t\ge x)\de x\,.
 \end{align*}
 Using Markov's inequality we get
  \begin{align*}
   \E\Big[\omega_t^2-(\omega_t\wedge \tau_t)^2\Big]& \le \int\limits_{\tau_t}^{+\infty} \frac{2\E_{p_b}[\omega^{\gamma}(S;p_0)^{2+\beta}]}{x^{1+\beta}}\\
   &=  \frac{2\E_{p_b}[\omega^{\gamma}(S;p_0)^{2+\beta}]}{\tau_t^{\beta}}\,.
  \end{align*}
  Using $\tau_t=t^{\alpha/2}$ in along with \eqref{eq: E_2_beta} we arrive at
  \begin{align}
 \frac{1}{T}\sum_{t=1}^T \E\Big[\omega_t^2-(\omega_t\wedge \tau_t)^2\Big] &\lesssim  \frac{1}{T} \E_{p_b}\Big[\omega^{\gamma}(S;p_0)^{2+\beta}\Big]\sum_{t=1}^T t^{-\frac{1}{2}\alpha \beta}\nonumber\\
 &\lesssim T^{-\frac{1}{2}\beta\alpha}\label{eq: lind-4}\,.
 \end{align}
 This implies that 
 \begin{align}
 \left|\sum_{t=1}^T\E\Big[X_{t,T}^2\Big]-\sigma_b^2\right|&=\left|\frac{1}{T}\sum_{t=1}^T\E\left[\big(\omega_t^2\wedge \tau_t^2-\omega_t^2\big)\eta_t^2(R_t+\gamma v_{t+1}-q_{t})^2\right]\right|\nonumber\\
 &\lesssim \frac{1}{T}\sum_{t=1}^T\E\left[\big(\omega_t^2-\omega_t^2\wedge \tau_t^2\big) \right]\label{eq: lind-5}\,.
 \end{align}
 Finally, by employing \eqref{eq: lind-4} in \eqref{eq: lind-5} we arrive at
 \[
\lim\limits_{T\to \infty}^{} \sum_{t=1}^T\E\Big[X_{t,T}^2\Big]=\sigma_b^2\,.
\]
This proves \eqref{eq: var-stability} and completes the proof.





\section{Proof of Proposition \ref{prop-unif-bound}}

 We want to provide an upper bound for the function $Q_e(s,a)$, that holds for all states $s$ and action values $a$. We suppose that under the evaluation policy $\pi_e$, in the stationary regime, the reward density distribution is $q_e(.)$. From the ergodicity theorem for Markov chains, we know that $\E_{q_e}[R]=\myavg$.  We then focus on a class of reward distributions $\{q_{k}(.|S_0=s,A_0=a)\}_{k\ge 0}$, where $q_k$ denote the distribution on reward value after $k$ steps under Markov dynamics when we start from the initial state $S_0=s$ and the action $A_0=a$ is taken.  Given that the chain has a geometric mixing property with parameter $t_0$, we have $d_{\tv}(q_k,q) \le \exp\left(-\frac{k}{t_0}\right)$. This gives us
\begin{align*}
\left|\E[R_k|S_0=s,A_0=a]-\myavg\right|&=\left|\int u \left(q_k(u|S_0=s,A_0=a)-q_e(u)\right)\de \lambda_r(u)\right|\\
&\le \int u\left| q_k(u|S_0=s,A_0=a)-q_e(u)\right|\de \lambda_r(u)\\
&\le R_{\max}\cdot d_{\tv}(q_k,q_e)\\
&\le R_{\max}\exp\left(-\frac{k}{t_0}\right)\,.
\end{align*}
In the next step, we consider the definition of $Q(s,a)$ function to obtain
\begin{align*}
Q_e(s,a)&=\E\left[\sum\limits_{k=0}^{\infty} (R_k-\myavg)|s_0=s,a_0=a\right]\nonumber\\
&\le R_{\max}\sum\limits_{k\ge 0}^{}\exp\left(-\frac{k}{t_0}\right)<+\infty \,.
\end{align*}
If we define 
$C_Q=R_{\max}\sum\limits_{k\ge 0}^{}\exp\left(-\frac{k}{t_0}\right),$ then for every $(s,a)$ value we have $Q(s,a)\le C_Q$. Accordingly, given that $V_e(s)=\E_{\pi_e}[Q_e(s,a)|s]$, we realize that the obtained bound works for function $V_e(.)$ as well. 


\section{Proof of Theorem \ref{thm: mse-gamma-1}}\label{proof: thm: mse-gamma-1}
We let $\eta(s,a)=\frac{\pi_e(s,a)}{\pi_b(s,a)}$. We use shorthands $\eta_t=\eta(S_t,A_t)$, $W_t=\omega(S_t)$, $\hW_t=\homg(S_t)$, $V_t=V_e(S_t)$, $\hV_t=\hat{V}_e(S_t)$, $Q_t=Q_e(S_t)$, and $\hQ_t=\hat{Q}_e(S_t)$. 
{
We first show that when $\delta\le 1$ given that the value $\E_{p_b}[\|\homg(S)-\omega(S)\|_2^2]$ is finite, the same tail inequality as in $\delta$-weak distributional overlap condition can be established for $\homg(.)$. In particular, using the triangle's inequality we have
\begin{align*}
\prob_{p_b}\left(\homg(S)^{1+\delta}\ge x\right) &=\prob_{p_b}\left(\homg(S)\ge x^{\frac{1}{1+\delta}}\right)  \\
&\le \prob_{p_b}\left(\omega(S)\ge \frac{x^{\frac{1}{1+\delta}}}{2}\right) +  \prob_{p_b}\left(\Big|\homg(S)-\omega(S)\Big|\ge \frac{x^{\frac{1}{1+\delta}}}{2}\right)\,.
\end{align*}
We next use the $\delta$-weak distributional overlap assumption in along with the Markov's inequality to obtain
\begin{align*}
\prob_{p_b}\left(\homg(S)^{1+\delta}\ge x\right) &\le \frac{C 2^{1+\delta}}{x}+\frac{2^{1+\delta}\E\left[|\homg(S)-\omega(S)|^{1+\delta}\right]}{x}\,.
\end{align*} 
Given that we are in the setting of $\delta \le 1$, from Assumption \ref{assumption: model-estimates} we have $\E\left[|\homg(S)-\omega(S)|^{2}\right]<+\infty$, therefore $\E\left[|\homg(S)-\omega(S)|^{1+\delta}\right]<+\infty$. Using this in the above implies that there exists constant $C'$ with finite value such that 
\begin{align}\label{eq: mse-homg-tail-gamma-1}
\prob_{p_b}\left(\homg(S)^{1+\delta}\ge x\right) \le \frac{C'}{x}\,, \quad \text{ when } \delta \le 1\,.
\end{align}
In the next step, by rewriting the $\tdr$ estimator we obtain 
\[
\mytdravg-\myavg=\frac{\frac{1}{T}\sum\limits_{t=1}^T (\hW_t \wedge \tau_t)\eta_t(R_t+\hV_{t+1}-\hQ_t-\myavg) }{\frac{1}{T}\sum\limits_{t=1}^T (\hW_t \wedge \tau_t)\eta_t }\,.
\]

From the policy overlap assumption and Proposition \ref{prop-unif-bound}, we get that 
\begin{equation}\label{eq: boundedness-lambda}
\eta_t(R_t+\hV_{t+1}-\hQ_t-\myavg)\le B\overset{\Delta}{=}C_\eta(R_{\max}+ 3C_Q)
\end{equation}
In the next step, by considering the self-normalization term in the denominator, we realize that $|\mytdravg-\myavg|\le B$. This gives us
\begin{equation}\label{eq: tdr-prob}
\E\left[(\mytdravg-\myavg)^2\right]=\int\limits_{0}^{B} 2x\prob\left(|\mytdravg-\myavg| \ge x \right) \de x\,.
\end{equation}

In addition, for every positive constant $M$ and $\eps \in (0,M)$ and non-negative random variables $(U,V)$ we have $\prob\left(\frac{U}{V}\ge x\right)\le \prob\left(\frac{U}{M-\eps}\ge x\right)+\prob\left(|V-M|\ge \eps\right).$ This hold because if $\frac{U}{V}\ge x$ then either $\frac{U}{M-\eps}\ge x$ or $|V-M|\ge \eps$. In fact, if none of these two events hold, we get
$\frac{U}{V}<\frac{U}{M-\eps} <x\,,$ which contradicts $\frac{U}{V}\ge x$. 
Using this for \eqref{eq:  tdr-prob} gives us
\begin{align}
\E\left[(\mytdravg-\myavg)^2\right]&=\int\limits_{0}^{B} 2x\prob\left(\Big|\frac{1}{T}\sum\limits_{t=1}^T (\hW_t \wedge \tau_t)\eta_t(R_t+\hV_{t+1}-\hQ_t-\myavg)\Big|  \ge x(M-\eps) \right) \de x\label{eq: big-tdr-1}\\
&\,~+ \int\limits_{0}^{B} 2x\prob\left( \Big|\frac{1}{T}\sum\limits_{t=1}^T (\hW_t \wedge \tau_t)\eta_t-a\Big| \ge \eps \right) \de x \label{eq: big-tdr-2}\,.
\end{align}

We start by focusing on \eqref{eq: big-tdr-1}. By change of variables in the integration given in \eqref{eq: big-tdr-1} for $x'=x(M-\eps)$ we get that \eqref{eq: big-tdr-1}  is upper bounded by the following 
\begin{align*}
\Delta_1&=\frac{1}{(M-\eps)^2}\E\left[ \Big(\frac{1}{T}\sum\limits_{t=1}^{T} \hW_t \eta_t(R_t+\hV_{t+1}-\hQ_t-\myavg)\Big)^2 \right]\,.
\end{align*}

Let $\lambda_t = \eta_t(\hW_t \wedge \tau_t)(R_t + \hV_{t+1} - \hQ_t-\myavg)$, we then have
\begin{align}\label{eq: mse-numerator}
(M-\eps)^2\Delta_1&=\E\left[ \Big(\frac{1}{T}\sum\limits_{t=1}^{T} \lambda_t\Big)^2  \right]\nonumber \\
&=\var\Big(\frac{1}{T}\sum\limits_{t=1}^{T} \lambda_t\Big)+\Big(\sum\limits_{t=1}^{T}\frac{1}{T}\E[\lambda_t]\Big)^2\,.
\end{align}

 We first focus on the variance term. For this, we have
\begin{align}
\var\left(\frac{1}{T}\sum_{t=1}^T \lambda_t  \right)&=\frac{1}{T^2}\sum_{t=1}^T \var(\lambda_t)+ \frac{2}{T^2}\sum\limits_{1\le \ell <t\le T}^{} \cov(\lambda_\ell,\lambda_t)\nonumber\\
&= \frac{1}{T^2}\sum_{t=1}^T \var(\lambda_t) + \frac{2}{T^2}\sum\limits_{\ell<t}^{} {\corr(\lambda_\ell,\lambda_t)}{\var(\lambda_\ell)^{1/2}\var(\lambda_t)^{1/2}}\nonumber\,,
\end{align}
where in the last relation we used the correlation function definition. Given that $\E[\lambda_t^2]\lesssim \tau_t^2$, so $\lambda_t,\lambda_{k+t}$ belong to the space of square integrable functions and we have $\corr(\lambda_\ell, \lambda_t)\le \rho_{|t-\ell|}$, following the definition of $\rho$-mixing coefficient $\rho_n$,  we arrive at
\begin{align}
\var\left(\frac{1}{T}\sum_{t=1}^T \lambda_t  \right)&\leq \frac{1}{T^2}\sum_{t=1}^T \var(\lambda_t) + \max\limits_{\ell<t}\left\{\var(\lambda_\ell)^{1/2}\var(\lambda_t)^{1/2}\right\} \frac{2}{T^2}\sum\limits_{\ell<t}^{} {\corr(\lambda_\ell,\lambda_t)}\nonumber \\
&\leq  \frac{1}{T^2}\sum_{t=1}^T \var(\lambda_t) + \max\limits_{\ell<t}\left\{\var(\lambda_\ell)^{1/2}\var(\lambda_t)^{1/2}\right\} \frac{2}{T^2}\sum\limits_{\ell<t}^{} \rho_{t-\ell}\nonumber \\
&\le \frac{1}{T^2}\sum_{t=1}^T \var(\lambda_t) +\frac{2}{T} \max\limits_{\ell<t}\left\{\var(\lambda_\ell)^{1/2}\var(\lambda_t)^{1/2}\right\} \sum_{m=1}^{T}\rho_m \nonumber \\
&\le  \frac{1}{T^2}\sum_{t=1}^T \var(\lambda_t) +\frac{2C_\rho}{T} \max\limits_{\ell<t}\left\{\var(\lambda_\ell)^{1/2}\var(\lambda_t)^{1/2}\right\}\,.
\label{eq: tmp-var-upper-gamma-1}
\end{align}
The last inequality follows $\rho$-mixing Assumption \ref{assu:rho-mix}. From \eqref{eq: tmp-var-upper-gamma-1} we need to upper bound $\var(\lambda_t)$,  which we use $\var(\lambda_t)\le \E[\lambda_t^2]$. Specifically,  we have $\var(\lambda_t)\le B \E[(\hW_t\wedge \tau_t)^2]$, where we used \eqref{eq: boundedness-lambda}.  Put all together, we focus on upper bounding $\E[(\hW_t\wedge \tau_t)^2]$.

We start by considering the $\delta>1$ case. In this case, given that $\delta>1$, we have $\E_{p_b}[\omega(S)^2]$ is bounded, therefore by using triangle's inequality for estimate Assumption \ref{assumption: model-estimates-gamma-1} we get $\E_{p_b}[\hat{\omega}(S)^2]<+\infty$. This implies that in this case there exists a finite value constant $C_V$ such that $\E[(\hW_t\wedge \tau_t)^2]\le C_V$. Using this in \eqref{eq: tmp-var-upper-gamma-1} yields 
\begin{equation}\label{eq: long-run-var-delta_large}
\var\left(\frac{1}{T}\sum\limits_{t=1}^{T}\lambda_t\right) \lesssim \frac{1}{T}\,, \quad \text{ when } \delta>1\,.
\end{equation}

By rewriting this expectation in terms of probability density functions we get
\begin{equation}\label{eq: var-tmp-gamma-1}
 \E\Big [(\hW_t\wedge \tau_t)^2\Big]=\int\limits_{0}^{\tau_t} 2x \prob(\hW_t\geq x)\de x\,.
\end{equation}
We next choose $\beta_t$ such that  $\beta_t<\tau_t$, we arrive at
\begin{align}
\int\limits_{0}^{\tau_t} 2x \prob(\hW_t\ge x)&= \int\limits_{0}^{\beta_t} 2x \prob(\hW_t\ge x)\de x+ \int\limits_{\beta_t}^{\tau_t} 2x \prob(\hW_t \ge x)\de x  \nonumber \\
&\leq \beta_t^2+ \int\limits_{\beta_t}^{\tau_t} 2x \prob(\hW_t\ge x)\de x \nonumber \,.
\end{align}
We next employ the tail bound on $\homg$ given in \eqref{eq: mse-homg-tail-gamma-1} to obtain
\begin{align}
\int\limits_{0}^{\tau_t} 2x \prob(\hW_t\ge x)& \leq \beta_t^2+ 
\int\limits_{\beta_t}^{\tau_t}  2x\frac{C'}{x^{1+\delta}}\nonumber\\
&=  \beta_t^2+ 2C' \int \limits_{\beta_t}^{\tau_t} \frac{1} {x^\delta}  \de x\,. \label{eq: general-delta-gamma-1}
\end{align}
 In the next step, by using \eqref{eq: general-delta-gamma-1} we get
 \begin{equation}\label{eq: tmp-upper-integration-gamma-1}
\int\limits_{0}^{\tau_t} 2x \prob(\hW_t\ge x)  \le 
\begin{cases}
\beta_t^2+2C'\frac{\tau_t^{1-\delta}}{1-\delta} & \delta<1\,,\\
\beta_t^2+ 2C'\log(\tau_t/\beta_t)  & \delta=1\,.
\end{cases}
\end{equation}
By an appropriate choice of $\beta_t$, specifically  $\beta_t=\tau_t^{\frac{1-\delta}{2}}$ for $\delta<1$ and $\beta_t=\log(\tau_t)^{\frac{1}{2}}$ for $\delta=1$  in \eqref{eq: tmp-upper-integration-gamma-1}, and then combining \eqref{eq: var-tmp-gamma-1} and \eqref{eq: tmp-upper-integration-gamma-1} we arrive at
\begin{align}\label{eq: var-truncated-gamma-1}
\var(\lambda_t) &\le B\E\left[(\hW_t \wedge \tau_t)^2\right] \nonumber\\
&\lesssim
\begin{cases}
\tau_t^{1-\delta} & \delta<1\,,\\
\log(\tau_t) & \delta=1\,.
\end{cases}
\end{align}
We suppose that $\tau_t$ is an non-decreasing sequence in $t$. Using \eqref{eq: var-truncated-gamma-1} in \eqref{eq: tmp-var-upper-gamma-1} gives us 
\begin{equation}\label{eq: var-tdr-gamma'-1}
\var\left(\frac{1}{T}\sum\limits_{t=1}^{T}\lambda_t\right)\lesssim  
\begin{cases}
\frac{1}{T^2}\sum\limits_{t=1}^T\tau_{t}^{1-\delta}+\frac{1}{T}\tau_T^{1-\delta} \,,& \delta<1 \,, \\ 
\frac{1}{T^2}\sum\limits_{t=1}^{T}\log(\tau_t)+ \frac{1}{T}\log \tau_T\,,& \delta=1\,. 
\end{cases}
\end{equation}
By combining \eqref{eq: long-run-var-delta_large} and
\eqref{eq: var-tdr-gamma'-1} we arrive at

\begin{equation}\label{eq: var-tdr-gamma-1}
\var\left(\frac{1}{T}\sum\limits_{t=1}^{T}\lambda_t\right)\lesssim  
\begin{cases}
\frac{1}{T^2}\sum\limits_{t=1}^T\tau_{t}^{1-\delta}+\frac{1}{T}\tau_T^{1-\delta} \,,& \delta<1 \,, \\ 
\frac{1}{T^2}\sum\limits_{t=1}^{T}\log(\tau_t)+ \frac{1}{T}\log \tau_T\,,& \delta=1\,,\\
\frac{1}{T}\,,& \delta>1\,.
\end{cases}
\end{equation}

We next move to upper bound the bias term in \eqref{eq: mse-numerator}. In particular, we have

\begin{align}
\frac{1}{T}\sum\limits_{t=1}^{T}\E[\lambda_t]&=\frac{1}{T}\sum\limits_{t=1}^T  \E[\eta_t (\hW_t\wedge \tau_t)(R_t+\hV_{t+1}-\hQ_t-\myavg)]\nonumber\\
&=\frac{1}{T}\sum\limits_{t=1}^{T}\E[ ((\hW_t\wedge \tau_t)-\hW_t)\eta_t(R_t+\hV_{t+1}-\hQ_t-\myavg)]\nonumber\\
&\, +\frac{1}{T}\sum\limits_{t=1}^T  \E[\eta_t \hW_t(R_t+\hV_{t+1}-\hQ_t-\myavg)]\nonumber \,.
\end{align}
We next use \eqref{eq: boundedness-lambda} and doubly robust lemma \ref{lemma-second-little-o-gamma-1} in the above to get
\begin{align}\label{eq: tmp-bias-upper-bound-gamma-1}
\frac{1}{T}\sum\limits_{t=1}^{T}\E[\lambda_t]&\le C_\eta K_T \Xi_T+ \frac{B}{T}\sum\limits_{t=1}^{T}\E\left[\hW_t-\hW_t\wedge \tau_t\right]\,.
\end{align}

We next provide an upper bound on $\E\left[\hW_t-\hW_t\wedge \tau_t\right]$. For the case $\delta>1$, as we discussed in the variance upper bound above, we have 
$\E_{p_b}[\homg(S)^2]<+\infty$, so by using Markov's inequality we arrive at
\begin{align*}
\E[\hW_t-\hW_t\wedge \tau_t]
&=\int\limits_{\tau_t}^{\infty}\prob(\hW_t\ge u)\de u\\
&\le \int\limits_{\tau_t}^{\infty} \frac{\E[\hW_t^2]}{u^2} \de u\\
&=\E[\hW_t^2]\frac{1}{\tau_t}\,.
\end{align*} 
This implies that 
\begin{align}\label{eq: bias-long-run-delta-large}
\E[\hW_t-\hW_t\wedge \tau_t] \lesssim \frac{1}{\tau_t}\,, \quad \text{ when }\delta >1\,.
\end{align}
We next move to the $\delta\le 1$ setting, and by using the tail inequality of $\homg$ given in \eqref{eq: mse-homg-tail-gamma-1} we obtain:
\begin{align*}
\E[\hW_t-\hW_t\wedge \tau_t]&=\int\limits_{\tau_t}^{\infty}\prob(\hW_t\ge u)\de u\\
&\le \int\limits_{\tau_t}^{\infty}\frac{C'}{u^{1+\delta}} \de u\,.
\end{align*}
This brings us
\begin{align}\label{eq: bias-bound-tmp-gamma-1}
\E\left[\hW_t-\hW_t\wedge \tau_t\right]&\le \frac{C'}{\delta\tau_t^\delta}\,, \quad \text{ when } \delta\le 1\,.
\end{align}
By employing \eqref{eq: bias-bound-tmp-gamma-1} and \eqref{eq: bias-long-run-delta-large} in \eqref{eq: tmp-bias-upper-bound-gamma-1} and utilizing the identity $(a+b)^2\le 2(a^2+b^2)$, we obtain the following result:
 \begin{equation}\label{eq: bias-gamma-1}
\left(\frac{1}{T}\sum\limits_{t=1}^{T}\E[\lambda_t] \right)^2  \lesssim 
\begin{cases} 
K_T^2\Xi_T^2  + \frac{1}{T^2} \left(\sum\limits_{t=1}^T \frac{1}{\tau_t^{\delta}}\right)^2\,, & \delta\le 1\,,\\
K_T^2\Xi_T^2  + \frac{1}{T^2} \left(\sum\limits_{t=1}^T \frac{1}{\tau_t}\right)^2\,, & \delta>1\,.
\end{cases}
\end{equation}

By using \eqref{eq: var-tdr-gamma-1} and \eqref{eq: bias-gamma-1} in \eqref{eq: mse-numerator}  we get the following
\begin{equation}\label{eq: main-mse-gamma-1}
(M-\eps)^2 \Delta_1 \lesssim
\begin{cases}
 K_T^2\Xi_T^2+ \frac{1}{T^2}\Big(\sum\limits_{t=1}^{T}\tau_t^{-\delta}\Big)^2  +\frac{1}{T^2}\sum\limits_{t=1}^{T}\tau_t^{1-\delta}+\frac{\tau_{T}^{1-\delta} }{T} \,, & \delta<1\,, \\
K_T^2\Xi_T^2+\frac{1}{T^2}\Big(\sum\limits_{t=1}^{T}\tau_t^{-1}\Big)^2 +\frac{1}{T^2}\sum\limits_{t=1}^{T}\log \tau_t+\frac{1}{T}\log(\tau_T) \,,  & \delta=1\,, \\ 
K_T^2\Xi_T^2+\frac{1}{T^2}\Big(\sum\limits_{t=1}^{T}\tau_t^{-1}\Big)^2 +\frac{1}{T}\,, & \delta>1\,. 
\end{cases}
\end{equation}

We next focus on providing an upper bound for \eqref{eq: big-tdr-2}. We have
 \begin{align}
 \int\limits_{0}^{B} 2x\prob\left( \Big|\frac{1}{T}\sum\limits_{t=1}^T (\hW_t \wedge \tau_t)\eta_t-M\Big| \ge \eps \right) \de x &\le B^2 ~ \prob\left( \Big|\frac{1}{T}\sum\limits_{t=1}^T (\hW_t \wedge \tau_t)\eta_t-M\Big| \ge \eps \right)\,. \nonumber\\
 &\le  {\E\left[\Big|\frac{1}{T}\sum\limits_{t=1}^T (\hW_t \wedge \tau_t)\eta_t-M\Big| ^2\right]} \frac{B^2}{\eps^2}\label{eq: mse-2-gamma-1} \,,
\end{align}
where in the last relation we used Markov's inequality. We then consider the following specific value $M=\E_{p_b}[\homg(S)]$. This implies that for $(S,A)\sim p_b(S) \pi_b(A|S)$, we have $M=\E[\homg(S)\eta(S,A)]$. Using this value of $M$ in the above yields
\begin{align}\label{eq: mse-infinity-gamma-1}
\E\left[\Big|\frac{1}{T}\sum\limits_{t=1}^T (\hW_t \wedge \tau_t)\eta_t-M\Big| ^2\right] =\var\left(\frac{1}{T}\sum\limits_{t=1}^T (\hW_t \wedge \tau_t)\eta_t \right)+
\left[\Big(\frac{1}{T}\sum\limits_{t=1}^T \E[(\hW_t \wedge \tau_t- \hW_t)\eta_t]\Big) ^2\right]\,.
\end{align}
By considering $\lambda'_t=(\hW_t \wedge \tau_t)\eta_t$ similar to the argument used earlier for $\lambda_t$ in \eqref{eq: var-tdr-gamma-1} we obtain

\begin{align}
\var\left(\frac{1}{T}\sum\limits_{t=1}^T (\hW_t \wedge \tau_t)\eta_t \right) \lesssim
\begin{cases}
\frac{1}{T^2}\sum\limits_{t=1}^T\tau_{t}^{1-\delta}+\frac{1}{T}\tau_T^{1-\delta} \,,& \delta<1 \,,\\ 
\frac{1}{T^2}\sum\limits_{t=1}^{T}\log(\tau_t)+ \frac{1}{T}\log \tau_T\,,& \delta=1\,, \\
\frac{1}{T}\,, & \delta>1\,.
\end{cases}
 \label{eq: self-normalization-1} 
\end{align}
In addition, using \eqref{eq: bias-bound-tmp-gamma-1} and \eqref{eq: bias-long-run-delta-large} we get

\begin{align}\label{eq: self-normalization-2} 
\Big(\frac{1}{T}\sum\limits_{t=1}^T \E\big[(\hW_t \wedge \tau_t- \hW_t)\eta_t\big]\Big) ^2 \lesssim
\begin{cases}
    \frac{1}{T^2} \left(\sum\limits_{t=1}^{T}\frac{1}{\tau_t^\delta}\right)^2\,, & \delta \le 1\,,\\
     \frac{1}{T^2} \left(\sum\limits_{t=1}^{T}\frac{1}{\tau_t}\right)^2\,, & \delta > 1\,.
\end{cases}
\end{align}

We next consider $\eps=\frac{M}{2}$. We then plug \eqref{eq: self-normalization-1} and \eqref{eq: self-normalization-2} into \eqref{eq: mse-infinity-gamma-1},  and next deploy \eqref{eq: mse-2-gamma-1} to get
\begin{align}
 \int\limits_{0}^{B} 2x\prob\left( \Big|\frac{1}{T}\sum\limits_{t=1}^T (\hW_t \wedge \tau_t)\eta_t-M\Big| \ge \eps \right)\lesssim
\begin{cases}
\frac{1}{T^2}\sum\limits_{t=1}^T\tau_{t}^{1-\delta}+\frac{1}{T}\tau_T^{1-\delta}+ \frac{1}{T^2} \left(\sum\limits_{t=1}^{T}\frac{1}{\tau_t^\delta}\right)^2 \,,& \delta<1 \,,\\ 
\frac{1}{T^2}\sum\limits_{t=1}^{T}\log(\tau_t)+ \frac{1}{T}\log \tau_T+ \frac{1}{T^2} \left(\sum\limits_{t=1}^{T}\frac{1}{\tau_t}\right)^2\,,& \delta=1\,, \\
\frac{1}{T}+ \frac{1}{T^2} \left(\sum\limits_{t=1}^{T}\frac{1}{\tau_t}\right)^2\,, & \delta>1\,.
\end{cases}
 \label{eq: self-normalization-mse} 
\end{align}

We then respectively use \eqref{eq: main-mse-gamma-1} and \eqref{eq: self-normalization-mse} in \eqref{eq: big-tdr-1} and \eqref{eq: big-tdr-2} to obtain

\begin{equation}\label{eq: main-mse-two-terms}
\E\left[(\mytdravg-\myavg)^2\right] \lesssim 
\begin{cases}
 K_T^2\Xi_T^2+ \frac{1}{T^2}\Big(\sum\limits_{t=1}^{T}\tau_t^{-\delta}\Big)^2  +\frac{1}{T^2}\sum\limits_{t=1}^{T}\tau_t^{1-\delta}+\frac{\tau_{T}^{1-\delta} }{T} \,, & \delta<1\,, \\
K_T^2\Xi_T^2+\frac{1}{T^2}\Big(\sum\limits_{t=1}^{T}\tau_t^{-1}\Big)^2 +\frac{1}{T^2}\sum\limits_{t=1}^{T}\log \tau_t+\frac{1}{T}\log(\tau_T) \,,  & \delta=1\,. \\ 
K_T^2\Xi_T^2+\frac{1}{T^2}\Big(\sum\limits_{t=1}^{T}\tau_t^{-1}\Big)^2 +\frac{1}{T}\,,  & \delta>1\,. \\ 
\end{cases}
\end{equation}

To strike a balance between different terms present in \eqref{eq: main-mse-two-terms},  for $\delta<1$, the truncation rate can be chosen as either $\tau_t=t^{\frac{1}{1+\delta}}$ or $\tau_t=T^{\frac{1}{1+\delta}}$. Both options result in an $\mse$ rate of $T^{-\frac{2\delta}{1+\delta}}\vee  K_T^2\Xi_T^2$. For the case of $\delta=1$, the truncation policies $\tau_t=\sqrt{t}$ or $\tau_t=\sqrt{T}$ can be applied, leading to an $\mse$ rate of $\frac{\log T}{T} \vee   K_T^2\Xi_T^2$. Finally, when $\delta>1$, having $\tau_t=t^{\alpha}$ or $\tau_t=T^{\alpha}$ for $\alpha\ge \frac{1}{2}$ results in $\mse$ convergence rate of 
$K_T^2\Xi_T^2 \vee \frac{1}{T}$. This completes the proof. 


}

\section{Proof of  Theorem \ref{thm: truncated-gamma-1-clt}} \label{proof: thm: truncated-gamma-1-clt}
We follow the same shorthands given in Section \ref{proof: thm: mse-gamma-1}. We start from the $\tdr$ estimator for the average reward estimation, we have
\begin{align*}
\mytdravg&=\frac{\sum\limits_{t=1}^{T} (\hW_t\wedge \tau_t) \eta_t (R_t+\hV_{t+1}-\hQ_t)}{\sum\limits_{t=1}^{T} (\hW_t\wedge \tau_t)\eta_t}\,.
\end{align*}
By subtracting the average reward value $\myavg$ from both sides, the $\tdr$ estimator can be written as the following
\begin{equation}\label{eq: rewritten-tdr}
\mytdravg-\myavg=\frac{\sum\limits_{t=1}^{T} (\hW_t\wedge \tau_t) \eta_t (R_t+\hV_{t+1}-\hQ_t-\myavg)}{\sum\limits_{t=1}^{T} (\hW_t\wedge \tau_t)\eta_t}\,.
\end{equation}
For the rest of the proof, for the above expression, we show that the numerator has CLT convergence to $\normal(0,\sigma_b^2)$, and the denominator convergence in probability to a constant term. We then employ Slutsky's theorem and gets the convergence in distribution result. We start by the numerator expression $\sum\limits_{t=1}^{T} (\hW_t\wedge \tau_t) \eta_t (R_t+\hV_{t+1}-\hQ_t-\myavg)$. In the interest of brevity we adopt the shorthands  $c_t=W_t\eta_t$, $\hc_t=\hW_t\eta_t$, $b_t=(R_t+V_{t+1}-Q_t-\myavg)$, and $\hb_t=(R_t+\hV_{t+1}-\hQ_t-\myavg)$. We have 
\begin{align}
\frac{1}{\sqrt{T}}\sum\limits_{t=1}^{T} (\hW_t\wedge \tau_t) \eta_t \hb_t
&~=\frac{1}{\sqrt{T}}\sum\limits_{t=1}^{T}  (\hW_t\wedge \tau_t -\hW_t) \eta_t \hb_t\label{eq: martingale-1} \\
&~\,+ \frac{1}{\sqrt{T}}\sum\limits_{t=1}^{T}\left(\hc_t \hb_t-c_tb_t-\E[\hc_t\hb_t]+\E[c_tb_t] \right)\label{eq: martingale-2}\\
&~\,+ \frac{1}{\sqrt{T}}\sum\limits_{t=1}^{T}\left(\E[\hc_t\hb_t]-\E[c_tb_t] \right)\label{eq: martingale-3}\\
&~\,+\frac{1}{\sqrt{T}}\sum\limits_{t=1}^{T}(W_t-W_t\wedge \tau_t)\eta_tb_t\label{eq: martingale-5}\\
&~\,+\frac{1}{\sqrt{T}}\sum\limits_{t=1}^{T}(W_t\wedge \tau_t)\eta_tb_t\label{eq: martingale-4}\,.
\end{align}
We claim that \eqref{eq: martingale-1}, \eqref{eq: martingale-2}, \eqref{eq: martingale-3}, and \eqref{eq: martingale-5} are $o_P(1)$ and we later establish the CLT convergence to $\normal(0,\sigma_b^2)$ for \eqref{eq: martingale-4}.  Expressions \eqref{eq: martingale-2} and \eqref{eq: martingale-3} are $o_P(1)$ following Lemmas  \ref{lemma-first-little-o-gamma-1} and \ref{lemma-second-little-o-gamma-1}.  We next consider the following value
\begin{equation*}
\beta\overset{\Delta}{=}\delta-1-\frac{1}{2}\cdot \min\left\{ \delta-1,\frac{1+\delta}{2}-\alpha,\alpha-\frac{2}{1+\delta}\right\}\,.
\end{equation*}
It is easy to check that $\delta-1>\beta>0$ and $\frac{2}{2+\beta}< \alpha < \frac{2+\beta}{2}$, given that $\frac{2}{1+\delta}< \alpha < \frac{1+\delta}{2}$. In addition, since $2+\beta<1+\delta$ and from $\delta$-weak distributional overlap assumption for the density ratio function $\omega(s)$ we have $\prob_{p_b}\big(\omega(S)^{1+\delta}>x\big)\le \frac{C}{x}$ and estimate assumption $\prob_{p_b}\big(\homg(S)^{1+\delta}>x\big)\le \frac{C'}{x}$ we get 
\begin{equation}\label{eq: E_2_beta-gamma-1}
\E_{p_b}[\omega(S)^{2+\beta}]<+\infty\,,\quad \E_{p_b}[\homg(S)^{2+\beta}]<+\infty\,. 
\end{equation}
For the expression \eqref{eq: martingale-1}, define
\[
Z_t=\hW_t\eta_t(R_t+\hV_{t+1}-\hQ_t)\,, \quad Z'_t=(\hW_t\wedge \tau_t)\eta_t(R_t+V_{t+1}-Q_t)\,.
\]
We will prove that $\frac{1}{\sqrt{T}}\sum_{t=1}^T (Z_t-Z'_t)=0$, almost surely. For this end, first not that we have 
\begin{align*}
\sum_{t=1}^\infty \prob(Z_t\neq Z'_t)&=\sum_{t=1}^\infty \prob(\hW_t > \tau_t)\nonumber\\
&\le \sum_{t=1}^\infty \frac{\E\Big[\hW_t^{2+\beta}\Big]}{\tau_t^{2+\beta}}\nonumber\\
&=\E\Big[\hW_t^{2+\beta}\Big] \sum_{t=1}^{+\infty}\frac{1}{t^{\alpha(2+\beta)}}<+\infty\,.
\end{align*}
where in the last relation we used the boundedness of $2+\beta$-th moment of $\homg(s)$ in along with inequality $\alpha (2+\beta)>1$. Using above for the Borel–Cantelli lemma give s us the following
\[
\prob\left(\lim\inf_{t\to \infty }\left \{s\in \cS: Z_t(s)=Z'_t(s)\right\}\right)=1\,.
\]
This implies that if $B=\bigcup\limits_{t=1}^{\infty} \bigcap\limits_{r\geq t}^{\infty} \{s\in \cS: Z_r(s)=Z'_r(s)\}$, then we have $\prob(B)=1$.  Concretely, for every  $s\in B$, there exists an integer $\Lambda(s)$ such that if $t\geq \Lambda(s)$ then $Z_t(s)=Z'_t(s)$. This implies that for each $s\in B $ the following holds

 \begin{align*}
 \lim\sup_{T\to \infty} \frac{1}{\sqrt{T}}\sum_{t=1}^T (Z_t(s)-Z'_t(s))&= \lim\sup_{T\to \infty} \frac{1}{\sqrt{T}}\sum_{t=1}^{\Lambda(s)} (Z_t(s)-Z'_t(s))=0\,.
 \end{align*}

Given that $\prob(B)=1$, this gives us the almost sure convergence of $\frac{1}{\sqrt{T}}\sum_{t=1}^T (Z_t-Z'_t)$ to zero, and therefore \eqref{eq: martingale-1} is $o_P(1)$. By a similar chain of arguments it can be shown that  \eqref{eq: martingale-5} is also $o_P(1)$.  We now proceed to prove the CLT convergence result for \eqref{eq: martingale-4}. We consider the following triangular array $\{G_{t,T}\}_{t\leq T, T\geq 1}$:
\[
G_{t,T}=\frac{1}{\sqrt{T}}\sum_{\ell=1}^{t} (W_\ell \wedge \tau_\ell)\eta_\ell(R_\ell+V_{\ell+1}-Q_\ell-\myavg)\,.
\]
For the filtration $\cF_{t}$ generated by $\{(S_\ell,A_\ell,R_\ell)\}_{\ell\le {t+1}}$ we show that $G_{t,T}$ is a triangular martingale array with respect to $\cF_t$. As a first step, it is easy to observe that $G_{t,T}$ is measurable with respect to $\cF_t$, and therefore implies adaptivity. We next show that $\E[G_{t,T}|\cF_{t-1}]=G_{t-1,T}$ and establish the martingale property:
\begin{align*}
\E[G_{t,T}|\cF_{t-1}]&=G_{t-1,T}+\frac{1}{\sqrt{T}} \E[(W_t \wedge \tau_t)\eta_t(R_t+V_{t+1}-Q_t-\myavg)|\cF_{t-1}]\\
&=G_{t-1,T}+\frac{1}{\sqrt{T}}\E\Big[((W_t\wedge \tau_t) \eta_t(R_t+ V_{t+1}-Q_t-\myavg)|S_t,A_t\Big]\\
&=G_{t-1,T}+\frac{1}{\sqrt{T}} (W_t\wedge \tau_t)\eta_t \E[(R_t+V_{t+1}-Q_t-\myavg)|S_t,A_t]\\
&=G_{t-1,T}\,,
\end{align*}
where the last relation follows Lemma \ref{lemma: bell-q}. 
Let $X_{\ell,T}=\frac{1}{\sqrt{T}} (W_\ell\wedge \tau_\ell) \eta_\ell(r_\ell+V_{\ell+1}-Q_\ell-\myavg)$, so $G_{t,T}=\sum_{\ell\le t }X_{\ell,T}$. We next show that $\{G_{t,T}\}$ is square-integrable.  We have

\begin{align*}
\E[G_{t,T}^2]=\sum_{\ell\le t}\E[X_{\ell,T}^2] +\sum_{\ell_1\neq \ell_2}\E[X_{\ell_1,T}X_{\ell_2,T}]\,.
\end{align*}
For $\ell_1<\ell_2$, again by using a similar argument used above by Lemma \ref{lemma: bell-q} we get $\E[X_{\ell_1,T}X_{\ell_2,T}]=\E[X_{\ell_1,T}\E[X_{\ell_2,T}|\cF_{\ell_2-1}]]=0$. This gives us
\begin{align*}
\E[G_{t,T}^2]&=\sum_{\ell\le t}\E[X_{\ell,T}^2]\\
&= \frac{t}{T}\E\left[(W_t \wedge \tau_t)^2\eta_t^2(R_t+V_{t+1}-Q_t-\myavg)^2\right]\\
& \le \sigma_b^2 <+\infty.  
\end{align*}

So far, we have shown that $G_{t,T}$ is a centered square-integrable triangular Martingale array. We then use \citep{hall2014martingale}, Corollary 3.1 to establish the CLT convergence result. For this purpose, we first show that the Lindeberg's condition holds. Formally, for every $\eps>0$ we have to prove
\begin{equation}\label{eq: lindeberg's conditon-gamma-1}
\sum_{t=1}^{T}\E[X_{t,T}^2\ind(|X_{t,T}|>\eps)|\cF_{t-1}]\overset{(p)}{\to} 0.
\end{equation}

This means that for every $c>0$ we must have
\begin{equation}\label{eq: Lind-gamma-1}
\lim_{T\to \infty} \prob\left(\sum_{t=1}^{T}\E[X_{t,T}^2\ind(|X_{t,T}|>\eps)|\cF_{t-1}]>c \right)=0
\end{equation}

We start by using Markov's inequality to obtain
\begin{align}
 \prob\left(\sum_{t=1}^{T}\E[X_{t,T}^2\ind(|X_{t,T}|>\eps)|\cF_{t-1}]>c \right)&\leq \frac{\E\left[\sum\limits_{t=1}^T X_{t,T}^2 \ind(|X_{t,T}|> \eps) \right]}{c}\nonumber \\
 &\leq \frac{1}{c}\E\left[\max_{1\le t\le T} |X_{t,T}|^2 \sum_{t=1}^T \ind(|X_{t,T}|>\eps)\right]\label{eq: martingale-6}\,.
 \end{align}
In the next step, by using Proposition \ref{prop-unif-bound} we get $X_{t,T}\lesssim \frac{\tau_t}{\sqrt{T}} $. Plugging this into \eqref{eq: martingale-6} yields

 \begin{align}\label{eq: linden-1-gamma-1}
 \prob\left(\sum_{t=1}^{T}\E\Big[X_{t,T}^2\ind(|X_{t,T}|>\eps)|\cF_{t-1}\Big]>c \right) &\lesssim \frac{\max\limits_{1\le t\le T}^{}{\tau_t}^2}{c T}\sum_{t=1}^{T}\prob(|X_{t,T}|>\eps)\,.
 \end{align}
 We next employ $X_{t,T}\lesssim \frac{W_t}{\sqrt{T}}$ and by an application of Markov's inequality we get 
 \[
 \prob(|X_{t,T}|\ge \eps)\lesssim \frac{\E_{p_b}[W(S)^{2+\beta}]}{\eps^{2+\beta}\, T^{\frac{2+\beta}{2}}}\,.
 \]
 Using this in \eqref{eq: linden-1-gamma-1} yields
 \begin{align*}
\prob\left(\sum_{t=1}^{T}\E[X_{t,T}^2\ind(|X_{t,T}|>\eps)|\cF_{t-1}]>c\right) & \lesssim \Big({\max\limits_{1\le t\le T}^{}{\tau_t}^2}\Big) \frac{\E_{p_b}[W(S)^{2+\beta}]}{c\eps^{2+\beta}\, T^{1+\frac{\beta}{2}}}\,.
\end{align*}
Having $\tau_t=t^{\frac{\alpha}{2}}$ gives us 
\[
\frac{\max\limits_{1\le t\le T}{\tau_t}^2}{T^{1+\frac{\beta}{2}}}=T^{\alpha- \frac{2+\beta}{2} }\,.
\]
Using this in the earlier expression, in along with  the fact that $\alpha< \frac{2+\beta}{2}$ brings us
\[
\lim_{T\to \infty}\prob\left(\sum_{t=1}^{T}\E[X_{t,T}^2\ind(|X_{t,T}|>\eps)|\cF_{t-1}]>c\right)=0\,.
\]
This completes the proof for \eqref{eq: Lind-gamma-1}. Having shown the Lindeberg's condition \eqref{eq: lindeberg's conditon-gamma-1}, the only remaining part for establishing CLT is to show the following:
 \begin{equation}\label{eq: var-stability-gamma-1}
 \sum_{t=1}^{T} X_{t,T}^2\overset{(p)}{\to} \sigma_b^2\,.
 \end{equation}
  We have
 \begin{align*}
 \E\Big[W_t^2-(W_t\wedge \tau_t)^2\Big]= \int\limits_{\tau_t}^{+\infty} 2x \prob(W_t\ge x)\de x\,.
 \end{align*}
 Using Markov's inequality we get
  \begin{align*}
   \E\Big[W_t^2-(W_t\wedge \tau_t)^2\Big]& \le \int\limits_{\tau_t}^{+\infty} \frac{2\E_{p_b}[\omega(S)^{2+\beta}]}{x^{1+\beta}}\\
   &=  \frac{2\E_{p_b}[\omega(S)^{2+\beta}]}{\tau_t^{\beta}}\,.
  \end{align*}
  Using $\tau_t=t^{\alpha/2}$ in along with \eqref{eq: E_2_beta-gamma-1} we arrive at
  \begin{align}
 \frac{1}{T}\sum_{t=1}^T \E\Big[W_t^2-(W_t\wedge \tau_t)^2\Big] &\lesssim  \frac{1}{T} \E_{p_b}\Big[\omega(s)^{2+\beta}\Big]\sum_{t=1}^T t^{-\frac{1}{2}\alpha \beta}\nonumber\\
 &\lesssim T^{-\frac{1}{2}\beta\alpha}\label{eq: lind-4-gamma-1}\,.
 \end{align}
 This implies that 
 \begin{align}
 \left|\sum_{t=1}^T\E\Big[X_{t,T}^2\Big]-\sigma_b^2\right|&=\left|\frac{1}{T}\sum_{t=1}^T\E\left[\big(W_t^2\wedge \tau_t^2-W_t^2\big)\eta_t^2(R_t+V_{t+1}-Q_{t}-\myavg)^2\right]\right|\nonumber\\
 &\lesssim \frac{1}{T}\sum_{t=1}^T\E\left[\big(W_t^2-W_t^2\wedge \tau_t^2\big) \right]\label{eq: lind-5-gamma-1}\,.
 \end{align}
 Finally, by employing \eqref{eq: lind-4-gamma-1} in \eqref{eq: lind-5-gamma-1} we arrive at
 \[
\lim\limits_{T\to \infty}^{} \sum_{t=1}^T\E\Big[X_{t,T}^2\Big]=\Sigma_b^2\,.
\]
This proves \eqref{eq: var-stability-gamma-1}. Going back to the primary expression of interest \eqref{eq: rewritten-tdr}, so far we have shown that
\begin{equation}\label{eq: numerator}
\frac{1}{\sqrt{T}}\sum\limits_{t=1}^{T} (\hW_t\wedge \tau_t) \eta_t (R_t+\hV_{t+1}-\hQ_t-\myavg)\to \normal(0,\Sigma_b^2)\,.
\end{equation}
In addition, we have
\begin{align}
\frac{1}{T}\sum\limits_{t=1}^{T}(\hW_t\wedge \tau_t)\eta_t&=\frac{1}{T}\sum\limits_{t=1}^{T}(\hW_t\wedge \tau_t-\hW_t)\eta_t \label{eq: denom-1}\\
&\, + \frac{1}{T}\sum\limits_{t=1}^{T}(\hW_t-W_t)\eta_t\label{eq: denom-2}\\
&\, +\frac{1}{T}\sum\limits_{t=1}^{T}W_t\eta_t\label{eq: denom-3}\,.
\end{align}
Similar to the previous arguments used earlier for \eqref{eq: martingale-1} and \eqref{eq: martingale-5} we can get \eqref{eq: denom-1} is $o_P(1)$. For \eqref{eq: denom-2}, by using Assumption \ref{assumption: model-estimates-gamma-1} we obtain
$ \frac{1}{T}\sum\limits_{t=1}^{T}\E[|\hW_t-W_t|\eta_t] \le \Xi_T C_\eta$. Given that $\Xi_T=o_p(1)$, we realize that \eqref{eq: denom-2} is also $o_P(1)$. Finally, for \eqref{eq: denom-3}, given that $\E[W_t \eta_t]=1$, we have
\begin{align*}
\var\left(\frac{1}{T}\sum\limits_{t=1}^{T}(W_t\eta_t-1)\right)&\le \frac{C_\eta^2 \E_{p_b}[\omega(S)^2]}{T}+ \frac{1}{T^2}\sum\limits_{k\neq \ell}{} \cov(W_k\eta_k, W_\ell\eta_\ell)\\
&\le \frac{C_\eta^2 \E_{p_b}[\omega(S)^2]}{T} +  \frac{C_\rho C_\eta^2 \E_{p_b}[\omega(S)^2] }{T}
\end{align*}
This implies that  $\var(\frac{1}{T}\sum\limits_{t=1}^{T}(W_t\eta_t-1))$ is $o_p(1)$, and by Chebyshev's inequality we realize that $\frac{1}{T}\sum\limits_{t=1}^{T}(W_t\eta_t-1)=o_P(1)$ as well. Put all together, we get the following
\begin{equation}\label{eq: denom}
\frac{1}{T}\sum\limits_{t=1}^{T}(\hW_t\wedge \tau_t)\eta_t \overset{(p)}{\to} 1\,.
\end{equation}
Combining \eqref{eq: numerator} and \eqref{eq: denom} and by an application of Slutsky's theorem we realize that
\[
 \frac{\sqrt{T} \sum\limits_{t=1}^{T}(\hW_t\wedge \tau_t)\eta_t(R_t-\hV_{t+1}+Q_t-\myavg) }{\sum\limits_{t=1}^{T}(\hW_t\wedge \tau_t)\eta_t } \to \normal(0,\Sigma_b^2)\,.
\] 
This completes the proof.

{ \section{Details on $\omega^{\gamma}(.;p_0)$ estimation process of Algorithm \ref{alg: omega-estimate}}
\label{sec: omega-estimation}
By recalling the Bellman equation given in Lemma \ref{lemma: bell-prob}. Under Assumptions \ref{assu:MDP} and \ref{assu:random}, for $(S,A,S')$ (under behavior policy) we have
\[
\E[\omega^{\gamma}(S;p_0)f(S)]=(1-\gamma)\int f(s) p_0(s)\de \lambda_{\cS}(s) +\gamma \E\left[ \omega^{\gamma}(S;p_0)\frac{\pi_e(A|S)}{\pi_b(A|S)}f(S') \right]\,.
\]
For this example, we consider the off-policy evaluation problem for the stationary regime of the evaluation policy with $p_0=p_e$. In this case, $\omega^{\gamma}(s;p_e)=\frac{p_e(s)}{p_b(s)}$, with $p_e(\cdot)$ and $p_b(\cdot)$ denoting the stationary state distributions induced by the evaluation and behavior policies. Hence, for $p_0=p_e$ we arrive at
\[
\int f(s) p_0(s) \de \lambda_{\cS} (s)=
\int f(s) \omega^{\gamma}(s;p_e) p_b(s) \de \lambda_{\cS} (s)=\E[f(S) \omega^{\gamma}(S;p_e)]\,.
\]
Using this in the earlier relation gives us
\begin{equation}\label{eq: omega-estimate-exp}
\E[f(S) \omega^{\gamma}(S;p_e)]=\E\left[ \omega^{\gamma}(S;p_e)\frac{\pi_e(A|S)}{\pi_b(A|S)}f(S') \right]\,,
\end{equation}
where this must be held for all measurable functions $f$. We next use this moment condition to estimate the value of $\omega^{\gamma}(S;p_e)$, where \cite{liu2018breaking} establishes the uniqueness of $\omega^{\gamma}(.;p_e)$ as the only solution satisfying this relation under certain conditions. For a single behavior trajectory $\{(S_i,A_i,R_i)\}_{i=1:T}$ and for a finite number of states $\cS=\{1,2,\dots,m\}$, consider the indicator function $f(s)=e_s$, where $e_j$ is the basis vector with exactly one nonzero entry at its $j$-th coordinate equal to one—this choice for $f(s)$ is based on the guideline for estimating $\omega(.)$ given in \cite{kallus2022efficiently}. For simplicity, let $\beta$ be the unknown parameter to estimate with $\beta_j=\omega^{\gamma}(j;p_e)$. Using the ergodicity of the process we know that as $T$ grows to infinity then $\frac{1}{T-1} \sum_{t=1}^{T-1}f(S_t)\beta_{S_t}\to \E[f(S)\omega^{\gamma}(S;p_e)]$, rewriting this for each state $j\in[m]$ we obtain (by using sample average for both sides of \eqref{eq: omega-estimate-exp}):
\begin{align*}
\frac{1}{T-1} \sum_{t=1}^{T-1} \ind(S_t=j)\beta_j&= \frac{1}{T-1} \sum_{t=1}^{T-1}\ind(S_{t+1}=j)\beta_{S_t} \frac{\pi_e(A_t|S_t)}{\pi_b(A_t|S_t)}\\
&=\frac{1}{T-1}\sum_{t=1}^{T-1}\ind(S_{t+1}=j)\sum_{i=1}^m\ind(S_t=i)\beta_{S_t} \frac{\pi_e(A_t|S_t)}{\pi_b(A_t|S_t)}\,.
\end{align*}
For brevity, we introduce the matrix $M^{(T)}\in \reals^{m\times m}$ such that
\[
M^{(T)}_{j,i}=\sum_{t=1}^{T-1} \ind(S_t=i,S_{t+1}=j) \frac{\pi_e(A_t|S_t=i)}{\pi_b(A_t|S_t=i)}\,, \quad \forall i,j \in [m]\,.
\]
Using this in the above yields, for every $j\in [m]$,
\[
\beta_j \sum_{t=1}^{T-1} \ind(S_t=j)=\sum_{i=1}^m M^{(T)}_{j,i}\beta_i\,.
\]
By considering $N_j=\sum_{t=1}^{T-1} \ind(S_t=j)$; then the above relation becomes
\begin{equation}\label{eq: omega-estimage-balance}
\mathsf{diag}([N_1,\dots,N_m])\beta=M^{(T)}\beta\,.
\end{equation}
In addition, given that $\omega$ is the density ratio in the stationary, we must have $\E_{\pi_b}[\omega(S)]=1$; then the ergodic theorem reads as
\[
\lim\limits_{T\to \infty}\frac{1}{T}\sum \limits_{t=1}^{T}\beta_{S_t}=1\,.
\]
This implies that we must have
\[
\lim\limits_{T\to\infty}^{}\frac{1}{T}\sum_{j=1}^{m} N_j\beta_j=1\,.
\]
Finally, letting $c=\frac{1}{T-1}\big[N_1,\dots,N_m\big]$, $H=\mathsf{diag}([N_1,\dots,N_m])-M^{(T)}$, to balance the both sides of \eqref{eq: omega-estimage-balance}, we try to minimize $\|H\beta\|$ while respecting the density ratio property of $\omega^{\gamma}(.;p_e)$, i.e., $c^\sT \beta=1$. In addition, as $\beta$ is the density ratio, it must have non-negative coordinates. Finally,  this gives us the following estimator for $\omega^\gamma(.;p_e)$:
\[
\hat{\beta}=\arg\min_{\beta \in \reals^m} \|H\beta \|_2^2\,,\quad \text{s.t.,} \quad c^\sT \beta=1\,,\quad \beta_i\ge0\,,~~\forall i \in [m]\,.
\]



}

{\color{blue}

}


\end{document}